%% file: manuscript_v3.tex
\documentclass{article} % For LaTeX2e
\usepackage{iclr2025_conference,times}

\input{math_commands.tex}

\DeclareMathOperator{\dist}{dist}

\usepackage{hyperref}
\usepackage{url}
\usepackage{algorithm}
\usepackage{algorithmic}
\usepackage{booktabs}
\usepackage{multirow}
\usepackage{pdflscape}
\usepackage{graphicx}
\usepackage{amsmath,amssymb}
\usepackage{mathtools}
\usepackage{amsthm}
\usepackage{enumitem}

\theoremstyle{plain}
\newtheorem{theorem}{Theorem}
\newtheorem{corollary}{Corollary}
\newtheorem{proposition}{Proposition}
\newtheorem{lemma}{Lemma}
\newtheorem*{conjectureR}{Conjecture~R}
\theoremstyle{definition}
\newtheorem{definition}{Definition}

\title{EB-gMCR: Energy-Based Generative \\Modeling for Signal Unmixing and \\ Multivariate Curve Resolution}

\author{
Yu-Tang Chang \quad Shih-Fang Chen \\
Department of Biomechatronics Engineering \\
National Taiwan University \\
\texttt{\{b05611038,sfchen\}@ntu.edu.tw}
}

\iclrfinalcopy


\begin{document}

\maketitle
\lhead{Preprint}

\begin{abstract}
A single measurement of a chemical mixture, a reaction mixture,
a natural extract, or a tissue, records the sum of the profiles
of the few components it contains, each weighted by its
concentration. Recovering the components and their
concentrations from a collection of such samples is
multivariate curve resolution (MCR). Classical MCR factorizes
the data matrix, takes the component count as input, and
leaves a rotational ambiguity that constraints narrow. This
paper keeps the forward direction and models how a sample is
made: each sample activates a few components from a pool of candidate components and is observed as their linear superposition plus noise. Under this model the continuous ambiguity of factorization collapses to the numbering of the components, and among all decompositions that reproduce the data, the one with the fewest component occurrences across samples is the true one. We prove three results. Under a spark condition, minimal usage identifies the true supports. A decomposition that reconstructs every sample within a tolerance set by the noise, with no more usage than the truth, is the truth up to the numbering of the components, and the recovered pool then decodes new samples on its own. As samples accumulate, what is recovered almost surely is the process itself, the pool and the component count; the decoding of any single sample stays limited by the noise, for every method. The solver EB-gMCR selects components per sample with an
energy-based gate under a usage penalty. It recovers the
component count on synthetic mixtures of up to $256$ components
and on two public spectroscopy datasets without being told the
count, and a frozen model decodes unseen mixtures at the noise
floor. Code: \url{https://github.com/b05611038/ebgmcr_solver}.
\end{abstract}

\section{Introduction}
\label{sec:intro}

A single measurement of a complex sample, such as a reaction mixture, a natural extract, or a tissue, records many chemical constituents at once. No part of the measurement belongs to one constituent alone. Determining which constituents are present, and at what concentrations, requires separating the measurement into one contribution per constituent. The measurement makes this
separation both possible and difficult. Each constituent, hereafter a component, contributes a fixed component profile. The profiles add in proportion to the concentrations, and the sensor records only the sum. The same structure holds for any
sensor that records a sum of non-interacting sources: the observed sample is a linear superposition of fixed component profiles.
Unmixing recovers the composition from the measurement. Diagnosis, reaction monitoring, and library search all require the composition.

Unmixing is ill-posed for two reasons. Both the component profiles and the concentrations are unknown, and both must be estimated from the samples. A product of two unknown factors has infinitely many exact factorizations. In chemistry this task
is multivariate curve resolution (MCR) \citep{dejuan2021mcr}:
given $M$ samples with $d$ features,
$D \in \mathbb{R}^{M \times d}$, find $N$ component profiles
$S \in \mathbb{R}^{N \times d}$ and their concentrations
$C \in \mathbb{R}^{M \times N}$ with
\begin{equation}
D = CS + \varepsilon,
\label{eq:1}
\end{equation}
where $\varepsilon$ is measurement noise. Classical methods
compute the factorization directly. MCR by alternating least squares (MCR-ALS), non-negative matrix factorization (NMF), and related methods update $C$ and $S$ in turn, under constraints such as non-negativity, closure, and unimodality, until $CS$ reproduces $D$. The constraints enforce physical plausibility and narrow the feasible set \citep{dejuan2021mcr}. This approach
has three limitations. First, the number of components $N$ is
an input, although it is unknown and often the quantity of
interest. Second, any invertible $T$ yields $CT$ and $T^{-1} S$ with the same fit. Constraints narrow this rotational ambiguity, and remove the ambiguity only under selectivity and local-rank conditions \citep{tauler1995selectivity, dejuan2021mcr}. Third, a
factorization describes one data matrix. The feasible set
depends on which samples the matrix contains
\citep{tauler1995selectivity, dejuan2021mcr}, and a new sample
from the same process requires a new factorization. The formulation below is different. Real samples have structure that
the factorization does not exploit: each sample contains few
components, while the pool of candidates numbers in the
hundreds \citep{barcaru2017bayesian, heinecke2021bayesian}. The
model below encodes this structure as a sparse selection from a
pool. Under this model the solution set is still not a single point. The paper studies the structure of that set.

This paper models the mixing process directly. The generative MCR model (gMCR, \S\ref{sec:gmcr}) fixes a pool of candidate components. Each sample activates a small set of components at sample-specific concentrations, and the observed sample is the linear superposition of those components plus noise. This paper proves three properties of this model: identifiability of the generative process, certified recovery from finite noisy
data, and large-sample limits. Under gMCR the per-sample support determines which components explain each sample, and the remaining
ambiguity is a relabeling of the pool. Usage is the total number of components active over all samples. Minimal usage identifies the generative process: a decomposition with higher usage explains
at least one sample with a component that sample does not
contain. Minimal usage makes the problem well-posed. Attaining minimal usage is the computational task of the solver EB-gMCR. The contributions are as follows.
\begin{itemize}[leftmargin=2em,itemsep=2pt]
\item \textbf{Identifiability by minimal usage.} Assume a spark condition on the pool: no set of up to $2k$ component profiles can nearly cancel, where $k$ is the per-sample sparsity. Then the true supports are the unique minimizer of total usage among exact factorizations over the pool, and a usage penalty with weight in the $\lambda$-window of Theorem~\ref{thm:ident} prefers those supports.
\item \textbf{A certificate for finite noisy data.} With the fit tolerance set by the noise, a decomposition passes the certificate when it reconstructs every sample within that tolerance and uses no more components than the generative process. A decomposition that passes is that process: it matches the component count and every per-sample support exactly, and the concentrations to the accuracy the noise permits (Theorem~\ref{thm:cert}). The certificate is a test on the returned solution. A pool that passes the certificate decodes new samples of the same process (Corollary~\ref{cor:reuse}).
\item \textbf{Large-sample limits.} As the sample count grows,
the component count and the generative process are recovered
almost surely up to relabeling, while no estimator decodes every sample correctly (Theorem~\ref{thm:limits}).
\item \textbf{An energy-based solver.} EB-gMCR implements
per-sample selection with an energy-based gate, trains under a
usage penalty, and returns the checkpoint with the fewest
components at each reconstruction quality. Proposition~\ref{prop:dyn} gives the convergence mechanism. Its component count is within $4\%$ of the true count with $8N$ samples, for up to $256$ components, at a signal-to-noise ratio (SNR) of 30\,dB. It recovers the component count of two public spectroscopy datasets with no count supplied. A frozen model decodes held-out samples with error at the noise level (\S\ref{sec:exp}).
\end{itemize}

\section{Related Work}
\label{sec:related}

\paragraph{When is the factorization unique?}
Several literatures share the bilinear model $D = CS + \varepsilon$ of Eq.~\ref{eq:1}. They differ in which of the unknowns are given: the pool $S$, the component count $N$, the per-sample supports, or the sign of the factors. Each literature has its own answer to when the factorization is unique.
With the pool known, the problem is sparse coding and compressed sensing: recover a sparse concentration vector \citep{donohoelad2003optimally, candes2006robust, donoho2006compressed}.
With the pool learned and the component count fixed, the problem
is dictionary learning: a sparsely used dictionary is
identifiable \citep{aharon2006uniqueness, spielman2012exact},
stable under noise \citep{garfinkle2019uniqueness}, and learnable
from a bounded number of samples \citep{gribonval2015sample}.
With non-negative factors and the count fixed, the problem is NMF. NMF is identifiable under separability-type and sufficiently scattered conditions \citep{donoho2003parts, fu2019nmfidentifiability}. In chemometrics the same problem is MCR. Its solution set was first described as a feasible region for two components \citep{lawton1971smcr}. Non-negativity alone leaves the set ambiguous, because rotational and intensity freedom can survive that constraint \citep{tauler1993mcr}. The factorization is unique under selectivity and local-rank conditions \citep{tauler1995selectivity, sawall2020ambiguity}. The gMCR model of \S\ref{sec:gmcr} takes fewer unknowns as given than any of these settings: the pool is
learned, the component count is unknown, each sample has a
sparse support, and the concentrations are non-negative. Every result above fixes the count in advance. Theorem~\ref{thm:ident} is the identifiability result with the count unknown and determined by minimal usage.

\paragraph{Choosing the count, and selecting per sample.}
Two practical literatures bear on the experiments of \S\ref{sec:exp}. In prior work the count is chosen before the fit. Chemometrics reads the count from the eigenvalues of the data matrix \citep{malinowski1977indicator} or from the rank of growing submatrices \citep{maeder1987efa, keller1991efa}. Automatic relevance determination for NMF (ARD-NMF) prunes an oversized factorization to a count determined by the data; that count depends on a noise-scale hyperparameter \citep{tan2013automatic}. The solvers compared in \S\ref{sec:exp} take the count as input \citep{tauler1995selectivity, lee1999nmf, hoyer2004sparse, schmidt2009bayesian, comon1994ica}.
Spike-and-slab sparse coding \citep{goodfellow2012s3c}, nonparametric Bayesian dictionary learning \citep{zhou2009bpfa}, and, in chemistry, Bayesian library search \citep{barcaru2017bayesian, heinecke2021bayesian} share one model: a binary selection variable per sample and per component, with the number of components inferred. gMCR uses the same model and adds the identifiability, the certificate, and the limits. Consistency of estimators defined up to relabeling
\citep{wald1949note, redner1981note, pollard1981strong} is the
statistical counterpart of Theorem~\ref{thm:limits}.

\section{The Generative MCR Model}
\label{sec:gmcr}

% --- opening: A; definition: A's equation + B's figure ---
The classical MCR formulation represents mixed signals as a linear
combination of $N$ unknown components. In principle $N$ should be
large enough to capture the true solution space (library searches
can involve thousands of candidates), yet specifying a large $N$
is numerically unstable and overfits noise, which is why most MCR
approaches restrict $N$ to a small number. The gMCR formulation
recasts MCR as a generative process: each sample
activates a small set of components, draws their concentrations,
and observes their linear superposition plus measurement noise.
Only the samples are observed; the pool, the supports, and the
concentrations are all latent and are the objects of inference.
The model serves to decode samples, not to synthesize them.
This paper analyzes linear superposition throughout.

The explicit modeling of the selection indicator $\delta$, defined
formally in the model below, is central to gMCR.
Classical MCR implicitly encodes selection through concentration magnitude: a component is ``absent'' when its concentration is near zero.
In gMCR, $\delta$ explicitly separates selection (which components participate) from intensity (how much each contributes), enabling discrete component discovery rather than soft shrinkage toward zero.

\paragraph{The gMCR model.}
A pool is a set of unit-norm component profiles
$s_1, \dots, s_N \in \R^d$, fixed across samples. A sample, indexed
by a draw $\omega$ of the generative process, is
\begin{equation}
x(\omega) \;=\; \sum_{i \in J(\omega)} c_i(\omega)\, s_i
\;+\; \varepsilon(\omega),
\qquad
|J(\omega)| \le k,
\quad
c_i(\omega) \in [c_{\min},\, c_{\max}]
\;\; \text{for } i \in J(\omega),
\label{eq:model}
\end{equation}
where $J(\omega) \subseteq [N] = \{1, \dots, N\}$ is the sample's
support ($\delta_i = \mathbf{1}[i \in J]$ recovers the indicator
above) and the noise $\varepsilon(\omega)$ is sub-Gaussian with proxy
$\sigma$ (formal definition in the theory appendix). Each constraint
has a physical reading: $k$ caps how many components any single sample
contains (few analytes per specimen); $c_{\min}$ is the detection
floor, so that active means detectably active; $c_{\max}$ is the
dynamic range; unit-norm profiles fix the scale ambiguity between a
profile and its concentration; sub-Gaussian $\varepsilon$ is
instrument noise with controlled tails. This constraint set is the
mathematical form of standard assumptions in spectroscopic unmixing:
linear detector response, a detection floor, finite dynamic range,
and few analytes per sample. Any domain satisfying them inherits
every result of this paper.

\begin{definition}[Spark condition with margin]
\label{def:spark}
The pool satisfies the \emph{spark condition with margin}
$s_{\min} > 0$ if $\sigma_{\min}(S_A) \ge s_{\min}$ for every
$A \subseteq [N]$ with $|A| \le 2k$, where $S_A$ collects the
profiles indexed by $A$ as columns.
\end{definition}

In words: no group of up to $2k$ profiles can nearly cancel; small
subsets of the pool are not merely linearly independent but
independent with a quantitative margin. This is the quantitative
form of the spark condition of \citet[Definition~1]{donohoelad2003optimally}.
Here $\sigma_{\min}(S_A)$, the smallest singular value, is computed
from the pool, whereas $s_{\min}$ is an assumed constant of the
model: the condition asserts that the computed quantity never falls
below the assumed margin. In particular the condition requires
$2k \le d$; it constrains the per-sample sparsity against the
feature dimension, not the pool size $N$, which may exceed $d$.
Two clarifications guard common misreadings. First, the condition
constrains the profiles themselves, not the selection indicators:
which combinations of components occur plays no role here (such
occurrence conditions appear separately, in \S\ref{sec:theory}).
Second, visually similar profiles are typically still linearly
independent; what similarity degrades is the margin $s_{\min}$, and
the margin, not bare independence, is what every guarantee
consumes.

\begin{proposition}[The decomposition is well-defined]
\label{prop:welldef}
Assume the spark condition with margin $s_{\min}$. Let
$x_0 = \sum_{i \in J} c_i s_i$ with $|J| \le k$ and
$c_i \ge c_{\min}$ on $J$. If a pair $(J', c')$ with $|J'| \le k$
and $c'_i \ge c_{\min}$ on $J'$ satisfies
$\bigl\| x_0 - \sum_{i \in J'} c'_i s_i \bigr\| \le \epsilon$ for
some $0 \le \epsilon < c_{\min}\, s_{\min}$ (a deterministic
tolerance, distinct from the model noise $\varepsilon(\omega)$),
then $J' = J$ and
$|c'_i - c_i| \le \epsilon / s_{\min}$ for every $i \in J$. In
particular, at $\epsilon = 0$ the $k$-sparse decomposition of a
noiseless sample is unique among decompositions with coefficients
at least $c_{\min}$.
\end{proposition}

\begin{proof}[Proof sketch]
Set $v_i = c_i - c'_i$ on $A = J \cup J'$, extending both coefficient
vectors by zero. Then
$\|S_A v\| = \bigl\| x_0 - \sum_{i \in J'} c'_i s_i \bigr\| \le
\epsilon$ and $|A| \le 2k$, so the spark margin gives
$s_{\min} \|v\| \le \epsilon$, hence
$|v_i| \le \|v\| \le \epsilon / s_{\min} < c_{\min}$ for every $i$. An index
$i \in J \setminus J'$ would have $|v_i| = c_i \ge c_{\min}$, which
is impossible; symmetrically $J' \setminus J = \emptyset$. Hence
$J' = J$, and the coefficient bound follows. The formal statement
and proof are given in Appendix~\ref{app:prop1}.
\end{proof}

The quantity $c_{\min} s_{\min}$, the weakest admissible
concentration times the pool margin, is the threshold that governs
every guarantee in this paper.
Proposition~\ref{prop:welldef} makes each sample's decomposition
well-defined \emph{given} the pool; the uniqueness of the pool
itself is a separate question, taken up in \S\ref{sec:theory}.

\paragraph{Reusability.}
Because the target of inference is the generative process rather
than a fixed data matrix, whatever recovers the process decodes new
samples from the same process without re-estimation; decoding a
fresh sample over a known pool is exactly the setting of
Proposition~\ref{prop:welldef}. A toy illustration is given in
Appendix~\ref{app:toy}.

% --- A's well-posedness paragraph, verbatim ---
\section{What Can Be Known from Mixtures: The Theory}
\label{sec:theory}

Solving MCR is an inverse problem: given the observed samples,
infer the pool, the supports, and the concentrations. Posed over
the unconstrained factorization $D = CS$, the problem is
ill-posed: any invertible $T$ yields another solution with
identical fit, the ambiguity persists at the correct rank, and
non-negativity alone, the constraint of the NMF family, does not
remove it in general. The gMCR model changes the object:
selection is combinatorial, and invertible transformations do not
preserve the constraint set of Eq.~\ref{eq:model}, so the target
is well-defined (\S\ref{sec:gmcr}). Three theorems connect this target to a working solver.
Theorem~\ref{thm:ident} gives the objective: its optimum is the
true decomposition. Theorem~\ref{thm:cert} gives the check: a
solution that passes the certificate is the truth, no matter how
it was found. Theorem~\ref{thm:limits} gives the limit: not every sample
can be decoded, but the generative process is recovered as
samples accumulate. One combinatorial statement remains open
(Conjecture~R, \S\ref{sec:theory-frontier}).\footnote{Formal statements and
proofs: Theorem~1 in Appendix~\ref{app:thm1},
Theorem~2 and Corollary~1 in Appendix~\ref{app:thm2}, Theorem~3 in
Appendix~\ref{app:thm3}.} Throughout, the
data follow the gMCR model of \S\ref{sec:gmcr}
(Eq.~\ref{eq:model}) with the spark condition of
Definition~\ref{def:spark}.

\subsection{A Criterion That Points at the Truth}
\label{sec:theory-t1}

The solver still requires an explicit target to search.
\S\ref{sec:gmcr} shows the true decomposition exists; what
remains is a searchable quantity under which the truth is
optimal. \emph{Minimal selection} provides it. A sample contains
few constituents, each present or absent, and the occurrence
matrix $\delta \in \{0,1\}^{M \times N}$ ($\delta_{mj} = 1$ when
component $j$ serves sample $m$) records exactly this; among all
decompositions that reproduce the data exactly, prefer the fewest
occurrences. This is the principle compressed sensing built on in
a fixed basis: the sparsest representation is identifiable under
the spark condition \citep[Corollary~1]{donohoelad2003optimally} and findable
from incomplete measurements
\citep{candes2006robust,donoho2006compressed}. gMCR poses the
same question for a learned pool across many samples, and part
(a) of the theorem below answers it: the minimal-selection
decomposition is unique and is the truth.

Noise precludes exact fits, so the principle is expressed as a
per-sample criterion: for a candidate selection $A \subseteq [N]$
and a sample $x$,
\begin{equation}
\ell_\lambda(A;\, x) \;=\; \min_{c \,\ge\, 0}\,
\Bigl\| x - \sum_{j \in A} c_j s_j \Bigr\|^2 + \lambda\, |A| ,
\label{eq:penobj}
\end{equation}
the best nonnegative fit plus a penalty $\lambda$ per selected
component. The penalty acts on the row of $\delta$, not on the
coefficient magnitudes: Eq.~\ref{eq:penobj} is best-subset
($\ell_0$-type) selection, not ridge or lasso
\citep{hoerl1970ridge,tibshirani1996regression}, which penalize
$\|c\|$ and impose no cost on the selection count
(Appendix~\ref{app:thm1}).

\begin{theorem}[Identifiability at minimal usage]
\label{thm:ident}
Assume the gMCR model (Eq.~\ref{eq:model}) with the spark
condition (Definition~\ref{def:spark}).
\emph{(a)} For noiseless data, the true occurrence matrix
$\delta^{\star}$ is the unique minimizer of total usage
$\sum_{m,j} \delta_{mj}$ among all exact factorizations over the
pool.
\emph{(b)} Set $L_\alpha = \sqrt{2 \log(6NM/\alpha)}$. If
$c_{\min} s_{\min} > 2 \sigma L_\alpha$ and
$\sigma^2 L_\alpha^2 < \lambda <
(c_{\min} s_{\min} - \sigma L_\alpha)^2$, then with probability at
least $1 - \alpha$, simultaneously for every sample, the true
support is a strict local minimizer of
$\ell_\lambda(\,\cdot\,;\, x^{(m)})$ under single-component
changes; at $\sigma = 0$ it is the unique global minimizer among
selections with $|A| \le k$.
\end{theorem}

Theorem~\ref{thm:ident} indicates that an added component can
improve the fit only by absorbing noise, a gain below $\lambda$,
while removing a true component increases the fit error by more
than $\lambda$. Between these two bounds, the true support is the
strict optimum, sample by sample. The formal statement and proof
are given in Appendix~\ref{app:thm1}.

\subsection{The Structure of the Solution Set}
\label{sec:theory-cert}
\label{sec:theory-frontier}

In the criterion of Theorem~\ref{thm:ident}, selection is a
variable. Which components serve which samples, and how many
components appear in total, are decided by the search, not fixed
in advance. This is necessary: a component count fixed at any
value other than the unknown true count $K^{\ast}$ leaves the
true decomposition outside the search space. Selection therefore
has structure of its own, and that structure is what the results
of this section describe. The problem changes accordingly. The
goal is not to fit a single decomposition. It is to describe the
set of decompositions the data admit, and to locate a returned
solution within that set.

Three facts give the basic description of this set. The data
cannot see the order of the pool slots: renumbering the slots,
and renumbering the columns of $\delta$ to match, produces
exactly the same samples, so components can be recovered at best
up to relabeling. When the supports rarely distinguish one
component from another, the data admit a continuum of
decompositions, and Appendix~\ref{app:toy} computes one; this
ambiguity belongs to the data, not to any method. When
occurrences are diverse, the set collapses to the truth up to
relabeling.

Verification is a check on whatever the solver returns:
\begin{equation}
\Bigl\| x^{(m)} - \sum_{j \in A_m} \hat c^{(m)}_j s_j \Bigr\|
\le \varepsilon_{\mathrm{fit}}
\ \text{ and } \ |A_m| \le k \ \text{ for every } m,
\qquad
\sum_m |A_m| \;\le\; \sum_m |J_m| .
\label{eq:cert}
\end{equation}
The first is a condition on fit, the second on usage. Together
they form a \emph{certificate}: a solution that passes them is
proved to be the truth, and the component count is read off that
solution.

\begin{theorem}[Certified recovery; informal]
\label{thm:cert}
\emph{(a)} Over a known pool, every solution passing
(\ref{eq:cert}) is the truth: $A_m = J_m$ for every sample,
hence $|\bigcup_m A_m| = K^{\ast}$, with concentrations correct
to the accuracy the noise permits.
\emph{(b)} Over a learned pool without noise, every solution
passing the learned-pool check is the truth up to a relabeling
$\pi$ of the pool slots: $\hat s_{\pi(i)} = s_i$,
$A_m = \pi(J_m)$, $\hat c^{(m)}_{\pi(i)} = c^{(m)}_i$; the count
again equals $K^{\ast}$.
\emph{(c)} Under noise the same holds up to Conjecture~R, stated
below; one component per sample needs no conjecture.
\end{theorem}

\noindent
The formal statements (Appendix~\ref{app:thm2}) add the precise
noise, probability, and regularity conditions; they sharpen this
picture without changing it.

Part (b) of Theorem~\ref{thm:cert} rests on a pinning step: a
pair of samples sharing exactly one component pins that component
to one pool slot. Under noise the step needs more: the two
samples must occur in proportions different enough that no
coordinated error imitates the component across both. Exactly
this is the open statement: fix
$\gamma_{\mathrm{c}} > 0$ and say a sample $m$ is
\emph{diversely served} if some $m_1, \dots, m_{|J_m|}$ among the
samples sharing its support and candidate set satisfy
\[
\sigma_{\min}\Bigl( \bigl[\, c^{(m_1)} \ \cdots \
c^{(m_{|J_m|})} \,\bigr] \Bigr) \;\ge\; \gamma_{\mathrm{c}} .
\]

\begin{conjectureR}[isolating pairs; informal]
For some $\gamma_{\mathrm{c}} > 0$: every active component $i$
admits samples $m, m'$ with $J_m \cap J_{m'} = \{ i \}$, both
diversely served at level $\gamma_{\mathrm{c}}$.
\end{conjectureR}

The case of one component per sample ($k = 1$) is proved, and
the noiseless limit needs no conjecture. Moreover, whether the
required pairs exist can be checked directly on a returned
solution, by locating them among its supports and concentrations;
whenever they are present, part (c) holds for that solution with
nothing assumed. The formal version is given in
Appendix~\ref{app:thm2}.

\begin{corollary}[Reuse; informal]
\label{cor:reuse}
Let the pool be the components recovered by part (b), retained
after training, and let
$x = \sum_{i \in J} c_i s_i + \varepsilon$ be a fresh sample of
the same process, with $J$ among the recovered components. Then
the check of (\ref{eq:cert}), applied to $x$ alone, again returns
the truth: $A = J$, with concentrations correct to the accuracy
the noise permits.
\end{corollary}

\noindent
This is what it means to learn the process rather than fit one
dataset. The recovered pool serves as a component library. A new
sample is decoded by itself, under the same check, with nothing
re-trained.

A caution is in order. The guarantee covers any solution that
passes the check, no matter how the solution was found. But not
every condition of the check can be verified in practice. The fit
and the isolating pairs can be read off the data and the returned
solution. The usage condition is different. It asks that the solution use
no more selections than the truth, and the truth is exactly what
is unknown, so no computation on the data and the solution can
evaluate it. What the theory offers instead is
Theorem~\ref{thm:ident}: inside the penalty window, the criterion
itself favors minimal usage, so any solver that minimizes it is
pushed toward a condition it cannot check. A solution meeting
every condition is certified, and the truth itself meets them;
but whether a returned solution has met this one cannot be
observed. This is all that can be said at a fixed number of
samples. As samples accumulate, the two targets
of the inverse problem diverge: decoding every sample drifts out
of reach, while the process itself comes within it. The next
theorem states both.

\begin{theorem}[Impossibility and consistency; informal]
\label{thm:limits}
\emph{(a)} As the number of samples $M$ grows, no estimator
decodes every sample correctly: the probability that all $M$
decodings are right decays geometrically in $M$.
\emph{(b)} With a known noise law and diverse occurrences, the
estimated process converges almost surely, up to relabeling, to
the true process, and the estimated count is eventually exactly
$K^{\ast}$.
\end{theorem}

\noindent
The formal statements (Appendix~\ref{app:thm3}) add the precise
assumptions, the constant noise level of a static process among
them; they sharpen this picture without changing it. The reason
for (a) is elementary: the largest noise among $M$ independent
draws grows like $\sqrt{2 \log M}$, so some sample is eventually
overwhelmed, no matter the estimator. Behind (b), among processes
reproducing the limiting data law, the usage-minimal one is the
truth, and compactness converts identification into convergence.

Read together, the theorems tell a user what to expect.
Per-sample decodings are trustworthy at the scale of the dataset
in hand, where Theorem~\ref{thm:cert} certifies them. What
improves as data accumulate is the process, and above all the
component count, the quantity to trust most. Failure has known
addresses, not a general shadow. With too little noise margin,
the window of Theorem~\ref{thm:ident} closes and over-selection
follows. With too little occurrence diversity, isolating pairs
disappear and components merge. With an unknown noise law, the
concentration \emph{law} is out of reach, while supports,
profiles, and the count are not. Away from these conditions, the
inverse problem is not merely solvable: a returned solution can
be checked.

\section{EB-gMCR: A Solver for the Generative Model}
\label{sec:solver}

EB-gMCR is the solver that turns the model of \S\ref{sec:gmcr}
and the results of \S\ref{sec:theory} into an engineering
solution: an energy-based model that infers selections,
component profiles, and concentrations from the observed samples
alone, with each design element stating, where it is built, the
result it serves. \S\ref{sec:solver-ebselect}
builds the selection gate; \S\ref{sec:solver-objective} the
objective; \S\ref{sec:solver-checkpoint} the checkpointing
rule; and \S\ref{sec:solver-convergence} analyzes when a run's
checkpoints can be trusted.

\subsection{EB-select: Hard Gating for Discrete Selection}
\label{sec:solver-ebselect}

Theorem~\ref{thm:ident} requires discrete per-sample selection:
continuous shrinkage cannot collapse the factorization ambiguity
to a finite relabeling group, and only a selection that is
exactly on or exactly off lets a support be exactly right or
exactly wrong, which is what makes the certified exact recovery
of Theorem~\ref{thm:cert} a meaningful claim. EB-select is the gate that realizes this requirement: zeros in $\delta$ encode non-selection, not near-zero amplitudes; the gate is soft during training so that gradients flow, and hard at evaluation.

For each observed sample $x$, a neural network evaluates
selection energies $E(x) \in \R^{N \times 2}$, one energy for
selecting component $j$ ($E_{j,1}$) and one for not selecting
it ($E_{j,0}$). Learning a discrete selection by gradient is
the obstacle; the Gumbel-softmax reparameterization
\citep{jang2016gumbel,maddison2016concrete} is the alternative.
With $Z_{j,\ell} \sim \mathrm{Gumbel}(0,1)$ i.i.d.,
\begin{equation}
[f_e(x)]_j
\;=\;
\frac{\exp\bigl( ( Z_{j,1} - E_{j,1} ) / \tau \bigr)}
{\sum_{\ell \in \{0,1\}} \exp\bigl( ( Z_{j,\ell} - E_{j,\ell} ) / \tau \bigr)}
\;=\;
\varsigma\Bigl( \frac{ ( E_{j,0} - E_{j,1} )
+ ( Z_{j,1} - Z_{j,0} ) }{\tau} \Bigr),
\label{eq:gumbel}
\end{equation}
the logistic function $\varsigma$ of the noisy energy gap at
temperature $\tau$, so that $f_e(x) \approx \mathbb{E}[\delta \mid x]$ is
differentiable in the energies. Training uses a positive
$\tau$; evaluation uses a near-zero temperature and selects
component $j$ when its probability exceeds a fixed threshold,
equivalently when the energy gap $E_{j,0} - E_{j,1}$ exceeds a
floor. The threshold, the floor it induces through
Eq.~\ref{eq:gumbel}, and the gate's decision surface are reported in
Appendix~\ref{app:gate}. To improve
exploration, the energies are perturbed during training by
additive Gaussian noise before they enter the gate:
\begin{equation}
E' \;=\; E + s_{\mathrm{ex}}\, \xi, \qquad \xi \sim \mathcal{N}(0, I),
\label{eq:5}
\end{equation}
with a fixed scale $s_{\mathrm{ex}}$ (Appendix~\ref{app:repro}),
applied in training only. The energies are the model's own state costs, so
noise there perturbs the selection distribution directly,
without moving the parameters; the step is the diffusion part
of a Langevin iteration with the drift omitted.

\subsection{Objective: Fidelity, Parsimony, Distinctness}
\label{sec:solver-objective}

The gate supplies the selection; two further functions complete
the generative model. Profiles are learnable vectors $\mathbf{s}_i \in \R^d$, one per pool slot, normalized to unit norm and retrieved for the selected components; concentrations come from a network
$f_c: \R^d \to \R^N$ evaluated on the sample. The reconstruction
$\hat x$ is the superposition of the selected profiles weighted
by their concentrations, the model of Eq.~\ref{eq:model} with
the latent objects replaced by their learned counterparts.

The training loss is the differentiable form of the criterion of
Eq.~\ref{eq:penobj}, fit plus a penalty on selection, with two
regularizers:
\begin{equation}
\mathcal{L} \;=\; \| x - \hat x \|^2 + \lambda\, C(x)
+ \lambda_{\mathrm{se}} \| E \|_2^2
+ \lambda_{\mathrm{amb}}\, \mathcal{R}(B),
\qquad
C(x) \;=\; \frac{1}{N} \sum_{i=1}^{N} [f_e(x)]_i .
\label{eq:9}
\end{equation}
The usage term $C(x)$ is the selection count in soft form: as
$\tau \to 0$ it tends to the number of selected components
divided by $N$, so $\lambda\, C(x)$ is the count penalty $\lambda |A|$ of Eq.~\ref{eq:penobj} scaled by $1/N$, with $\delta$ soft during training. Its weight is
scheduled rather than fixed: $\lambda$ stays at zero until the normalized reconstruction error falls below a threshold (Appendix~\ref{app:repro}), then increases,
moving the optimization from reconstructing to reconstructing
with the fewest components. The order matches Theorem~\ref{thm:ident}: the $\lambda$-window, inside which the penalized fit strictly prefers the true selection, is defined for a fit already established. Whether a given run\'s $\lambda$ enters the window is not fixed by the schedule and is not claimed; the certificate judges the outcome.

The ambiguity term keeps the learned pool in the regime the
theory assumes. Theorems~\ref{thm:ident}--\ref{thm:cert}
identify components up to the relabeling group only under the
spark condition, and a pool with duplicated or nearly parallel
profiles violates it: two slots carrying the same pattern can
trade concentration freely. $\mathcal{R}$ penalizes similarity among the profiles in use in a minibatch $B$,
\begin{equation}
\mathcal{R}(B) \;=\;
\frac{1}{|\mathcal{P}_B|}
\sum_{\{j, k\} \in \mathcal{P}_B}
K(\mathbf{s}_j, \mathbf{s}_k),
\label{eq:7}
\end{equation}
with $K$ a radial basis function (RBF) kernel, $\mathcal{I}_B$ the components selected for at least one sample of $B$, and $\mathcal{P}_B$ the pairs in $\mathcal{I}_B$: the Hilbert--Schmidt independence criterion \citep{gretton2005kernel,gretton2005hsic} restricted to the components in use. The restriction is what keeps the term cheap and its weight small: duplication matters only among components that serve the same data, so a weight just large enough to break ties suffices, and the fit term is left to shape the profiles. The
term promotes the spark condition; it does not certify it. The
energy regularizer $\lambda_{\mathrm{se}} \| E \|_2^2$ removes
the free common shift of each energy pair
(Appendix~\ref{app:gate}). All weights are fixed across every
experiment (Appendix~\ref{app:repro}).

\subsection{Checkpointing: The Certificate}
\label{sec:solver-checkpoint}

The theorems say what a good answer is, not how to reach it.
Theorem~\ref{thm:ident} puts the true selection at minimal usage
inside its window; Theorem~\ref{thm:cert} says that any answer
which fits and uses no more than the truth is the truth; and a
run under persistent noise has no fixed point to wait for. A
program follows by reading these as instructions. Because the
endpoint is a checked object rather than a converged one, the
solver returns checkpoints: states sampled during training and
judged on observable criteria. Because the truth sits at minimal
usage, the checkpoint kept is the one that reconstructs the data
with the fewest components; the loss already pushes usage down
sample by sample, and the checkpoint rule keeps the
least-component state the run has reached. Because the
certificate's fit tolerance depends on the noise level, and the
noise of a real instrument varies, the reconstruction quality
that defines a checkpoint band is left to the user, stated as an
$R^2$ band.

The user's statement is a convenience, not a requirement. One
run yields checkpoints over several bands and the certificate
judges each, so the mechanism does not need the noise level to
be known; knowing it selects the band where the theorems'
window lies. A band the data cannot support leaves the run
outside that region: too strict a band is never entered and
returns no checkpoint, too loose a band accepts states that have
dropped true components. Of the certificate's two conditions,
fit is computed at every checkpoint, while usage compares
against the unknown truth and cannot be certified during the
run; Theorem~\ref{thm:cert} guarantees that the truth itself
passes, so a correct solution exists to be found. Non-negative
concentrations and unit-norm profiles hold by construction, so
no projection separates a checkpoint from the certificate. The
rules, the bands used, and termination are given in
Appendix~\ref{app:repro}; when a run's checkpoints are stable
enough to trust is the question of
\S\ref{sec:solver-convergence}.

\subsection{Convergence: Stability After the Anchor}
\label{sec:solver-convergence}

We analyze training under the checkpointing rule: when does a
run stop producing transients, so that a checkpoint records a
stable selection rather than a passing one? Call the moment
every accept and every reject of the gate holds a definite
margin, an event computable from the gate's energies, an \emph{anchor}. From an observed anchor, training
enters a certified two-phase regime.

\begin{proposition}[Two-phase convergence; informal]
\label{prop:dyn}
Let $t_0$ be an observed anchor with margin $\gamma > 0$: a
time at which every selection margin of the gate is at least
$\gamma$. With $L_\Delta$ the margin Lipschitz constant, $G$
the update-displacement constant, $\eta$ the constant step size
after $t_0$, $\hat J_t$ the support the gate selects at step
$t$, and $J_{\sharp} = \hat J_{t_0}$, assume the
local-geometry, concentration, and compatibility conditions of
Appendix~\ref{app:dyn} (curvature constant $\mu$, noise floor
$\Phi(\eta)$, noise-floor level $p \le \tfrac12$), with
expectations and probabilities conditional on the run's log at
the anchor. Then:
\emph{(a)} deterministically, $\hat J_t = J_{\sharp}$ for all
$t \in [\,t_0,\; t_0 + T(\gamma)\,]$, where
\[
T(\gamma)
\;=\;
\Bigl\lfloor \frac{\gamma}{4\, L_\Delta\, G\, \eta}
\Bigr\rfloor ;
\]
\emph{(b)} with the selection frozen, the objective $f$
contracts in expectation: for $t$ in the window,
\[
\mathbb{E}\bigl[ f( \theta_t ) - f^{\star} \bigr]
\;\le\;
\Bigl( 1 - \frac{\mu\eta}{2} \Bigr)^{t - t_0}
\bigl( f( \theta_{t_0} ) - f^{\star} \bigr) + \Phi(\eta),
\]
and the contraction completes inside the window;
\emph{(c)} on the concentration event over the sample, after the
contraction, the anchored selection persists: over any horizon
of $T$ further steps,
\[
\Pr\Bigl( \hat J_t = J_{\sharp}
\text{ for every } t \text{ in the horizon} \Bigr)
\;\ge\;
1 - \zeta(T),
\qquad
\zeta(T) \;=\; \frac12 + p + p\, \frac{\mu\eta}{2}\, T ,
\]
and the corresponding single-time failure bound decays
geometrically to $p$.
\end{proposition}

The formal statement and proof are given in
Appendix~\ref{app:dyn}. The analysis begins at the observed
anchor: a rate from initialization is not claimed, and no
conclusion depends on the run before the anchor. The three
parts are one mechanism. The
window of (a) is arithmetic, Lipschitz continuity times bounded
optimizer steps, and holds at any sample size; part (b) is
linear convergence, in the optimization sense, to a noise
floor; part (c) makes the stability self-sustaining where the
geometry and concentration conditions hold, and at the
implemented scale the concentration conditions exceed the
experiments' $M$, so (a) and (b) are the conclusions that bind
there (Appendix~\ref{app:dyn} quantifies the boundary). What
the mechanism gives the solver is the meaning of its
checkpoints: every checkpoint taken inside the window records
the same selection, so the rolling checkpoint of
\S\ref{sec:solver-checkpoint} samples a stable, checkable
object, and the certificate of Theorem~\ref{thm:cert} alone
decides whether that object is the truth.

The statement's two-phase shape follows the finite-identification
literature, where proximal methods provably reach the correct
support and then converge linearly on it
\citep{liang2017activity,lee2012manifold}. Its trigger
follows the certificate tradition of safe screening, where an
observable quantity yields a certified conclusion
\citep{ndiaye2017gap}. Three elements are new, each with a
consequence for the MCR task. The trigger is an event the
practitioner observes, not an existence claim about the run.
The guarantee therefore attaches to a real unmixing run at a
moment its log identifies, with no unverifiable condition on
how training began. The window protects the component set the
solver actually selected, correct or not. Stability and truth
are therefore separate verdicts: stability from this analysis,
truth from the certificate of Theorem~\ref{thm:cert}. The
persistence bound treats the whole horizon as one event. A
single number therefore bounds the risk that the selected set
drifts while the run continues past a checkpoint.

\section{Experiments}
\label{sec:exp}

This section measures whether the solver of \S\ref{sec:solver}
reaches, on data, what \S\ref{sec:gmcr} and \S\ref{sec:theory}
prove: the component count, the fit the certificate asks for, and
the reuse of a recovered process. Two quantities are reported
throughout. The reconstruction $R^2$ summarizes the per-sample
fit condition of the certificate over a dataset. The estimated
component count (EC) is the number of components a solver
returns, the quantity of Theorems~\ref{thm:cert} and
\ref{thm:limits}(b). Every baseline except ARD-NMF takes the
count as input; ARD-NMF chooses it through a noise-scale
hyperparameter $\phi$. Setups, exact numbers, and runtimes are
in Appendices~\ref{app:c}--\ref{app:f}; EB-gMCR is run five
times per configuration, the baselines one hundred.

\subsection{Synthetic Mixtures}
\label{sec:exp-synth}

\begin{figure}[t]
\centering
\includegraphics[width=\textwidth]{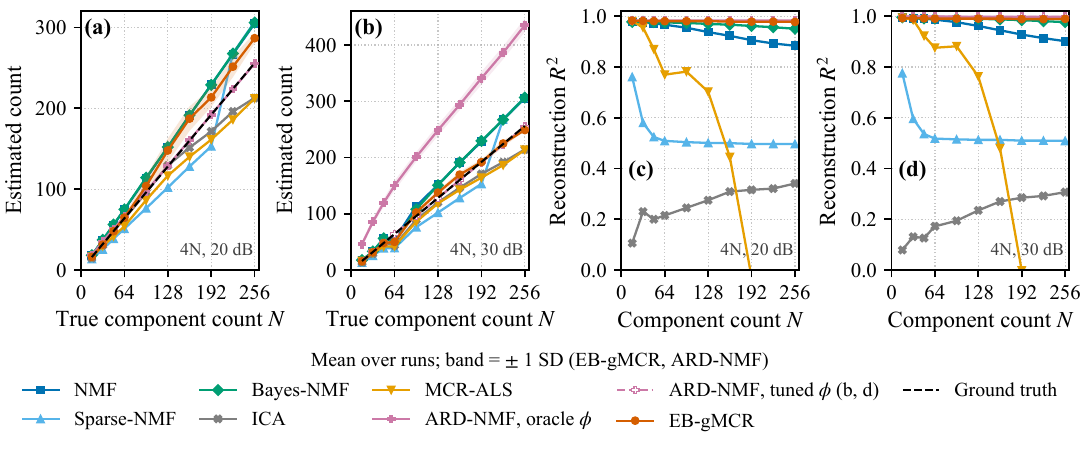}
\caption{Synthetic mixtures, $4N$ samples. Estimated against true
component count at (a) 20\,dB and (b) 30\,dB, and reconstruction
$R^2$ against the true count at (c) 20\,dB and (d) 30\,dB.}
\label{fig:count}
\end{figure}

The synthetic mixtures (Appendix~\ref{app:c}) have known
components, which lets the returned count be checked directly: unit-norm
non-negative profiles in $\R^{512}$, one to four active components
per sample, $N \in \{16, \dots, 256\}$ true components, $4N$ or
$8N$ samples, at 20 and 30\,dB; EB-gMCR starts from a pool of
$1024$ candidate components. Its count is read from one $R^2$
band per setting, the success band of the baseline search
(Appendix~\ref{app:d}); Table~\ref{tab:e1} gives every band.
Figure~\ref{fig:count}(a, b) and Tables~\ref{tab:e2}
and \ref{tab:e2b} report the counts; Figure~\ref{fig:count}(c, d)
the reconstruction quality.

At 30\,dB the count is within a few percent of the truth at every $N$ and both sample sizes, within $4\%$ with $8N$ samples, above the truth for $N$ between $48$ and $160$ and below it at the smallest and largest $N$ (Table~\ref{tab:e1}); no fixed-count method offers this. At
20\,dB the count exceeds the truth, by $8$ to $17\%$ for
$N \ge 96$ at $4N$ and by up to $26\%$ at $8N$, and doubling the
samples does not remove the excess. The window condition
$c_{\min} s_{\min} > 2 \sigma L_\alpha$ of
Theorem~\ref{thm:ident} gives one reading of the two noise
levels: at the edge of the window, extra components fit noise
and no usage weight can forbid them, so the count can only
exceed the truth there. Whether either noise level lies inside
the window is not computed here. Runs at $8N$ end sooner
(Table~\ref{tab:runtime}); Proposition~\ref{prop:dyn} gives no
rate in the sample count, and the shorter time is measured, not
given by the theory. The per-sample supports behind these counts
are taken up in \S\ref{sec:limits}.

The classical solvers take the count as input; for the
comparison each is given a search over counts near the truth
(Appendix~\ref{app:d}). Even with the search they leave the
truth as the count grows, and every one of them has left the
fit band or the true count by $N = 256$ (Fig.~\ref{fig:count}(a,
b), Tables~\ref{tab:e2} and \ref{tab:e2b}). ARD-NMF determines
the count on its own and tracks the truth at 20\,dB when its
noise-scale hyperparameter $\phi$ is set to the true noise
variance; the same setting returns $1.7$ to $2.9$ times the
truth at 30\,dB while reporting $R^2 \approx 0.998$, and a
$\phi$ chosen with the true count in view returns the exact
count (Appendix~\ref{app:e-ard}, Fig.~\ref{fig:ardnmf-synth}). Both
solvers take the noise level as input, ARD-NMF as the variance
scale $\phi$, EB-gMCR as the $R^2$ band. The band is a
reconstruction quality that every checkpoint is checked against,
and one run returns a checkpoint for every band; a value of
$\phi$ cannot be judged from the run itself. Runtimes are in
Table~\ref{tab:runtime} (Appendix~\ref{app:e}).

\subsection{Real Chemical Spectra}
\label{sec:exp-real}

\begin{figure}[t]
\centering
\includegraphics[width=\textwidth]{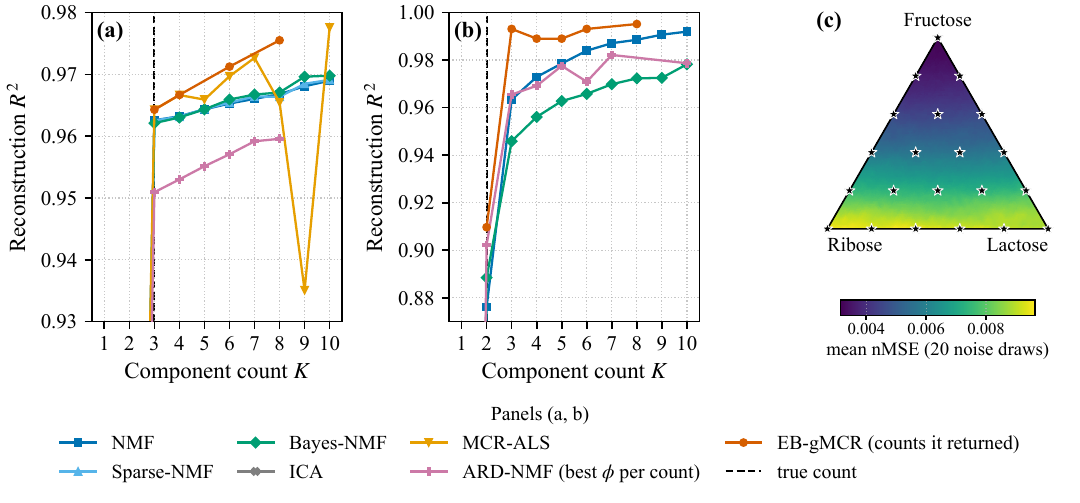}
\caption{Real chemical spectra. Reconstruction $R^2$ against
component count on (a) Carbs and (b) NIR; the black dashed line
is the true component count of the dataset. (c) Frozen-model
reuse on the Carbs simplex: mean nMSE at $861$ mixtures, the
$21$ training mixtures starred.}
\label{fig:real}
\end{figure}

Two public spectroscopy datasets are used. \textbf{Carbs}
\citep{Engelsen_SPECARB_Carbohydrates} is Raman: $M = 21$
mixtures of three carbohydrates. \textbf{NIR}
\citep{Liebmann_et_al_NIRdataset_2009} is first-derivative
near-infrared: $M = 166$ samples with two analytes. EB-gMCR
starts from a pool of $128$ candidates. Metrics at the true
count are in Table~\ref{tab:f1} (Appendix~\ref{app:f}) and
Figure~\ref{fig:real}(a, b).

At the end of each run the EC matches the true count, three on
Carbs and two on NIR, with no count given to the solver. On NIR
the end state sits below the lowest checkpoint band, since no
method exceeds $R^2$ of about $0.91$ there with two components
(Table~\ref{tab:f1}). Stricter $R^2$ bands of the same run
return more components, as a tighter fit tolerance must. At the
true count, EB-gMCR reconstructs as well as the best method that
was given the count (Fig.~\ref{fig:real}(a, b)), and on Carbs
its concentrations and profiles match those of MCR-ALS, the
reference method in chemometrics (Table~\ref{tab:f1}). ARD-NMF,
the one baseline that also chooses its count, decomposes equally
well at the true count and reaches that count for $\phi$ within
a range no single run reveals (Appendix~\ref{app:f},
Fig.~\ref{fig:ardnmf}).

What the run learns is the process that generated the mixtures,
and Corollary~\ref{cor:reuse} says a model holding that process
decodes mixtures it has never seen. The least-component Carbs
checkpoint, selected with no use of the known count, was frozen
and decoded $861$ mixtures synthesized from the reference
profiles across the concentration simplex, with fresh noise
(Appendix~\ref{app:b.5}). The model kept its three components
on every mixture, and the error tracks the noise floor, the nMSE
of the added noise alone, over the whole simplex and not only
near the $21$ training mixtures (Fig.~\ref{fig:real}(c);
Appendix~\ref{app:f}, Fig.~\ref{fig:floor}). A fixed-count fit
of the $21$ training mixtures decodes no new mixture; the frozen
model decodes all $861$.

\subsection{Ablation}
\label{sec:exp-ablation}

\begin{table}[t]
\centering
\caption{Ablation at $N = 256$: one term of the loss removed at a time. Mean $\pm$ one standard deviation (SD) over five runs, counts read from the band of Table~\ref{tab:e2}; Time in hours; runs capped at $10^5$ epochs.}
\label{tab:ablation}
\footnotesize
\setlength{\tabcolsep}{4pt}
\begin{tabular}{lccc|ccc}
\toprule
\multirow{2}{*}{\textbf{Solver}}
& \multicolumn{3}{c|}{\textbf{20dB}}
& \multicolumn{3}{c}{\textbf{30dB}} \\
\cmidrule(lr){2-4} \cmidrule(lr){5-7}
& $R^2$ & EC & Time & $R^2$ & EC & Time \\
\midrule
\multicolumn{7}{l}{\textit{$4N$ samples}} \\
\midrule
EB-gMCR & 0.978±0.003 & 286.7±6.4 & 3.6±0.8 & 0.990±0.000 & 248.8±1.1 & 3.3±0.6 \\
w/o exploration noise & 0.980±0.001 & 292.0±8.2 & 4.7±0.3 & 0.990±0.000 & 257.8±6.2 & 2.9±0.3 \\
w/o usage penalty $\lambda C(x)$ & 0.980±0.001 & 284.6±13.4 & 4.1±0.3 & 0.995±0.000 & 268.2±4.5 & 4.1±0.3 \\
w/o energy regularizer $\|E\|_2^2$ & 0.979±0.002 & 282.0±20.3 & 6.1±0.2 & 0.990±0.000 & 251.8±3.4 & 6.0±0.4 \\
w/o ambiguity term $\mathcal{R}(B)$ & 0.980±0.000 & 292.2±13.0 & 3.8±0.6 & 0.990±0.000 & 249.0±1.6 & 4.5±0.7 \\
\midrule
\multicolumn{7}{l}{\textit{$8N$ samples}} \\
\midrule
EB-gMCR & 0.975±0.000 & 301.0±6.4 & 2.0±0.2 & 0.991±0.001 & 255.0±1.4 & 2.1±0.2 \\
w/o exploration noise & 0.976±0.002 & 323.8±41.2 & 2.5±0.3 & 0.991±0.000 & 255.8±0.4 & 2.0±0.3 \\
w/o usage penalty $\lambda C(x)$ & 0.975±0.000 & 318.6±32.2 & 2.2±0.1 & 0.993±0.002 & 298.2±29.3 & 2.4±0.2 \\
w/o energy regularizer $\|E\|_2^2$ & 0.975±0.000 & 290.4±11.9 & 12.1±0.1 & 0.991±0.000 & 255.6±0.5 & 12.2±0.1 \\
w/o ambiguity term $\mathcal{R}(B)$ & 0.975±0.002 & 305.4±6.4 & 2.5±0.5 & 0.991±0.000 & 254.8±0.8 & 2.2±0.1 \\
\bottomrule
\end{tabular}%
\end{table}

Removing one loss term at a time at $N = 256$
(Table~\ref{tab:ablation}) measures what each term
of \S\ref{sec:solver} does. Without the usage penalty the count
inflates in three of the four settings, by up to $17\%$ at $8N$
and 30\,dB: nothing then prefers the minimal selection. Without
the energy regularizer the termination rule never fires, the run
ends at the epoch cap of the ablation, and runtime grows by
$1.7$ to $6$ times: the stopping signal is gone. Without the
exploration noise the count moves by a few percent, within one
standard deviation except at $4N$ and 30\,dB. Without the ambiguity term the count and the fit stay within
one standard deviation of the full solver. The usage penalty
already prefers one whole profile per pattern over several
partial ones; the ambiguity term acts on the remaining case, two
slots carrying the same profile and used by different samples,
which the count penalty does not distinguish from one slot.

\section{Discussion}
\label{sec:discussion}
\label{sec:limits}
\label{sec:conclusion}

\paragraph{What modeling the process established.}
This paper models the generative process rather than the data
matrix. That choice changes the object of inference and what a
returned decomposition can claim. Theorem~\ref{thm:ident} makes
minimal usage the objective, Theorem~\ref{thm:cert} makes the
certificate the check, and Theorem~\ref{thm:limits} states what
more samples do and do not resolve; Corollary~\ref{cor:reuse}
makes the recovered pool reusable on new samples of the same
process. A certified solution is the truth up to a relabeling,
its component count is the true component count, and its concentrations are the true concentrations of the components. A solution at any other component count is not
the truth, and its concentrations are parameters of a fit. Every method that discovers the component count rests on a statement about the noise, and the certificate makes that statement checkable on the returned decomposition. What the certificate cannot verify on the returned decomposition is whether its per-sample supports are usage-minimal.

\paragraph{What the solver reaches.}
Decoding the synthetic runs sample by sample
(Appendix~\ref{app:e-support}) shows two states of one run, and
the theory names both. Early in a run the gate selects almost as
few components per sample as the truth, but the pool is not yet
complete: some true components have no slot of their own. Late
in a run the pool is complete and each true profile is pinned to
a slot, but the gate has opened for many more components per
sample than any sample contains, while the concentrations it
assigns to the extra components are near zero. The component
count read off the pool is right in the late state; the
per-sample selection is not. The theory places this. A usage
penalty below the $\lambda$-window of Theorem~\ref{thm:ident}
keeps components that fit noise and drops none, which is the
late state exactly, and the solver's schedule does not put the
penalty inside the window. The certificate does not read a
support from the gate alone but from the coefficients that clear
its floor, condition (C8), and under that reading the late state
is close to usage-minimal. What no retained checkpoint has is a
complete pool and a usage-minimal gate at the same time. For a
chemical study this is the difference between two questions.
The late state answers which components the study contains and reads their concentrations; it answers which components a given sample contains through those concentrations, not through the selection. What remains
to be built is a solver whose gate stays usage-minimal until the
pool is complete. Theorem~\ref{thm:ident} says where the penalty
must sit for that, and the certificate says whether a candidate
did it.

\paragraph{Scope and open questions.}
The generative process fixes each component profile across
samples; profiles that shift between samples, as in
chromatographic drift, are outside the model. On Carbs the
candidate pool has more slots than there are samples, and the
certificate there rests on assumption (C7) of
Appendix~\ref{app:thm2}, which is not checked
(Appendix~\ref{app:f}).

One limit belongs to the data rather than to any solver, and it
has a practical form. The certificate pins a component to a slot only when two samples share exactly that component. If two components occur together in every sample that contains either, their joint contribution admits a continuum of decompositions: any pair of profiles spanning the same plane, with the concentrations transformed to match, reproduces every sample at the same usage. Proportions that vary across samples narrow the continuum, since every transformed concentration must stay admissible, but do not close it; proportions that are fixed make the pair indistinguishable from one component, which minimal usage then prefers (Appendix~\ref{app:toy}). No method separates such a pair from those samples, and at the sample counts used here most components lack an isolating pair even under the true supports (Appendix~\ref{app:e-support}). The condition
is one a study can design for, by including samples in which a component occurs alone or in different proportions, and the
certificate says when the design has succeeded.

With a pool larger than the true component count, minimal usage leaves the extra
slots unused (Theorem~\ref{thm:ident}). With a pool smaller than
the true component count, a solver must merge some components or
drop them, and the theorems do not say which. What such a solver
returns is the open question this paper points to. The same
question arises in the sparse autoencoders used for mechanistic
interpretability, which decompose network activations into sparse linear combinations of learned features whose number is chosen by the practitioner, not known \citep{bricken2023monosemanticity, cunningham2024sparse}; the
premise there, that a network represents more features than it
has dimensions, is itself a superposition claim
\citep{elhage2022superposition}. That setting raises the three questions this paper asks of chemical mixtures: how many features there are, when the decomposition is unique, and what is returned when the autoencoder is given fewer features than the activations contain, the case of a pool smaller than the true component count. For linear superposition with a sparse selection per sample, this paper answers the first two, the second up to Conjecture~R under noise, and leaves the third open.

\bibliography{references}
\bibliographystyle{iclr2025_conference}

\newpage
\appendix

\section{Formal Statement and Proof of Proposition~\ref{prop:welldef}}
\label{app:prop1}

This appendix proves Proposition~\ref{prop:welldef} in full. The
proposition says: if any $k$-sparse combination with detectably
active coefficients reconstructs a true sparse sample with error
below the threshold $c_{\min} s_{\min}$, then it is the true
decomposition, in support and in coefficients alike. The proof
subtracts the two candidate representations and applies the spark
condition to their difference: Steps~1--5 bound every coordinate
of the difference strictly below $c_{\min}$; Step~6 turns that cap
into support equality (a support disagreement would force a
coordinate of magnitude at least $c_{\min}$); Step~7 reads off the
coefficient error.

\paragraph{Notation.}
\begin{center}
\begin{tabular}{@{}ll@{}}
\toprule
$\|x\|$ & Euclidean norm of $x \in \R^d$; $d$ is the feature
dimension\\
$S_A$ & the $d \times |A|$ matrix with the profiles $s_i$,
$i \in A$, as columns\\
$S_A v$ & $= \sum_{i \in A} v_i s_i$, with $v$ indexed by the
elements of $A$\\
$\sigma_{\min}(S_A)$ & $= \min_{\|u\| = 1} \|S_A u\|$, the
smallest singular value of $S_A$\\
$s_{\min},\ c_{\min}$ & assumed constants of the model: spark
margin, concentration floor\\
$\epsilon$ & deterministic reconstruction tolerance; distinct from
the model noise $\varepsilon$\\
\bottomrule
\end{tabular}
\end{center}
Two conventions. In the matrix notation of Eq.~\ref{eq:1},
profiles are \emph{rows} of $S$, whereas $S_A$ stacks the selected
profiles as \emph{columns}, so that $S_A v$ maps a coefficient
vector to a synthetic sample in $\R^d$ and
$\|S_A u\| \ge \sigma_{\min}(S_A)\, \|u\|$ for every $u$.
$\sigma_{\min}(\cdot)$ always denotes a quantity \emph{computed}
from a matrix, while $s_{\min}$ (like $c_{\min}$) is an
\emph{assumed} constant of the model; the spark condition asserts
that the computed quantity dominates the assumed one,
$\sigma_{\min}(S_A) \ge s_{\min}$.

\paragraph{Proposition~\ref{prop:welldef}, restated.}
Fix $k \in \mathbb{N}$ and suppose the unit-norm profiles
$s_1, \dots, s_N \in \R^d$ satisfy the spark condition with margin
$s_{\min} > 0$: $\sigma_{\min}(S_A) \ge s_{\min}$ for every
$A \subseteq [N]$ with $|A| \le 2k$. Consider supports
$J, J' \subseteq [N]$ with $|J|, |J'| \le k$ and coefficients
$c_i \ge c_{\min} > 0$ for $i \in J$ and $c'_i \ge c_{\min}$ for
$i \in J'$, and write $x_0 = \sum_{i \in J} c_i s_i$. If
\[
\Bigl\| x_0 - \sum_{i \in J'} c'_i s_i \Bigr\| \;\le\; \epsilon
\qquad \text{for some } 0 \le \epsilon < c_{\min}\, s_{\min},
\]
then $J' = J$ and $|c_i - c'_i| \le \epsilon / s_{\min}$ for every
$i \in J$.

\begin{proof}
\emph{Step 1 (difference vector).} Let $A = J \cup J'$ and define
$v \in \R^{|A|}$, indexed by $A$, coordinate-wise:
\[
v_i =
\begin{cases}
c_i, & i \in J \setminus J',\\
c_i - c'_i, & i \in J \cap J',\\
-\,c'_i, & i \in J' \setminus J.
\end{cases}
\]
This is the difference of the two coefficient vectors after each is
extended by zero to all of $A$.

\emph{Step 2 (residual identity).} By linearity of the matrix--vector
product and Step 1,
\[
S_A v \;=\; \sum_{i \in A} v_i s_i
\;=\; \sum_{i \in J} c_i s_i - \sum_{i \in J'} c'_i s_i
\;=\; x_0 - \sum_{i \in J'} c'_i s_i ,
\]
and therefore $\|S_A v\| \le \epsilon$ by the hypothesis.

\emph{Step 3 (the spark condition applies).} $|A| \le |J| + |J'|
\le 2k$, so the spark condition gives
$\sigma_{\min}(S_A) \ge s_{\min}$.

\emph{Step 4 (norm bound).} By the definition of
$\sigma_{\min}$ (Notation above) and Steps 2--3,
\[
s_{\min} \|v\| \;\le\; \sigma_{\min}(S_A)\, \|v\|
\;\le\; \|S_A v\| \;\le\; \epsilon,
\qquad \text{hence} \qquad
\|v\| \;\le\; \epsilon / s_{\min} .
\]

\emph{Step 5 (coordinate bound).} For every $i \in A$,
$|v_i| \le \|v\|$ (a coordinate of a vector is bounded by its
Euclidean norm), so by Step 4 and the hypothesis
$\epsilon < c_{\min} s_{\min}$,
\[
|v_i| \;\le\; \epsilon / s_{\min} \;<\; c_{\min} .
\]

\emph{Step 6 (support equality).} This step is the reason the gap
between $\epsilon$ and $c_{\min} s_{\min}$ exists: a support
disagreement forces some coordinate of $v$ to reach magnitude at
least $c_{\min}$, while Step 5 caps every coordinate strictly below
$c_{\min}$. Concretely, suppose
$i \in J \setminus J'$. Then $v_i = c_i \ge c_{\min}$ by Step 1 and
the hypothesis $c_i \ge c_{\min}$, contradicting Step 5. Hence
$J \setminus J' = \emptyset$. Symmetrically, $i \in J' \setminus J$
gives $|v_i| = c'_i \ge c_{\min}$, again contradicting Step 5, so
$J' \setminus J = \emptyset$. Therefore $J' = J$.

\emph{Step 7 (coefficient bound).} With $J' = J$, Step 1 gives
$v_i = c_i - c'_i$ for every $i \in J$, and Steps 4--5 give
$|c_i - c'_i| \le \|v\| \le \epsilon / s_{\min}$.
\end{proof}

\noindent
The proof uses neither the unit-norm normalization of the profiles
nor any upper bound $c_{\max}$ on the coefficients; the proposition
therefore holds verbatim without these assumptions.

\section{Formal Statement and Proof of Theorem~\ref{thm:ident}}
\label{app:thm1}

This appendix proves Theorem~\ref{thm:ident} in full. Part (a)
says that among all exact reproductions of noiseless data over the
pool, the true occurrence matrix uses strictly fewer selections
than any alternative. Part (b) says that under noise, for
$\lambda$ inside the explicit window, every true support beats
every selection differing from it in one component, by a
quantified gap, simultaneously across all samples with probability
$1 - \alpha$; without noise the optimality is global among
$k$-sparse selections. Part (a) is proved in three steps:
uniqueness of sparse representations, a row-by-row spending bound,
and summation. Part (b) runs in eight steps: a one-flip fit
identity (Step~1), the costs of the two flips (Steps~2--3), dual
vectors (Step~4), the noise event (Step~5), inactivity of the
nonnegativity constraint at the truth (Step~6), the window verdict
(Step~7), and the noiseless global case (Step~8).

\paragraph{Notation.}
The notation of Appendix~\ref{app:prop1} remains in force. New
objects:
\begin{center}
\begin{tabular}{@{}ll@{}}
\toprule
$X_0 = \{x_0^{(m)}\}_{m=1}^{M}$ & noiseless data; the subscript is
the digit zero (zero noise)\\
$x_0^{(m)}$ & $= \sum_{i \in J_m} c^{(m)}_i s_i$; true support
$J_m$, $|J_m| \le k$; $c^{(m)}_i \in [c_{\min}, c_{\max}]$\\
$x^{(m)}$ & $= x_0^{(m)} + \varepsilon^{(m)}$, the observed data
(only these are observed)\\
$A_m$ & $= \{ j : \delta_{mj} = 1 \}$, the row supports of
$\delta \in \{0,1\}^{M \times N}$\\
$P_V$, $\dist(x, V)$ & orthogonal projection onto the subspace
$V$; $\dist(x, V) = \|x - P_V x\|$\\
$\ell_\lambda(A;\, x)$ & $= \min_{c \ge 0}
\| x - \sum_{j \in A} c_j s_j \|^2 + \lambda |A|$
(Eq.~\ref{eq:penobj})\\
\bottomrule
\end{tabular}
\end{center}
A matrix $\delta$ with row supports $A_m$ \emph{admits an exact
factorization} of $X_0$ over the pool if for every $m$ there are
real coefficients (of any sign) supported on $A_m$ with
$x_0^{(m)} = \sum_{j \in A_m} c'_j s_j$. The noise vectors
$\varepsilon^{(m)}$ are independent, mean-zero, and sub-Gaussian
with proxy $\sigma$
\citep[cf.\ \S\S 2.5 and 3.4]{vershynin2018high}: for every fixed
$a \in \R^d$ and every $t > 0$,
\[
\Pr\Bigl( \bigl|\langle \varepsilon^{(m)}, a \rangle\bigr| > t
\Bigr)
\;\le\; 2 \exp\Bigl( -\frac{t^2}{2 \sigma^2 \|a\|^2} \Bigr) .
\]

\paragraph{Theorem~\ref{thm:ident}, restated.}
Suppose the unit-norm profiles $s_1, \dots, s_N \in \R^d$ satisfy
the spark condition with margin $s_{\min}$
(Definition~\ref{def:spark}) and the data follow the model above.

\emph{(a)} The matrix $\delta^{\star}$ with rows
$\mathbf{1}_{J_m}$ is the unique minimizer of
$\sum_{m,j} \delta_{mj}$ among all $\delta$ admitting an exact
factorization of $X_0$ over the pool.

\emph{(b)} Fix $\alpha \in (0,1)$ and set
$L_\alpha = \sqrt{2 \log(6NM/\alpha)}$. Assume the SNR condition
and the window
\[
c_{\min} s_{\min} \;>\; 2 \sigma L_\alpha ,
\qquad
\sigma^2 L_\alpha^2 \;<\; \lambda \;<\;
\bigl( c_{\min} s_{\min} - \sigma L_\alpha \bigr)^2 .
\]
Then with probability at least $1 - \alpha$ over the noise,
simultaneously for every sample $m$ and every $A \subseteq [N]$
obtained from $J_m$ by adding or removing one component,
\[
\ell_\lambda(A;\, x^{(m)}) - \ell_\lambda(J_m;\, x^{(m)})
\;\ge\;
\min\Bigl\{\, \lambda - \sigma^2 L_\alpha^2,\ \
(c_{\min} s_{\min} - \sigma L_\alpha)^2 - \lambda \,\Bigr\}
\;>\; 0 .
\]
At $\sigma = 0$: for every $0 < \lambda < c_{\min}^2 s_{\min}^2$,
each $J_m$ is the unique global minimizer of
$\ell_\lambda(\,\cdot\,;\, x_0^{(m)})$ among selections with
$|A| \le k$.

\begin{proof}[Proof of part (a)]
\emph{Step 1 (two sparse representations coincide).} Suppose
\[
\sum_{i \in J} c_i s_i \;=\; \sum_{i \in J'} c'_i s_i ,
\qquad |J| \le k, \quad |J'| \le k ,
\]
with all listed coefficients nonzero. Set $A = J \cup J'$, so
$|A| \le |J| + |J'| \le 2k$, and define $v \in \R^{|A|}$
coordinate-wise, exactly as in Step~1 of
Appendix~\ref{app:prop1}:
\[
v_i =
\begin{cases}
c_i, & i \in J \setminus J',\\
c_i - c'_i, & i \in J \cap J',\\
-\,c'_i, & i \in J' \setminus J.
\end{cases}
\]
By linearity of the matrix--vector product,
\[
S_A v \;=\; \sum_{i \in J} c_i s_i - \sum_{i \in J'} c'_i s_i
\;=\; 0 .
\]
If $(J, c) \ne (J', c')$ then $v \ne 0$, and the spark condition
gives
\[
0 \;=\; \|S_A v\| \;\ge\; \sigma_{\min}(S_A)\, \|v\|
\;\ge\; s_{\min} \|v\| \;>\; 0 ,
\]
a contradiction. Hence $J = J'$ and $c = c'$: a noiseless sample
has exactly one representation using at most $k$ pool components
with nonzero coefficients.

\emph{Step 2 (every row spends at least the truth).} Let $\delta$
admit an exact factorization with row supports $A_m$ and realizing
coefficients $c'$, and let
\[
A'_m \;=\; \{\, j \in A_m : c'_j \ne 0 \,\} \;\subseteq\; A_m
\]
be the nonzero pattern on row $m$. Two cases.

\emph{Case $|A_m| \le k$.} Then $|A'_m| \le k$, and sample $m$ has
the two representations
\[
x_0^{(m)} \;=\; \sum_{i \in J_m} c^{(m)}_i s_i
\;=\; \sum_{j \in A'_m} c'_j s_j ,
\]
each using at most $k$ components with nonzero coefficients.
Step~1 forces $A'_m = J_m$, hence $J_m \subseteq A_m$ and
\[
\sum_{j} \delta_{mj} \;=\; |A_m| \;\ge\; |J_m| ,
\qquad \text{with equality exactly when } A_m = J_m .
\]

\emph{Case $|A_m| > k$.} Then directly
\[
\sum_{j} \delta_{mj} \;=\; |A_m| \;>\; k \;\ge\; |J_m| .
\]

\emph{Step 3 (summation).} Summing the row bounds of Step~2 over
$m$,
\[
\sum_{m,j} \delta_{mj} \;\ge\; \sum_{m} |J_m|
\;=\; \sum_{m,j} \delta^{\star}_{mj} ,
\]
with equality exactly when $A_m = J_m$ for every $m$, that is,
when $\delta = \delta^{\star}$. Hence $\delta^{\star}$ is the
unique minimizer.
\end{proof}

\begin{proof}[Proof of part (b)]
\emph{Step 1 (one-flip fit identity).} Every index set appearing
in Steps~1--7 has at most $k + 1 \le 2k$ elements, so by the spark
condition every $S_A$ below has full column rank. Write
$V = \mathrm{span}(S_A)$; the \emph{unconstrained} least-squares
fit equals the squared distance to the span,
\[
\min_{c \in \R^{|A|}}
\Bigl\| x - \sum_{j \in A} c_j s_j \Bigr\|^2
\;=\; \| x - P_V x \|^2 ,
\]
and we write, for the unconstrained counterpart of
Eq.~\ref{eq:penobj},
\[
\ell^{\mathrm{u}}_\lambda(A;\, x)
\;=\; \| x - P_V x \|^2 + \lambda |A| .
\]
Let $V'$ extend $V$ by one component and let $u$ be the unit
vector spanning $V' \cap V^{\perp}$. Since
$P_{V'} x = P_V x + \langle x, u \rangle u$ (projection onto an
orthogonal decomposition of $V'$), the Pythagorean identity gives
\[
\| x - P_{V'} x \|^2
\;=\; \| x - P_V x \|^2 - \langle x, u \rangle^2 :
\]
one flip changes the unconstrained fit by exactly
$\langle x, u \rangle^2$.

\emph{Step 2 (adding a spurious component).} Fix a sample $m$ and
$j \notin J_m$; take $V = \mathrm{span}(S_{J_m})$ and
$V' = \mathrm{span}(S_{J_m \cup \{j\}})$. Since
$x_0^{(m)} \in V$ and $u \perp V$,
\[
\langle x^{(m)}, u \rangle
\;=\; \langle x_0^{(m)}, u \rangle
+ \langle \varepsilon^{(m)}, u \rangle
\;=\; \langle \varepsilon^{(m)}, u \rangle ,
\]
and by Step~1 the objective change of the addition is
\[
\ell^{\mathrm{u}}_\lambda(J_m \cup \{j\};\, x^{(m)})
- \ell^{\mathrm{u}}_\lambda(J_m;\, x^{(m)})
\;=\; \lambda - \langle \varepsilon^{(m)}, u \rangle^2 .
\]

\emph{Step 3 (removing a true component).} Fix $i \in J_m$; take
$V = \mathrm{span}(S_{J_m \setminus \{i\}})$ and
$V' = \mathrm{span}(S_{J_m})$. Here $u$ is the normalized
component of $s_i$ orthogonal to $V$, so
\[
\langle x_0^{(m)}, u \rangle
\;=\; \sum_{i' \in J_m} c^{(m)}_{i'} \langle s_{i'}, u \rangle
\;=\; c^{(m)}_i \langle s_i, u \rangle ,
\qquad
\langle s_i, u \rangle = \dist(s_i, V) .
\]
The distance is bounded below by the spark margin: for any
coefficient vector $w$ on $J_m \setminus \{i\}$,
\[
\Bigl\| s_i - \sum_{j \in J_m \setminus \{i\}} w_j s_j \Bigr\|
\;=\; \| S_{J_m} v \|
\;\ge\; \sigma_{\min}(S_{J_m})\, \|v\|
\;\ge\; s_{\min} ,
\]
where $v$ carries $-w$ on $J_m \setminus \{i\}$ and $1$ at $i$,
so $\|v\| \ge 1$; minimizing over $w$ gives
$\dist(s_i, V) \ge s_{\min}$. Combining, with
$c^{(m)}_i \ge c_{\min}$,
\[
\bigl| \langle x^{(m)}, u \rangle \bigr|
\;\ge\; \bigl| \langle x_0^{(m)}, u \rangle \bigr|
- \bigl| \langle \varepsilon^{(m)}, u \rangle \bigr|
\;\ge\; c_{\min} s_{\min}
- \bigl| \langle \varepsilon^{(m)}, u \rangle \bigr| ,
\]
and by Step~1 the objective change of the removal is
\[
\ell^{\mathrm{u}}_\lambda(J_m \setminus \{i\};\, x^{(m)})
- \ell^{\mathrm{u}}_\lambda(J_m;\, x^{(m)})
\;=\; \langle x^{(m)}, u \rangle^2 - \lambda .
\]

\emph{Step 4 (dual vectors).} For each sample $m$ and each
$i \in J_m$, set $V_i = \mathrm{span}(S_{J_m \setminus \{i\}})$
and define
\[
a^{(m)}_i \;=\;
\frac{ s_i - P_{V_i} s_i }{ \| s_i - P_{V_i} s_i \|^2 } ,
\]
well defined since
$\| s_i - P_{V_i} s_i \| = \dist(s_i, V_i) \ge s_{\min} > 0$ by
Step~3. The numerator is orthogonal to $V_i$, so
\[
\langle a^{(m)}_i, s_j \rangle = 0
\quad \text{for } j \in J_m \setminus \{i\} ,
\qquad
\langle a^{(m)}_i, s_i \rangle
= \frac{ \bigl\langle s_i - P_{V_i} s_i,\,
( s_i - P_{V_i} s_i ) + P_{V_i} s_i \bigr\rangle }
{ \| s_i - P_{V_i} s_i \|^2 }
= 1 ,
\]
and its norm is controlled by the same distance:
\[
\| a^{(m)}_i \|
\;=\; \frac{1}{ \| s_i - P_{V_i} s_i \| }
\;=\; \frac{1}{ \dist(s_i, V_i) }
\;\le\; \frac{1}{ s_{\min} } .
\]
Both $s_i$ and $P_{V_i} s_i$ lie in $\mathrm{span}(S_{J_m})$, so
$a^{(m)}_i \in \mathrm{span}(S_{J_m})$.

\emph{Step 5 (the noise event).} The directions $u$ of
Steps~2--3 are at most $2NM$ fixed unit vectors ($N$ additions
and $k \le N$ removals per sample), and the dual vectors of
Step~4 are at most $kM \le NM$ further fixed vectors; all are
determined by $S$ and $\{J_m\}$, never by the noise. Apply the
sub-Gaussian tail (Notation) with $t = \sigma L_\alpha \|a\|$ to
each of these at most $3NM$ fixed vectors:
\[
\Pr\Bigl( \bigl| \langle \varepsilon^{(m)}, a \rangle \bigr|
> \sigma L_\alpha \|a\| \Bigr)
\;\le\; 2 \exp\bigl( -L_\alpha^2 / 2 \bigr)
\;=\; 2 \cdot \frac{\alpha}{6NM}
\;=\; \frac{\alpha}{3NM} .
\]
By the union bound over the at most $3NM$ events, with probability
at least $1 - \alpha$ the following event $E$ holds:
\begin{gather*}
E: \qquad
\bigl| \langle \varepsilon^{(m)}, u \rangle \bigr|
\le \sigma L_\alpha
\ \ \text{for every flip direction,}\\
\bigl| \langle \varepsilon^{(m)}, a^{(m)}_i \rangle \bigr|
\le \frac{\sigma L_\alpha}{s_{\min}}
\ \ \text{for every } m,\ i \in J_m .
\end{gather*}
Steps~6--7 run on $E$.

\emph{Step 6 (the nonnegativity constraint is inactive at the
truth).} $\ell_\lambda$ constrains $c \ge 0$; Steps~1--3 used
unconstrained fits. Write the projection of $x^{(m)}$ onto
$V = \mathrm{span}(S_{J_m})$ in pool coordinates,
\[
P_V x^{(m)} \;=\; \sum_{j \in J_m} \hat c^{(m)}_j s_j .
\]
Pairing with the dual vectors of Step~4: for every $i \in J_m$,
\begin{align*}
\hat c^{(m)}_i
\;&=\; \bigl\langle a^{(m)}_i,\, P_V x^{(m)} \bigr\rangle
\;=\; \bigl\langle a^{(m)}_i,\, x^{(m)} \bigr\rangle\\
\;&=\; \bigl\langle a^{(m)}_i,\, x_0^{(m)} \bigr\rangle
+ \bigl\langle a^{(m)}_i,\, \varepsilon^{(m)} \bigr\rangle
\;=\; c^{(m)}_i
+ \bigl\langle a^{(m)}_i,\, \varepsilon^{(m)} \bigr\rangle ,
\end{align*}
by, in order: biorthogonality (Step~4) applied to the sum;
$a^{(m)}_i \in V$, so the residual
$x^{(m)} - P_V x^{(m)} \perp a^{(m)}_i$ contributes nothing;
$x^{(m)} = x_0^{(m)} + \varepsilon^{(m)}$; biorthogonality applied
to $x_0^{(m)} = \sum_{j \in J_m} c^{(m)}_j s_j$. On $E$,
entrywise,
\[
\hat c^{(m)}_i
\;\ge\; c_{\min} - \frac{\sigma L_\alpha}{s_{\min}}
\;>\; c_{\min} - \frac{c_{\min}}{2}
\;=\; \frac{c_{\min}}{2} \;>\; 0 ,
\]
using the SNR condition in the form
$\sigma L_\alpha / s_{\min} < c_{\min}/2$. The coefficients
$\hat c^{(m)}$ are therefore feasible for the constraint
$c \ge 0$ and attain the unconstrained fit value
$\| x^{(m)} - P_V x^{(m)} \|^2$, so the two fits coincide at
$J_m$:
\[
\ell_\lambda(J_m;\, x^{(m)})
\;=\; \ell^{\mathrm{u}}_\lambda(J_m;\, x^{(m)}) .
\]
For every other selection $A$, the constrained minimum runs over a
subset of coefficient vectors, so
\[
\ell_\lambda(A;\, x^{(m)})
\;\ge\; \ell^{\mathrm{u}}_\lambda(A;\, x^{(m)}) .
\]

\emph{Step 7 (the window verdict).} On $E$, chaining Step~6,
Step~2, and Step~5, an addition costs
\[
\ell_\lambda(J_m \cup \{j\};\, x^{(m)})
- \ell_\lambda(J_m;\, x^{(m)})
\;\ge\; \lambda - \langle \varepsilon^{(m)}, u \rangle^2
\;\ge\; \lambda - \sigma^2 L_\alpha^2
\;>\; 0
\]
by the left edge of the window. Chaining Step~6, Step~3, and
Step~5 (the SNR condition makes
$c_{\min} s_{\min} - \sigma L_\alpha > 0$), a removal costs
\[
\ell_\lambda(J_m \setminus \{i\};\, x^{(m)})
- \ell_\lambda(J_m;\, x^{(m)})
\;\ge\; \langle x^{(m)}, u \rangle^2 - \lambda
\;\ge\; \bigl( c_{\min} s_{\min} - \sigma L_\alpha \bigr)^2
- \lambda
\;>\; 0
\]
by the right edge. Every one-flip change therefore costs at least
\[
\min\Bigl\{\, \lambda - \sigma^2 L_\alpha^2,\ \
\bigl( c_{\min} s_{\min} - \sigma L_\alpha \bigr)^2 - \lambda
\,\Bigr\} ,
\]
the stated gap.

\emph{Step 8 (noiseless global optimality).} Let $\sigma = 0$, so
$x^{(m)} = x_0^{(m)}$ and the constrained fit at $J_m$ is exact
with the feasible coefficients $c^{(m)}$. Let $A \ne J_m$ with
$|A| \le k$ and set $r = |J_m \setminus A|$. Two cases.

\emph{Case $r = 0$.} Then $J_m \subsetneq A$, the fit is $0$ at
both selections, and
\[
\ell_\lambda(A;\, x_0^{(m)}) - \ell_\lambda(J_m;\, x_0^{(m)})
\;=\; \lambda \bigl( |A| - |J_m| \bigr) \;\ge\; \lambda \;>\; 0 .
\]

\emph{Case $r \ge 1$.} For any coefficient vector $c$ supported on
$A$ (of any sign),
\[
x_0^{(m)} - S_A c \;=\; S_{J_m \cup A}\, v ,
\]
where $v$ carries $c^{(m)}_j$ at $j \in J_m \setminus A$,
$c^{(m)}_j - c_j$ on $J_m \cap A$, and $-c_j$ on
$A \setminus J_m$. Since $|J_m \cup A| \le 2k$, the spark
condition gives
\[
\bigl\| x_0^{(m)} - S_A c \bigr\|^2
\;\ge\; s_{\min}^2 \|v\|^2
\;\ge\; s_{\min}^2 \sum_{j \in J_m \setminus A}
\bigl( c^{(m)}_j \bigr)^2
\;\ge\; r\, c_{\min}^2 s_{\min}^2 .
\]
The penalty difference is bounded using $|A| \ge |J_m| - r$:
\[
\lambda \bigl( |A| - |J_m| \bigr) \;\ge\; -\,\lambda r ,
\]
hence
\[
\ell_\lambda(A;\, x_0^{(m)}) - \ell_\lambda(J_m;\, x_0^{(m)})
\;\ge\; r\, c_{\min}^2 s_{\min}^2 - \lambda r
\;=\; r \bigl( c_{\min}^2 s_{\min}^2 - \lambda \bigr)
\;>\; 0 .
\]
Both cases are strict, so $J_m$ is the unique global minimizer
among selections with $|A| \le k$.
\end{proof}

\noindent
Part (a) uses only the spark condition and the sparsity cap:
neither the concentration floor, the ceiling, nor nonnegativity of
the competing coefficients enters, so uniqueness of the minimal
occurrence matrix holds against arbitrary sign patterns. The floor
and nonnegativity enter part (b): the floor sets the removal cost
$(c_{\min} s_{\min} - \sigma L_\alpha)^2$, and nonnegativity is
neutralized in Step~6.

\medskip
\noindent
\emph{Relation to magnitude penalties.} Eq.~\ref{eq:penobj}
charges the selection cardinality
$|A| = \sum_j \delta_{mj}$, the per-sample summand of the total
usage $\sum_{m,j} \delta_{mj}$ minimized in part (a); the inner
minimization over $c$ carries no penalty. A ridge or lasso penalty
\citep{hoerl1970ridge,tibshirani1996regression} instead charges
$\|c\|_2^2$ or $\|c\|_1$ and perturbs every retained coefficient.
Ridge sets no coefficient to zero, so its support must be fixed by
thresholding outside the objective; the lasso does set
coefficients to zero through its objective, but which ones is a
by-product of magnitude shrinkage: the surviving support depends
on coefficient scales and the regularization path, and neither
objective prices the count $|A|$ itself.
Relevance-determination methods \citep{tipping2001sparse} likewise
prune by magnitude, with no priced count.
Appendix~\ref{app:toy} computes the difference on data that admit
infinitely many exact solutions: $\ell_1$ and $\ell_2$ each select
a specific wrong member with certainty. On such data no criterion
is guaranteed correct, the cardinality criterion included; what
differs is the cause of failure, not the outcome. A magnitude
penalty selects by built-in preference and is wrong with
certainty; the cardinality criterion fails there only for lack of
information, and is provably correct once the data supply it (in
the toy, one further sample; in general,
Theorem~\ref{thm:ident}).

\section{Formal Statement and Proof of Theorem~2 (Certified
Recovery)}
\label{app:thm2}

This appendix states and proves Theorem~2 in full, together with
Corollary~1. Its role in the argument is the second obstacle of
\S\ref{sec:theory}: Theorem~1 makes the truth the optimum of a
searchable criterion, but no optimizer is guaranteed to reach an
optimum, and a returned solution cannot be compared against the
unseen truth, so a search can succeed without anyone being able to
know it. Theorem~2 removes the need to trust the search: it
provides conditions on the returned solution (the certificate;
their observability status is stated with their definitions below)
under which any solution satisfying them, no matter how it was produced,
is the truth. Every guarantee below
therefore quantifies over certificate-passing solutions, never
over the algorithm that found them.

Part (a) treats a known pool: on one named noise event
of probability at least $1 - \alpha$, every solution that passes
the certificate (each sample reconstructed within tolerance by at
most $k$ components; total usage no larger than the truth's)
recovers every sample's support exactly, hence the exact component
count, and every concentration to noise-floor accuracy; the
guarantee quantifies over all certificate-passing solutions at
once, with no assumption on how they were found. Part (b) treats a
learned pool on noiseless data: every certificate-passing solution
recovers the active components exactly and correctly signed, up to
an injective relabeling $\pi$, together with exact supports,
component count, and concentrations. Part (c) treats the learned
pool under noise and is conditional: its linear-algebraic content
is proved in full below, and the one remaining step, a
combinatorial statement about how samples distribute over candidate
sets, is stated formally as Conjecture~R and assumed, never
claimed. Corollary~1 specializes part (a) to a single fresh sample
decoded over the recovered pool.

The proofs are organized in five layers. Part (a): Layer I
(Lemma~\ref{lem:thm2-event}) constructs the noise event $E$ and
extracts the single inequality (\ref{eq:thm2-nform}) through which
every later step uses $E$; Layer II (Lemmas~\ref{lem:thm2-detect}
and \ref{lem:thm2-count}) proves, deterministically on $E$,
detection of missing components and usage counting; Layer III
(Lemma~\ref{lem:thm2-coef}) adds coefficient stability, and the
assembly takes five steps. Part (b), Layer IV, runs through three
lemmas
(span intersection along an independent family,
Lemma~\ref{lem:thm2-spanint}; tight usage with exact spans,
Lemma~\ref{lem:thm2-tight}; anchoring of each active component to a
unique pool slot, Lemma~\ref{lem:thm2-anchor}) and a five-step
assembly. Part (c), Layer V, proves three unconditional robust
lemmas (a
crude coefficient bound, Lemma~\ref{lem:thm2-crude}; a
perturbation-stable dimension count, Lemma~\ref{lem:thm2-dim}; a
two-subspace pinch, Lemma~\ref{lem:thm2-pinch}), one unconditional
lemma quantifying why out-of-span errors cannot cancel across
coefficient-diverse samples (Lemma~\ref{lem:thm2-serve}), closes
the $1$-sparse case unconditionally (Lemma~\ref{lem:thm2-k1}), and
then proves the reduction to Conjecture~R in six steps.
Corollary~1 is part (a) at $M = 1$.

\paragraph{Notation.}
The notation of Appendices~\ref{app:prop1} and \ref{app:thm1}
remains in force. New objects:
\begin{center}
\begin{tabular}{@{}ll@{}}
\toprule
$J_{\mathrm{act}},\ K^{\ast}$ & $J_{\mathrm{act}} = \bigcup_m J_m$,
the active components; $K^{\ast} = |J_{\mathrm{act}}|$, their
count\\
$V_T$ & $= \mathrm{span}(s_i : i \in T)$, the span of the true
profiles indexed by $T$\\
$Q_V$ & $= I - P_V$, the projection onto the orthogonal complement
of the subspace $V$\\
$n_{2k}$ & $= \sum_{r=0}^{2k} \binom{N}{r}$, the number of pool
subsets of size at most $2k$\\
$\varepsilon_{\mathrm{ns}}(\alpha)$ & $= 2\sigma
\sqrt{2\bigl(d \log 5 + \log(2M/\alpha)\bigr)}$, the noise-norm
level\\
$\varepsilon_{\mathrm{prj}}(\alpha)$ & $= 2\sigma
\sqrt{2\bigl(2k \log 5 + \log(2M n_{2k}/\alpha)\bigr)}$, the
projected-noise level\\
$\varepsilon_{\mathrm{fit}}$ & fit tolerance of the certificate; a
constant chosen by the analyst\\
$\hat S = (\hat s_1, \dots, \hat s_N)$ & the learned pool,
unit-norm columns $\hat s_j$ (parts (b) and (c))\\
$\hat s_{\min}$ & assumed regularity margin of the learned pool\\
$\bar\varepsilon$ & $= \varepsilon_{\mathrm{fit}} +
\varepsilon_{\mathrm{ns}}(\alpha)$, the signal-level tolerance
(part (c))\\
$\vartheta$ & significance floor on used coefficients (part (c))\\
$\gamma_{\mathrm{c}}$ & coefficient-diversity quantile
(Conjecture~R)\\
$\hat C$ & $= (k\, c_{\max} + \bar\varepsilon)/\hat s_{\min}$, the
coefficient bound of Lemma~\ref{lem:thm2-crude}\\
$\pi$ & relabeling: an injective map
$J_{\mathrm{act}} \to [N]$ (see below)\\
$\|\cdot\|_F,\ \|\cdot\|_{\mathrm{op}}$ & Frobenius and operator
norms of matrices\\
\bottomrule
\end{tabular}
\end{center}
The noise vectors $\varepsilon^{(m)}$ are independent, mean zero,
and sub-Gaussian with proxy $\sigma$ in the moment-generating
sense: $\mathbb{E}\, e^{\lambda \langle u, \varepsilon^{(m)}
\rangle} \le e^{\lambda^2 \sigma^2 / 2}$ for every unit
$u \in \R^d$ and every $\lambda \in \R$
\citep[\S 2.5]{vershynin2018high}. This implies the one-sided tail
bound $\Pr(\langle u, \varepsilon^{(m)} \rangle \ge t) \le
e^{-t^2/2\sigma^2}$ used in Lemma~\ref{lem:thm2-event} and the
two-sided tail form of Appendix~\ref{app:thm1}.

A \emph{candidate solution over a known pool} is a family
$(\{A_m\}_{m=1}^{M}, \{\hat c^{(m)}\}_{m=1}^{M})$ of supports
$A_m \subseteq [N]$ and coefficient vectors $\hat c^{(m)}$
supported on $A_m$. The \emph{known-pool certificate}:
\begin{itemize}[leftmargin=*]
\item \emph{(C1)} $|A_m| \le k$ and
$\bigl\| x^{(m)} - S_{A_m} \hat c^{(m)} \bigr\| \le
\varepsilon_{\mathrm{fit}}$ for every $m \in [M]$;
\item \emph{(C2)} $\sum_{m} |A_m| \le \sum_{m} |J_m|$.
\end{itemize}
A \emph{candidate solution with learned pool} is a triple
$(\hat S, \{A_m\}, \{\hat c^{(m)}\})$ in which the pool $\hat S$ is
itself estimated, with $\|\hat s_j\| = 1$ and every used
coefficient strictly positive ($\hat c^{(m)}_j > 0$ for
$j \in A_m$; zero-coefficient slots are dropped from $A_m$, which
only lowers usage). In parts (b) and (c) the index set $[N]$
indexes the learned pool's slots, while the true supports $J_m$
retain their meaning from the model; the true profiles enter only
through $(s_i)_{i \in J_{\mathrm{act}}}$ and their spans $V_T$. The
\emph{noiseless certificate}:
\begin{itemize}[leftmargin=*]
\item \emph{(C3)} $|A_m| \le k$ and
$\sum_{j \in A_m} \hat c^{(m)}_j \hat s_j = x_0^{(m)}$ for every
$m$;
\item \emph{(C4)} $\sum_m |A_m| \le \sum_m |J_m|$;
\item \emph{(C5)} $\sigma_{\min}(\hat S_B) \ge \hat s_{\min} > 0$
for every $B \subseteq [N]$ with $|B| \le 2k$;
\item \emph{(C6)} for every $i \in J_{\mathrm{act}}$ there exist
samples $m, m'$ with $J_m \cap J_{m'} = \{i\}$;
\item \emph{(C7)} $\dim\bigl(\mathrm{span}(\hat S_{A_m}) \cap
V_{J_m}\bigr) = |J_m|$ for every $m$.
\end{itemize}
The \emph{robust certificate}:
\begin{itemize}[leftmargin=*]
\item \emph{(C8)} $|A_m| \le k$,
$\bigl\| x^{(m)} - \hat S_{A_m} \hat c^{(m)} \bigr\| \le
\varepsilon_{\mathrm{fit}}$, and $\hat c^{(m)}_j \ge \vartheta > 0$
for $j \in A_m$, for every $m$;
\item \emph{(C9)} $\sum_m |A_m| \le \sum_m |J_m|$;
\item \emph{(C10)} $\sigma_{\min}(\hat S_B) \ge \hat s_{\min} > 0$
for every $B \subseteq [N]$ with $|B| \le 2k$;
\item \emph{(C11)} $|A_m| \ge |J_m|$ for every $m$;
\end{itemize}
together with (C6). The conditions differ in what can be checked
at run time, as follows. The fit conditions (C1), (C3),
and (C8) are computable from data and solution alone. The
usage-parsimony conditions (C2), (C4), and (C9) compare against
the unknown truth and are not computable at run time: they are
hypotheses on what the run achieved, promoted by the usage penalty
inside the $\lambda$-window of Theorem~\ref{thm:ident}(b), and
they are not vacuous, because on the noise event below the true
configuration itself passes the fit condition with equality in
usage (final claim of part (a)). Conditions (C5) and (C10) are
promoted by the ambiguity penalty; (C6) is an occurrence-richness
assumption on the sampled supports; (C7) states that no candidate
set represents a sample's span with fewer directions than the
truth (its role is discussed after the proofs); (C11) is the
cardinality form of (C7) under noise. As with $s_{\min}$ and
$\sigma_{\min}(S_A)$ (Appendix~\ref{app:prop1}),
$\sigma_{\min}(\hat S_B)$ is a quantity computed from the learned
pool while $\hat s_{\min}$ is an assumed constant; (C5) and (C10)
assert that the computed quantity dominates the assumed margin.

Finally, slot indices in a learned pool are bookkeeping, not
structure: permuting the slots of $\hat S$ together with the
columns of the occurrence matrix changes no generated sample, so
no data can distinguish two pools that differ by a permutation.
Recovery of components is therefore stated up to a
\emph{relabeling} $\pi$: an
injective map $\pi : J_{\mathrm{act}} \to [N]$ assigning to each
active true component $i$ the pool slot $\pi(i)$ that represents
it. Exact recovery then reads $\hat s_{\pi(i)} = s_i$ and
$A_m = \pi(J_m)$; the relabeling is the only ambiguity the model
leaves.

\paragraph{Theorem~2, restated: parts (a) and (b).}
Fix $\alpha \in (0,1)$ and $\varepsilon_{\mathrm{fit}} \ge 0$.
Suppose the data follow the model of Appendix~\ref{app:thm1}:
$x^{(m)} = x_0^{(m)} + \varepsilon^{(m)}$ with
$x_0^{(m)} = \sum_{i \in J_m} c^{(m)}_i s_i$, $|J_m| \le k$,
$c^{(m)}_i \in [c_{\min}, c_{\max}]$, and noise as in the Notation.
Let $E$ denote the event that
\[
\bigl\| \varepsilon^{(m)} \bigr\| \le
\varepsilon_{\mathrm{ns}}(\alpha)
\qquad \text{and} \qquad
\bigl\| P_{V_B}\, \varepsilon^{(m)} \bigr\| \le
\varepsilon_{\mathrm{prj}}(\alpha)
\]
for every $m \in [M]$ and every $B \subseteq [N]$ with
$|B| \le 2k$.

\emph{(a) (Known pool.)} Suppose the pool $s_1, \dots, s_N$ is
known and satisfies the spark condition with margin $s_{\min}$
(Definition~\ref{def:spark}), and suppose the fixed-pool window
\begin{equation}
\varepsilon_{\mathrm{ns}}(\alpha) \;\le\;
\varepsilon_{\mathrm{fit}},
\qquad
\varepsilon_{\mathrm{fit}} + \varepsilon_{\mathrm{prj}}(\alpha)
\;<\; c_{\min}\, s_{\min} .
\label{eq:thm2-wfix}
\end{equation}
Then $\Pr(E) \ge 1 - \alpha$, and on $E$, for every candidate
solution $(\{A_m\}, \{\hat c^{(m)}\})$ that passes the certificate
(C1)--(C2), the following hold:
\emph{(i)} $A_m = J_m$ for every $m \in [M]$;
\emph{(ii)} $\bigl| \bigcup_m A_m \bigr| = K^{\ast}$;
\emph{(iii)} $\| \hat c^{(m)} - c^{(m)} \| \le
(\varepsilon_{\mathrm{fit}} +
\varepsilon_{\mathrm{prj}}(\alpha))/s_{\min}$ for every $m$.
Moreover, on $E$ the true configuration
$(\{J_m\}, \{c^{(m)}\})$ passes the certificate: (C1) holds, and
(C2) holds with equality.

\emph{(b) (Learned pool, noiseless.)} Suppose $\sigma = 0$, so
$x^{(m)} = x_0^{(m)}$, and suppose the true pool satisfies the
spark condition with margin $s_{\min}$
(Definition~\ref{def:spark}). Consider any candidate solution
$(\hat S, \{A_m\}, \{\hat c^{(m)}\})$ that passes the certificate
(C3)--(C7). Then
there is an injective $\pi : J_{\mathrm{act}} \to [N]$ such that:
\emph{(i)} $\hat s_{\pi(i)} = s_i$ for every
$i \in J_{\mathrm{act}}$;
\emph{(ii)} $A_m = \pi(J_m)$ for every $m$;
\emph{(iii)} $\bigl| \bigcup_m A_m \bigr| = K^{\ast}$;
\emph{(iv)} $\hat c^{(m)}_{\pi(i)} = c^{(m)}_i$ for every $m$ and
every $i \in J_m$.

\paragraph{Conjecture~R (isolating pairs), restated.}
(Called \emph{anchor spanning service}, Open Lemma R-A$''$, in
the technical report.)
Fix $\gamma_{\mathrm{c}} > 0$. For a support pattern
$J$ with $|J| = \varkappa$ and a candidate set $A \subseteq [N]$
with $|A| = \varkappa$, call the pair $(J, A)$
\emph{$\gamma_{\mathrm{c}}$-spanned} if there exist samples
$m_1, \dots, m_\varkappa \in [M]$ with
\[
J_{m_t} = J
\quad \text{and} \quad
A_{m_t} = A
\qquad \text{for every } t \in [\varkappa],
\]
whose coefficient matrix
$C = [\, c^{(m_1)} \cdots c^{(m_\varkappa)} \,] \in
\R^{\varkappa \times \varkappa}$ (rows indexed by $J$) satisfies
$\sigma_{\min}(C) \ge \gamma_{\mathrm{c}}$. Conjecture~R asserts:
under the model of Appendix~\ref{app:thm1}, the spark condition
(Definition~\ref{def:spark}), and the robust certificate
(C8)--(C11) with (C6), for every active
$i \in J_{\mathrm{act}}$ there exists a pair of samples $(m, m')$
with $J_m \cap J_{m'} = \{i\}$ such that $(J_m, A_m)$ and
$(J_{m'}, A_{m'})$ are both $\gamma_{\mathrm{c}}$-spanned.

Conjecture~R is open for $k \ge 2$; it is a theorem for $k = 1$
(Lemma~\ref{lem:thm2-k1}), and its role vanishes at
$\bar\varepsilon = 0$ (closing remarks). Part (c) is stated and
proved conditionally on it.

\paragraph{Theorem~2, restated: part (c).}
Retain the setup and the event $E$ above.

\emph{(c) (Learned pool, noisy; conditional on Conjecture~R.)}
Suppose the true pool satisfies the spark condition with margin
$s_{\min}$ (Definition~\ref{def:spark}), and consider candidate
solutions $(\hat S, \{A_m\}, \{\hat c^{(m)}\})$ passing the
robust certificate (C8)--(C11) with (C6). Assume Conjecture~R
holds with quantile $\gamma_{\mathrm{c}} > 0$. Define
\[
f \;=\; \frac{k\, \bar\varepsilon}
{s_{\min} \gamma_{\mathrm{c}} - \sqrt{k}\, \bar\varepsilon},
\qquad
f' \;=\; f + \frac{2\sqrt{2k}\, f}{s_{\min}},
\qquad
\kappa_{\mathrm{c}} \;=\;
\frac{\bar\varepsilon + \sqrt{k}\, \hat C f'}{s_{\min}},
\]
and suppose the robust window
\begin{equation}
\begin{gathered}
\sqrt{k}\, \bar\varepsilon < s_{\min} \gamma_{\mathrm{c}},
\qquad
\sqrt{2k}\, f < \hat s_{\min},
\qquad
f' \Bigl( 1 + \frac{2}{s_{\min}} \Bigr) < 1,
\\
\kappa_{\mathrm{c}} < c_{\min},
\qquad
\frac{\bar\varepsilon + k\, c_{\max} f' / (1 - f')}
{\hat s_{\min}} < \vartheta .
\end{gathered}
\label{eq:thm2-wrob}
\end{equation}
Then $\Pr(E) \ge 1 - \alpha$, and on $E$, every candidate solution
satisfying the robust certificate admits an injective
$\pi : J_{\mathrm{act}} \to [N]$ such that:
\emph{(i)} $\dist\bigl( \hat s_{\pi(i)},\,
\mathrm{span}(s_i) \bigr) \le f'$ and
$\langle \hat s_{\pi(i)}, s_i \rangle > 0$ for every
$i \in J_{\mathrm{act}}$;
\emph{(ii)} $A_m = \pi(J_m)$ for every $m$;
\emph{(iii)} $\bigl| \bigcup_m A_m \bigr| = K^{\ast}$;
\emph{(iv)} $\bigl| \hat c^{(m)}_{\pi(i)} - c^{(m)}_i \bigr| \le
\kappa_{\mathrm{c}} + \hat C f'$ for every $m$ and every
$i \in J_m$.

\paragraph{Corollary~1 (Reuse), restated.}
Fix $\alpha \in (0,1)$ and $\varepsilon_{\mathrm{fit}} \ge 0$.
Suppose the pool is $(s_i)_{i \in J_{\mathrm{act}}}$: either a
known component library, or the $K^{\ast}$ components recovered by
part (b) and retained by the frozen model. Consider a fresh sample
from the same generative process,
\[
x \;=\; \sum_{i \in J} c_i s_i + \varepsilon,
\qquad
J \subseteq J_{\mathrm{act}}, \quad |J| \le k, \quad
c_i \in [c_{\min}, c_{\max}],
\]
with $\varepsilon$ sub-Gaussian with proxy $\sigma$ (Notation).
Write $\varepsilon^{(1)}_{\mathrm{ns}}(\alpha)$ and
$\varepsilon^{(1)}_{\mathrm{prj}}(\alpha)$ for the levels of the
Notation evaluated at $M = 1$, with $N$ replaced by $K^{\ast}$ in
$n_{2k}$, and suppose the window (\ref{eq:thm2-wfix}) holds with
these levels. Then with probability at least $1 - \alpha$ over
$\varepsilon$, every decoding $(A, \hat c)$ of $x$ alone with
\[
|A| \le k,
\qquad
\| x - S_A \hat c \| \le \varepsilon_{\mathrm{fit}},
\qquad
|A| \le |J|,
\]
satisfies $A = J$ and
$\| \hat c - c \| \le
(\varepsilon_{\mathrm{fit}} +
\varepsilon^{(1)}_{\mathrm{prj}}(\alpha))/s_{\min}$.

\bigskip
\noindent
The proofs follow in five layers; each part's supporting layers
immediately precede its assembly. Layer I confines all randomness
to one event. Layer II (detection and counting) proves part (a)'s
support recovery, and Layer III adds coefficient stability. Layer
IV (span geometry of a learned pool) proves part (b). Layer V
(robust counterparts) proves part (c), and Corollary~1 closes.

\subsection*{Layer I: confining the randomness
(Lemma \ref{lem:thm2-event})}

This layer packs everything probabilistic into one event $E$ of
probability at least $1 - \alpha$: part (i) assembles $E$ from a
net bound and a union bound; part (ii) extracts the inner-product
form in which the event is consumed. Every argument after this
layer is deterministic on $E$.

\begin{lemma}[The recovery noise event]
\label{lem:thm2-event}
Let $E$ be the event: for every $m \in [M]$ and every
$B \subseteq [N]$ with $|B| \le 2k$,
\[
\| \varepsilon^{(m)} \| \le \varepsilon_{\mathrm{ns}}(\alpha)
\qquad \text{and} \qquad
\| P_{V_B}\, \varepsilon^{(m)} \| \le
\varepsilon_{\mathrm{prj}}(\alpha) .
\]
Then:
\emph{(i)} $\Pr(E) \ge 1 - \alpha$;
\emph{(ii)} on $E$, for every $m \in [M]$, every
$B \subseteq [N]$ with $|B| \le 2k$, and every coefficient vector
$v$ supported in $B$ with $y = \sum_{j} v_j s_j$,
\begin{equation}
\bigl| \langle \varepsilon^{(m)},\, y \rangle \bigr|
\;\le\; \varepsilon_{\mathrm{prj}}(\alpha)\, \| y \| .
\label{eq:thm2-nform}
\end{equation}
\end{lemma}

\begin{proof}
\emph{Step 1 (net).} Fix a subspace $V \subseteq \R^d$ with
$\dim V = d'$, chosen independently of the noise, and write
$\varepsilon$ for one sample's noise. Fix a $\tfrac12$-net
$\mathcal{N}$ of the unit sphere of $V$ with
\[
|\mathcal{N}| \;\le\; 5^{d'}
\]
\citep[\S 4.2]{vershynin2018high}.

\emph{Step 2 (net comparison).} For any unit $u \in V$ pick
$\bar u \in \mathcal{N}$ with $\| u - \bar u \| \le \tfrac12$.
Since $u - \bar u \in V$,
\[
\langle u, \varepsilon \rangle
\;=\; \langle \bar u, \varepsilon \rangle
+ \langle u - \bar u,\, P_V \varepsilon \rangle
\;\le\; \langle \bar u, \varepsilon \rangle
+ \tfrac12 \| P_V \varepsilon \| .
\]
The supremum of $\langle u, \varepsilon \rangle$ over unit
$u \in V$ equals $\| P_V \varepsilon \|$, so
\[
\| P_V \varepsilon \|
\;\le\; \max_{\bar u \in \mathcal{N}}
\langle \bar u, \varepsilon \rangle
+ \tfrac12 \| P_V \varepsilon \| ,
\qquad \text{hence} \qquad
\| P_V \varepsilon \|
\;\le\; 2 \max_{\bar u \in \mathcal{N}}
\langle \bar u, \varepsilon \rangle .
\]

\emph{Step 3 (tail at one net point).} By the Chernoff bound from
the moment-generating condition (Notation)
\citep[\S 2.5]{vershynin2018high}, for every $t > 0$,
\[
\Pr\bigl( \langle \bar u, \varepsilon \rangle \ge t \bigr)
\;\le\; e^{-t^2 / 2\sigma^2} .
\]

\emph{Step 4 (projected noise on one fixed subspace).} Set
$t = \sigma \sqrt{2 \log( 5^{d'} / \delta )}$ for
$\delta \in (0,1)$. By Steps 1 and 3,
\[
\Pr\Bigl( \max_{\bar u \in \mathcal{N}}
\langle \bar u, \varepsilon \rangle \ge t \Bigr)
\;\le\; 5^{d'} e^{- \log( 5^{d'} / \delta )}
\;=\; \delta ,
\]
and on the complementary event Step 2 gives
\[
\| P_V \varepsilon \| \;\le\; 2t
\;=\; 2\sigma \sqrt{2\bigl( d' \log 5
+ \log \tfrac{1}{\delta} \bigr)} ,
\]
a bound nondecreasing in $d'$.

\emph{Step 5 (norm bounds).} Apply Step 4 with
$V = \R^d$ ($P_V$ the identity, $d' = d$) and
$\delta = \alpha/(2M)$, once for each $m$: each bound
$\| \varepsilon^{(m)} \| \le \varepsilon_{\mathrm{ns}}(\alpha)$
fails with probability at most $\alpha/(2M)$.

\emph{Step 6 (projection bounds).} For each pair $(m, B)$, the
subspace $V_B$ is fixed: the pool is known, so $V_B$ does not
depend on the noise. Apply Step 4 with
$V = V_B$, $d' = \dim V_B \le 2k$, and
$\delta = \alpha/(2M n_{2k})$: each bound
$\| P_{V_B} \varepsilon^{(m)} \| \le
\varepsilon_{\mathrm{prj}}(\alpha)$ fails with probability at most
$\alpha/(2M n_{2k})$.

\emph{Step 7 (union bound; part (i)).} There are $M$ events in
Step 5 and at most $M n_{2k}$ in Step 6, so
\[
\Pr(E^{\mathrm{c}})
\;\le\; M \cdot \frac{\alpha}{2M}
+ M n_{2k} \cdot \frac{\alpha}{2M n_{2k}}
\;=\; \alpha .
\]

\emph{Step 8 (inner-product form; part (ii)).} On $E$, since
$y \in V_B$,
\[
\bigl| \langle \varepsilon^{(m)}, y \rangle \bigr|
\;=\; \bigl| \langle P_{V_B}\, \varepsilon^{(m)},\, y \rangle
\bigr|
\;\le\; \| P_{V_B}\, \varepsilon^{(m)} \| \, \| y \|
\;\le\; \varepsilon_{\mathrm{prj}}(\alpha)\, \| y \|
\]
by Cauchy--Schwarz and the event bound.
Inequality~(\ref{eq:thm2-nform}) is the only consequence of $E$
used in Lemmas~\ref{lem:thm2-detect} and \ref{lem:thm2-coef}: the
probabilistic layer ends here, and everything after it is
deterministic on $E$.
\end{proof}

\subsection*{Layer II: detection and counting
(Lemmas \ref{lem:thm2-detect} and \ref{lem:thm2-count})}

This layer proves part (a)'s mechanism: a candidate support
missing a true component leaves a residual too large to hide
(detection), and the certificate's usage condition then forces
every support to match (counting). The usage condition is a
hypothesis on the candidate solution, not on the data: it encodes
that the search achieved the frugality the objective pursues.
Theorem~1(a) makes it achievable (the truth satisfies it with
equality), Theorem~1(b) makes it pursued (inside the window the
penalized objective strictly prefers the true usage), and this
layer shows it is exactly what converts detection into support
equality.

\begin{lemma}[Detection of missing components]
\label{lem:thm2-detect}
On $E$: suppose a support $A \subseteq [N]$ with $|A| \le k$
misses at least one true component of sample $m$, that is,
$J_m \setminus A \ne \emptyset$. Then for every coefficient vector
$\hat c$ supported on $A$,
\[
\bigl\| x^{(m)} - S_A \hat c \bigr\|
\;\ge\; c_{\min}\, s_{\min} - \varepsilon_{\mathrm{prj}}(\alpha) .
\]
\end{lemma}

\begin{proof}
\emph{Step 1 (residual inequality).} For any
$y, \varepsilon \in \R^d$ and $b \ge 0$ with
$| \langle \varepsilon, y \rangle | \le b \| y \|$, it holds that
$\| y + \varepsilon \| \ge \| y \| - b$: for $y = 0$ this reads
$\| \varepsilon \| \ge -b$, true since $b \ge 0$; for $y \ne 0$,
Cauchy--Schwarz and bilinearity give
\[
\| y + \varepsilon \| \, \| y \|
\;\ge\; \langle y + \varepsilon,\, y \rangle
\;=\; \| y \|^2 + \langle \varepsilon, y \rangle
\;\ge\; \| y \|^2 - b \| y \| ,
\]
and dividing by $\| y \| > 0$ yields the claim.

\emph{Step 2 (mixed coefficient vector).} Define, for
$j \in [N]$,
\[
v_j \;=\; \mathbf{1}_{j \in J_m}\, c^{(m)}_j
- \mathbf{1}_{j \in A}\, \hat c_j ,
\qquad
y \;=\; \sum_{j} v_j s_j .
\]
Then $y = x_0^{(m)} - S_A \hat c$, the vector $v$ is supported in
$B = J_m \cup A$, and $|B| \le |J_m| + |A| \le 2k$.

\emph{Step 3 (coordinate mass).} Pick
$j_0 \in J_m \setminus A$. Then $v_{j_0} = c^{(m)}_{j_0} \ge
c_{\min}$, hence
\[
\sum_j v_j^2 \;\ge\; v_{j_0}^2 \;\ge\; c_{\min}^2 .
\]

\emph{Step 4 (spark).} By the spark condition on $B$
($|B| \le 2k$) and Step 3,
\[
\| y \| \;\ge\; s_{\min} \sqrt{\textstyle\sum_j v_j^2}
\;\ge\; s_{\min}\, c_{\min} .
\]

\emph{Step 5 (noise).} Lemma~\ref{lem:thm2-event}(ii) applied to
$(B, v)$ gives
\[
\bigl| \langle \varepsilon^{(m)},\, y \rangle \bigr|
\;\le\; \varepsilon_{\mathrm{prj}}(\alpha)\, \| y \| .
\]

\emph{Step 6 (conclusion).} Since
$x^{(m)} - S_A \hat c = y + \varepsilon^{(m)}$, Step 1 with
$b = \varepsilon_{\mathrm{prj}}(\alpha) \ge 0$ and Step 4 give
\[
\bigl\| x^{(m)} - S_A \hat c \bigr\|
\;\ge\; \| y \| - \varepsilon_{\mathrm{prj}}(\alpha)
\;\ge\; c_{\min}\, s_{\min} - \varepsilon_{\mathrm{prj}}(\alpha) .
\qedhere
\]
\end{proof}

\begin{lemma}[Usage counting]
\label{lem:thm2-count}
\emph{(i)} If $|J_m| \le |A_m|$ for every $m$ and
$\sum_m |A_m| \le \sum_m |J_m|$, then $|A_m| = |J_m|$ for every
$m$. \emph{(ii)} If moreover $J_m \subseteq A_m$ for every $m$,
then $A_m = J_m$ for every $m$.
\end{lemma}

\begin{proof}
\emph{Step 1 (cardinalities).} Suppose $|J_{m_0}| < |A_{m_0}|$ for
some $m_0$. Summing this strict inequality with
$|J_m| \le |A_m|$ over all other $m$,
\[
\sum_m |J_m| \;<\; \sum_m |A_m| ,
\]
contradicting the usage bound. Hence $|A_m| = |J_m|$ for every
$m$.

\emph{Step 2 (sets).} If $J_m \subseteq A_m$ and
$|A_m| = |J_m|$, a containment between finite sets of equal
cardinality is an equality: $A_m = J_m$.
\end{proof}

\subsection*{Layer III: coefficient stability
(Lemma \ref{lem:thm2-coef})}

Once supports match, the spark condition pins every concentration
to noise-floor accuracy; part (a)'s assembly follows.

\begin{lemma}[Coefficient stability]
\label{lem:thm2-coef}
On $E$: suppose sample $m$ is fitted on the true support,
$\| x^{(m)} - S_{J_m} \hat c^{(m)} \| \le
\varepsilon_{\mathrm{fit}}$. Then
\[
s_{\min}\, \| \hat c^{(m)} - c^{(m)} \|
\;\le\; \varepsilon_{\mathrm{fit}}
+ \varepsilon_{\mathrm{prj}}(\alpha) .
\]
\end{lemma}

\begin{proof}
\emph{Step 1 (signal part of the residual).} Set
\[
v_j \;=\; \mathbf{1}_{j \in J_m}
\bigl( c^{(m)}_j - \hat c^{(m)}_j \bigr),
\qquad
y \;=\; \sum_j v_j s_j
\;=\; S_{J_m} \bigl( c^{(m)} - \hat c^{(m)} \bigr) ,
\]
so that $x^{(m)} - S_{J_m} \hat c^{(m)} = y + \varepsilon^{(m)}$
and $v$ is supported in $J_m$, $|J_m| \le k \le 2k$.

\emph{Step 2 (degenerate case).} If $y = 0$, then the spark bound
of Step 5 reads
$s_{\min} \| \hat c^{(m)} - c^{(m)} \| \le \| y \| = 0$, and the
conclusion holds because
$\varepsilon_{\mathrm{fit}} + \varepsilon_{\mathrm{prj}}(\alpha)
\ge 0$.

\emph{Step 3 (Cauchy--Schwarz with the fit).} If $y \ne 0$,
\[
\langle y + \varepsilon^{(m)},\, y \rangle
\;\le\; \| y + \varepsilon^{(m)} \| \, \| y \|
\;\le\; \varepsilon_{\mathrm{fit}}\, \| y \| .
\]

\emph{Step 4 (bilinearity and noise).} By
Lemma~\ref{lem:thm2-event}(ii) applied to $(J_m, v)$,
\[
\langle y + \varepsilon^{(m)},\, y \rangle
\;=\; \| y \|^2 + \langle \varepsilon^{(m)}, y \rangle
\;\ge\; \| y \|^2 - \varepsilon_{\mathrm{prj}}(\alpha)\, \| y \| .
\]
Combining with Step 3 and dividing by $\| y \| > 0$,
\[
\| y \| \;\le\; \varepsilon_{\mathrm{fit}}
+ \varepsilon_{\mathrm{prj}}(\alpha) .
\]

\emph{Step 5 (spark conversion).} By the spark condition on $J_m$,
\[
s_{\min}\, \| \hat c^{(m)} - c^{(m)} \|
\;=\; s_{\min} \sqrt{\textstyle\sum_j v_j^2}
\;\le\; \| y \|
\;\le\; \varepsilon_{\mathrm{fit}}
+ \varepsilon_{\mathrm{prj}}(\alpha) .
\qedhere
\]
\end{proof}

\begin{proof}[Proof of part (a)]
\emph{Step 1 (the event).} By Lemma~\ref{lem:thm2-event}(i),
$\Pr(E) \ge 1 - \alpha$. Fix a realization in $E$; the remaining
steps are deterministic and hold simultaneously for every
candidate solution passing (C1)--(C2), since $E$ does not
depend on the candidate.

\emph{Step 2 (containment).} Fix $m$ and suppose
$J_m \not\subseteq A_m$, so $J_m \setminus A_m \ne \emptyset$. By
(C1), Lemma~\ref{lem:thm2-detect}, and the right edge of
(\ref{eq:thm2-wfix}),
\[
\varepsilon_{\mathrm{fit}}
\;\ge\; \bigl\| x^{(m)} - S_{A_m} \hat c^{(m)} \bigr\|
\;\ge\; c_{\min}\, s_{\min} - \varepsilon_{\mathrm{prj}}(\alpha)
\;>\; \varepsilon_{\mathrm{fit}} ,
\]
a contradiction. Hence $J_m \subseteq A_m$ for every $m$.

\emph{Step 3 (counting).} Step 2 and (C2) are the hypotheses of
Lemma~\ref{lem:thm2-count}(ii), so $A_m = J_m$ for every $m$:
part (i). Taking unions,
\[
\bigcup_m A_m \;=\; \bigcup_m J_m \;=\; J_{\mathrm{act}} ,
\qquad
\Bigl| \bigcup_m A_m \Bigr| \;=\; K^{\ast} :
\]
part (ii).

\emph{Step 4 (coefficients).} With $A_m = J_m$, condition (C1) is
the hypothesis of Lemma~\ref{lem:thm2-coef}, so
\[
\| \hat c^{(m)} - c^{(m)} \|
\;\le\; \frac{\varepsilon_{\mathrm{fit}}
+ \varepsilon_{\mathrm{prj}}(\alpha)}{s_{\min}} :
\]
part (iii).

\emph{Step 5 (achievability).} On $E$, by the left edge of
(\ref{eq:thm2-wfix}),
\[
\bigl\| x^{(m)} - S_{J_m} c^{(m)} \bigr\|
\;=\; \| \varepsilon^{(m)} \|
\;\le\; \varepsilon_{\mathrm{ns}}(\alpha)
\;\le\; \varepsilon_{\mathrm{fit}} ,
\]
and $|J_m| \le k$, so the true configuration passes (C1); its
total usage equals $\sum_m |J_m|$, so it meets (C2) with
equality.
\end{proof}

\subsection*{Layer IV: span geometry of a learned pool
(Lemmas \ref{lem:thm2-spanint}--\ref{lem:thm2-anchor})}

This layer proves part (b)'s mechanism, noiseless: tight usage
forces every used candidate component into the true spans, and
singleton-intersection patterns pin each component individually,
producing the relabeling $\pi$; non-negativity resolves the signs
at the last step. Wherever a lemma below assumes linearly
independent profiles, the assumption is supplied by the spark
condition: the index sets involved have at most $2k$ elements, and
a vanishing nontrivial combination over such a set would force
$\sigma_{\min} = 0 < s_{\min}$, contradicting
Definition~\ref{def:spark}.

\begin{lemma}[Span intersection along an independent family]
\label{lem:thm2-spanint}
Let $F, F'$ be finite index sets and let
$(v_\iota)_{\iota \in F \cup F'}$ be linearly independent vectors
in $\R^d$. Then
\[
\mathrm{span}(v_\iota : \iota \in F)
\,\cap\,
\mathrm{span}(v_\iota : \iota \in F')
\;=\;
\mathrm{span}(v_\iota : \iota \in F \cap F') .
\]
\end{lemma}

\begin{proof}
\emph{Step 1 ($\supseteq$).} Monotonicity of the span:
$F \cap F' \subseteq F$ and $F \cap F' \subseteq F'$.

\emph{Step 2 ($\subseteq$).} Suppose
\[
w \;=\; \sum_{\iota \in F} a_\iota v_\iota
\;=\; \sum_{\iota \in F'} b_\iota v_\iota .
\]
Subtracting,
\[
\sum_{\iota \in F \setminus F'} a_\iota v_\iota
+ \sum_{\iota \in F \cap F'} ( a_\iota - b_\iota ) v_\iota
- \sum_{\iota \in F' \setminus F} b_\iota v_\iota
\;=\; 0 ,
\]
a vanishing combination over $F \cup F'$, so independence forces
$a_\iota = 0$ on $F \setminus F'$, $b_\iota = 0$ on
$F' \setminus F$, and $a_\iota = b_\iota$ on $F \cap F'$. Hence
$w = \sum_{\iota \in F \cap F'} a_\iota v_\iota$.
\end{proof}

\begin{lemma}[Tight usage and exact spans]
\label{lem:thm2-tight}
Assume $\sigma = 0$, the spark condition on the true pool
(Definition~\ref{def:spark}), and (C3)--(C5) and (C7). Then for
every $m$,
\[
|A_m| \;=\; |J_m| ,
\qquad
\mathrm{span}\bigl( \hat S_{A_m} \bigr) \;=\; V_{J_m} .
\]
In particular every used candidate profile lies exactly in the
sample's true span, and the used profiles of sample $m$ span it.
\end{lemma}

\begin{proof}
\emph{Step 1 (row-wise lower bound).} For each $m$, by (C7),
\[
|A_m| \;\ge\; \dim \mathrm{span}\bigl( \hat S_{A_m} \bigr)
\;\ge\; \dim \bigl( \mathrm{span}( \hat S_{A_m} ) \cap V_{J_m}
\bigr)
\;=\; |J_m| .
\]

\emph{Step 2 (counting).} Step 1 and (C4) are the hypotheses of
Lemma~\ref{lem:thm2-count}(i), so $|A_m| = |J_m|$ for every $m$.

\emph{Step 3 (first rank squeeze).} By (C5) the columns of
$\hat S_{A_m}$ are linearly independent ($|A_m| \le k \le 2k$), so
by Step 2 and (C7),
\[
\dim \mathrm{span}\bigl( \hat S_{A_m} \bigr)
\;=\; |A_m|
\;=\; |J_m|
\;=\; \dim \bigl( \mathrm{span}( \hat S_{A_m} ) \cap V_{J_m}
\bigr) .
\]
The intersection is a subspace of
$\mathrm{span}(\hat S_{A_m})$ of the same finite dimension, hence
equals it; therefore
$\mathrm{span}(\hat S_{A_m}) \subseteq V_{J_m}$.

\emph{Step 4 (second rank squeeze).} Independence of the true
profiles on $J_m$ (spark condition, $|J_m| \le 2k$) gives
$\dim V_{J_m} = |J_m|$, equal by Step 3 to the dimension of the
contained subspace $\mathrm{span}(\hat S_{A_m})$; the containment
of Step 3 is therefore an equality.
\end{proof}

\begin{lemma}[Anchoring]
\label{lem:thm2-anchor}
Assume $\sigma = 0$, the spark condition on the true pool, and
(C3)--(C7). Then for every active $i \in J_{\mathrm{act}}$ there
is a pool index $\pi(i)$ with
$\hat s_{\pi(i)} \in \mathrm{span}(s_i)$; writing
$\hat s_{\pi(i)} = \tau_i s_i$, the scalar satisfies
$\tau_i \in \{\pm 1\}$; and the map $\pi$ is injective.
\end{lemma}

\begin{proof}
\emph{Step 1 (the pair and the line).} Fix $i$ and take $m, m'$
with $J_m \cap J_{m'} = \{i\}$ (C6). The true profiles indexed by
$J_m \cup J_{m'}$ are independent
($|J_m \cup J_{m'}| \le 2k$, spark condition), so
Lemma~\ref{lem:thm2-spanint} gives
\[
V_{J_m} \cap V_{J_{m'}} \;=\; \mathrm{span}(s_i) ,
\]
a subspace of dimension $1$ ($s_i \ne 0$ by unit norm).

\emph{Step 2 (rank bookkeeping).} Set $U = A_m \cup A_{m'}$. By
Lemma~\ref{lem:thm2-tight},
$|U| \le |A_m| + |A_{m'}| = |J_m| + |J_{m'}| \le 2k$, so (C5)
makes the columns of $\hat S_U$ independent and
\[
|U| \;=\; \dim \mathrm{span}\bigl( \hat S_U \bigr)
\;=\; \dim \bigl( V_{J_m} + V_{J_{m'}} \bigr) ,
\]
the second equality because
$\mathrm{span}(\hat S_U) = \mathrm{span}(\hat S_{A_m}) +
\mathrm{span}(\hat S_{A_{m'}}) = V_{J_m} + V_{J_{m'}}$
(Lemma~\ref{lem:thm2-tight}). The dimension formula with Step 1
gives
\[
\dim \bigl( V_{J_m} + V_{J_{m'}} \bigr)
\;=\; |J_m| + |J_{m'}| - 1 ,
\]
and hence, by inclusion--exclusion and
Lemma~\ref{lem:thm2-tight},
\[
| A_m \cap A_{m'} |
\;=\; |A_m| + |A_{m'}| - |U|
\;=\; 1 :
\]
exactly one shared index $j^{\ast}$.

\emph{Step 3 (location and sign).} The column
$\hat s_{j^{\ast}}$ belongs to both spans, so by
Lemma~\ref{lem:thm2-tight} and Step 1,
\[
\hat s_{j^{\ast}}
\;\in\; \mathrm{span}\bigl( \hat S_{A_m} \bigr) \cap
\mathrm{span}\bigl( \hat S_{A_{m'}} \bigr)
\;=\; V_{J_m} \cap V_{J_{m'}}
\;=\; \mathrm{span}(s_i) ,
\]
hence $\hat s_{j^{\ast}} = a\, s_i$ for a scalar $a$; unit norms
give $|a| = 1$, and we set $\tau_i = a \in \{\pm 1\}$ and
$\pi(i) = j^{\ast}$.

\emph{Step 4 (independence of the choice).} The slot is unique:
two distinct pool columns inside the line $\mathrm{span}(s_i)$
would form a linearly dependent pair, contradicting (C5) applied
to that pair ($2 \le 2k$). Hence $\pi(i)$ does not depend on the
pair $(m, m')$ chosen in Step 1.

\emph{Step 5 (injectivity).} If $\pi(i) = \pi(i')$ with
$i \ne i'$, then $\tau_i s_i = \tau_{i'} s_{i'}$, that is,
$s_{i'} = ( \tau_{i'}^{-1} \tau_i )\, s_i$: a member of the pair
$\{ s_i, s_{i'} \}$, independent by the spark condition
($2 \le 2k$), written as a multiple of the other. This is
impossible, so $\pi$ is injective.
\end{proof}

\begin{proof}[Proof of part (b)]
If $J_{\mathrm{act}} = \emptyset$, then (C4) reads
$\sum_m |A_m| \le 0$, every $A_m$ is empty, and all claims hold
with the empty relabeling; assume $J_{\mathrm{act}} \ne \emptyset$
from now on, whence $k \ge 1$. Let $\pi$ and $\tau_i \in \{\pm1\}$
be as in Lemma~\ref{lem:thm2-anchor}.

\emph{Step 1 (membership).} Fix $m$ and $i \in J_m$, and suppose
$\pi(i) \notin A_m$. The pool subset $\{\pi(i)\} \cup A_m$ has
cardinality $|A_m| + 1 \le k + 1 \le 2k$, hence its columns are
independent by (C5). All of them lie in $V_{J_m}$: the columns of
$A_m$ by Lemma~\ref{lem:thm2-tight}, and
$\hat s_{\pi(i)} = \tau_i s_i \in V_{J_m}$ since $i \in J_m$. An
independent family inside $V_{J_m}$ has at most
$\dim V_{J_m} = |J_m|$ members, so by
Lemma~\ref{lem:thm2-tight},
\[
|A_m| + 1 \;\le\; |J_m| \;=\; |A_m| ,
\]
which is absurd. Hence $\pi(i) \in A_m$ for every $i \in J_m$.

\emph{Step 2 (support identity).} Step 1 gives
$\pi(J_m) \subseteq A_m$, and by injectivity of $\pi$ and
Lemma~\ref{lem:thm2-tight},
\[
| \pi(J_m) | \;=\; |J_m| \;=\; |A_m| ,
\]
so $A_m = \pi(J_m)$: part (ii).

\emph{Step 3 (coefficient extraction).} Reindex the fit (C3) over
$A_m = \pi(J_m)$ ($\pi$ injective) and substitute
$\hat s_{\pi(i)} = \tau_i s_i$:
\[
\sum_{i \in J_m} \bigl( \hat c^{(m)}_{\pi(i)}\, \tau_i \bigr)\,
s_i
\;=\; x_0^{(m)}
\;=\; \sum_{i \in J_m} c^{(m)}_i\, s_i .
\]
The family $(s_i)_{i \in J_m}$ is independent (spark condition),
so the coefficients coincide:
\[
\hat c^{(m)}_{\pi(i)}\, \tau_i \;=\; c^{(m)}_i
\qquad \text{for every } i \in J_m .
\]

\emph{Step 4 (sign resolution).} For each active $i$ pick a
sample $m$ with $i \in J_m$ (the (C6) pair provides one). There
$c^{(m)}_i > 0$, and $\hat c^{(m)}_{\pi(i)} > 0$ because
$\pi(i) \in A_m$ (Step 1) and used coefficients are strictly
positive. Step 3 then forces $\tau_i = +1$, so
$\hat s_{\pi(i)} = s_i$: part (i); and Step 3 becomes
$\hat c^{(m)}_{\pi(i)} = c^{(m)}_i$ for every $m$ and
$i \in J_m$: part (iv). Only positivity of the coefficients is
used; $c_{\min}$ plays no quantitative role.

\emph{Step 5 (count).} By Step 2 and injectivity of $\pi$,
\[
\bigcup_m A_m \;=\; \pi\Bigl( \bigcup_m J_m \Bigr)
\;=\; \pi\bigl( J_{\mathrm{act}} \bigr) ,
\qquad
\Bigl| \bigcup_m A_m \Bigr| \;=\; K^{\ast} :
\]
part (iii).
\end{proof}

\subsection*{Layer V: robust counterparts
(Lemmas \ref{lem:thm2-crude}--\ref{lem:thm2-k1})}

This layer proves part (c): each exact-geometry fact of Layer IV
degrades continuously with the fit error (a crude coefficient
bound, a Weyl dimension count, a two-subspace pinch, and span
proximity from coefficient-diverse fits). Conjecture~R enters
exactly once, to supply the span-proximity hypothesis for every
active component; the $1$-sparse case closes without it.

\begin{lemma}[Crude coefficient bound]
\label{lem:thm2-crude}
Assume (C8) and (C10), and suppose the signal-level fit
$\| x_0^{(m)} - \hat S_{A_m} \hat c^{(m)} \| \le \bar\varepsilon$
holds for sample $m$. Then
$\| \hat c^{(m)} \| \le \hat C = ( k\, c_{\max} +
\bar\varepsilon ) / \hat s_{\min}$.
\end{lemma}

\begin{proof}
By (C10) with $|A_m| \le k \le 2k$, the triangle inequality, and
$\| x_0^{(m)} \| \le \sum_{i \in J_m} c^{(m)}_i \| s_i \| \le
k\, c_{\max}$,
\[
\hat s_{\min}\, \| \hat c^{(m)} \|
\;\le\; \sigma_{\min}\bigl( \hat S_{A_m} \bigr)\,
\| \hat c^{(m)} \|
\;\le\; \bigl\| \hat S_{A_m} \hat c^{(m)} \bigr\|
\;\le\; \| x_0^{(m)} \| + \bar\varepsilon
\;\le\; k\, c_{\max} + \bar\varepsilon .
\qedhere
\]
\end{proof}

\begin{lemma}[Robust dimension count]
\label{lem:thm2-dim}
Let $B \subseteq [N]$ with
$\sigma_{\min}(\hat S_B) \ge \hat s_{\min}$, let
$W \subseteq \R^d$ be a subspace of dimension $D$, and suppose
every column satisfies $\dist(\hat s_j, W) \le \eta$ for
$j \in B$. If $\sqrt{|B|}\, \eta < \hat s_{\min}$, then
$|B| \le D$.
\end{lemma}

\begin{proof}
\emph{Step 1 (rank).} Suppose $|B| > D$. The projected matrix
$P_W \hat S_B$ has rank at most $D < |B|$, so its smallest
singular value (of $|B|$ many) vanishes:
$\sigma_{\min}( P_W \hat S_B ) = 0$.

\emph{Step 2 (column-wise error).}
\[
\bigl\| \hat S_B - P_W \hat S_B \bigr\|_F^2
\;=\; \sum_{j \in B} \| \hat s_j - P_W \hat s_j \|^2
\;=\; \sum_{j \in B} \dist( \hat s_j, W )^2
\;\le\; |B|\, \eta^2 .
\]

\emph{Step 3 (Weyl).} Singular values are perturbed by at most
the operator norm of the difference
\citep[\S 7.3]{hornjohnson2013matrix}, so by Steps 1--2,
\[
\hat s_{\min}
\;\le\; \sigma_{\min}\bigl( \hat S_B \bigr)
\;\le\; \sigma_{\min}\bigl( P_W \hat S_B \bigr)
+ \bigl\| \hat S_B - P_W \hat S_B \bigr\|_{\mathrm{op}}
\;\le\; 0 + \sqrt{|B|}\, \eta
\;<\; \hat s_{\min} ,
\]
a contradiction. Hence $|B| \le D$.
\end{proof}

\begin{lemma}[Two-subspace pinch]
\label{lem:thm2-pinch}
Let $u$ be a unit vector with $\dist(u, V_J) \le \delta_1$ and
$\dist(u, V_{J'}) \le \delta_2$, where $J \cap J' = \{i\}$,
$|J \cup J'| \le 2k$, and
$\sigma_{\min}( S_{J \cup J'} ) \ge s_{\min}$. Then
\[
\dist\bigl( u,\ \mathrm{span}(s_i) \bigr)
\;\le\; \delta_1
+ \sqrt{2k}\, \frac{\delta_1 + \delta_2}{s_{\min}} .
\]
\end{lemma}

\begin{proof}
\emph{Step 1 (two decompositions).} Take the least-squares
decompositions
\[
u \;=\; S_J a + r_1 ,
\qquad
u \;=\; S_{J'} b + r_2 ,
\qquad
\| r_1 \| \le \delta_1 , \quad \| r_2 \| \le \delta_2 .
\]

\emph{Step 2 (difference through the spark).} Subtracting, the
coefficient vector $v$ on $J \cup J'$ (entries $a_j$ on
$J \setminus \{i\}$, $a_i - b_i$ at $i$, $-b_j$ on
$J' \setminus \{i\}$) satisfies
$S_{J \cup J'}\, v = r_2 - r_1$, so
\[
s_{\min}\, \| v \|
\;\le\; \bigl\| S_{J \cup J'}\, v \bigr\|
\;\le\; \delta_1 + \delta_2 ,
\qquad \text{hence} \qquad
\bigl\| a_{J \setminus \{i\}} \bigr\|
\;\le\; \| v \|
\;\le\; \frac{\delta_1 + \delta_2}{s_{\min}} .
\]

\emph{Step 3 (drop the off-$i$ part).} By Step 1,
$u - a_i s_i = S_{J \setminus \{i\}}\, a_{J \setminus \{i\}} +
r_1$. A matrix with at most $2k$ unit columns has operator norm
at most $\sqrt{2k}$, so by Step 2,
\[
\dist\bigl( u, \mathrm{span}(s_i) \bigr)
\;\le\; \| u - a_i s_i \|
\;\le\; \sqrt{2k}\,
\bigl\| a_{J \setminus \{i\}} \bigr\| + \delta_1
\;\le\; \delta_1
+ \sqrt{2k}\, \frac{\delta_1 + \delta_2}{s_{\min}} .
\qedhere
\]
\end{proof}

\begin{lemma}[Span proximity from coefficient-diverse fits]
\label{lem:thm2-serve}
Fix a pattern $J$ with $|J| = \varkappa \le k$ and
$\sigma_{\min}(S_J) \ge s$ for some $s > 0$, and a candidate set
$A \subseteq [N]$ with $|A| = \varkappa$. Suppose samples
$m_1, \dots, m_\varkappa$ share the pattern and the candidate
set, $J_{m_t} = J$ and $A_{m_t} = A$ for every $t$; that their
coefficient matrix
$C = [\, c^{(m_1)} \cdots c^{(m_\varkappa)} \,]$ has
$\sigma_{\min}(C) \ge \gamma > 0$; and that each meets the
signal-level fit
$\| S_J c^{(m_t)} - \hat S_A \hat c^{(m_t)} \| \le
\bar\varepsilon$. If
$\sqrt{\varkappa}\, \bar\varepsilon < s\, \gamma$, then for every
$j \in A$,
\[
\dist\bigl( \hat s_j,\, V_J \bigr)
\;\le\;
\frac{\varkappa\, \bar\varepsilon}
{s\, \gamma - \sqrt{\varkappa}\, \bar\varepsilon} .
\]
\end{lemma}

\begin{proof}
\emph{Step 1 (stacked fit).} Write
$\hat C = [\, \hat c^{(m_1)} \cdots \hat c^{(m_\varkappa)} \,]$
(rows indexed by $A$). Summing the squared fits,
\[
\bigl\| S_J C - \hat S_A \hat C \bigr\|_F^2
\;=\; \sum_{t=1}^{\varkappa}
\bigl\| S_J c^{(m_t)} - \hat S_A \hat c^{(m_t)} \bigr\|^2
\;\le\; \varkappa\, \bar\varepsilon^2 .
\]

\emph{Step 2 (projection off the span).} Write
$Q_J = I - P_{V_J}$. Since $Q_J S_J C = 0$ and
$\| Q_J \|_{\mathrm{op}} \le 1$,
\[
\bigl\| Q_J \hat S_A \hat C \bigr\|_F
\;=\; \bigl\| Q_J \bigl( \hat S_A \hat C - S_J C \bigr)
\bigr\|_F
\;\le\; \sqrt{\varkappa}\, \bar\varepsilon .
\]

\emph{Step 3 ($\hat C$ is well conditioned).} First,
\[
\sigma_{\min}\bigl( S_J C \bigr)
\;\ge\; \sigma_{\min}( S_J )\, \sigma_{\min}( C )
\;\ge\; s\, \gamma ,
\]
and singular values are perturbed by at most the operator norm of
the difference \citep[Theorem~4.11, p.~204]{stewartsun1990matrix}, so by Step 1,
\[
\sigma_{\min}\bigl( \hat S_A \hat C \bigr)
\;\ge\; \sigma_{\min}\bigl( S_J C \bigr)
- \bigl\| S_J C - \hat S_A \hat C \bigr\|_{\mathrm{op}}
\;\ge\; s\, \gamma - \sqrt{\varkappa}\, \bar\varepsilon
\;>\; 0 .
\]
Second, since $\hat S_A$ has $\varkappa$ unit columns,
$\| \hat S_A \|_{\mathrm{op}} \le \sqrt{\varkappa}$ and
\[
\sigma_{\min}\bigl( \hat S_A \hat C \bigr)
\;\le\; \| \hat S_A \|_{\mathrm{op}}\,
\sigma_{\min}\bigl( \hat C \bigr)
\;\le\; \sqrt{\varkappa}\, \sigma_{\min}\bigl( \hat C \bigr) .
\]
Combining, $\hat C$ is invertible with
\[
\bigl\| \hat C^{-1} \bigr\|_{\mathrm{op}}
\;=\; \frac{1}{\sigma_{\min}( \hat C )}
\;\le\; \frac{\sqrt{\varkappa}}
{s\, \gamma - \sqrt{\varkappa}\, \bar\varepsilon} .
\]

\emph{Step 4 (column bound).} For each $j \in A$, with $e_j$ the
corresponding coordinate vector, by Steps 2--3,
\[
\begin{aligned}
\dist\bigl( \hat s_j, V_J \bigr)
\;=\; \bigl\| Q_J \hat S_A e_j \bigr\|
&\;=\; \bigl\| \bigl( Q_J \hat S_A \hat C \bigr)
\bigl( \hat C^{-1} e_j \bigr) \bigr\|
\\
&\;\le\; \bigl\| Q_J \hat S_A \hat C \bigr\|_{\mathrm{op}}\,
\bigl\| \hat C^{-1} \bigr\|_{\mathrm{op}}
\;\le\;
\frac{\varkappa\, \bar\varepsilon}
{s\, \gamma - \sqrt{\varkappa}\, \bar\varepsilon} .
\end{aligned}
\]
This is the quantitative reason cancellation cannot persist: a
candidate set can cancel its out-of-span parts against one
coefficient vector, but not against $\varkappa$ linearly
independent ones; the stacked constraint of Step 2 controls the
whole matrix, hence each column.
\end{proof}

\begin{lemma}[Conjecture~R holds for $k = 1$]
\label{lem:thm2-k1}
Let $k = 1$, let the candidate solution pass (C8)--(C11) with
(C6), and suppose the signal-level fit
$\| x_0^{(m)} - \hat S_{A_m} \hat c^{(m)} \| \le \bar\varepsilon$
holds with $\bar\varepsilon < c_{\min}$. Then the conclusion of
Conjecture~R holds with quantile
$\gamma_{\mathrm{c}} = c_{\min}$; indeed
\[
\dist\bigl( \hat s_j,\, V_{J_m} \bigr)
\;\le\; \frac{\bar\varepsilon}{c_{\min} - \bar\varepsilon}
\qquad \text{for every } m \text{ and the single } j \in A_m .
\]
\end{lemma}

\begin{proof}
\emph{Step 1 (tight cardinalities).} (C11) and (C9) are the
hypotheses of Lemma~\ref{lem:thm2-count}(i), so
$|A_m| = |J_m| = 1$ for every $m$ with $J_m \ne \emptyset$.

\emph{Step 2 (each sample is its own witness).} Fix $m$ with
$J_m = \{i\}$. The pair $(J_m, A_m)$ is
$c_{\min}$-spanned by the single sample $m$ itself: the
coefficient matrix is the $1 \times 1$ matrix
$C = ( c^{(m)}_i )$ with
$\sigma_{\min}(C) = c^{(m)}_i \ge c_{\min}$, and
$\sigma_{\min}( S_{J_m} ) = \| s_i \| = 1$. In particular, for
every active $i$, any (C6) pair $(m, m')$ satisfies the
conclusion of Conjecture~R with
$\gamma_{\mathrm{c}} = c_{\min}$.

\emph{Step 3 (span proximity).} Lemma~\ref{lem:thm2-serve} with
$\varkappa = 1$, $s = 1$, $\gamma = c_{\min}$, and the hypothesis
$\bar\varepsilon < c_{\min}$ gives, for the single $j \in A_m$,
\[
\dist\bigl( \hat s_j, V_{J_m} \bigr)
\;\le\; \frac{\bar\varepsilon}{c_{\min} - \bar\varepsilon} .
\qedhere
\]
\end{proof}

\begin{proof}[Proof of part (c)]
\emph{Step 1 (signal-level fit on the event).} By
Lemma~\ref{lem:thm2-event}(i), $\Pr(E) \ge 1 - \alpha$; fix a
realization in $E$. Only the norm bounds
$\| \varepsilon^{(m)} \| \le \varepsilon_{\mathrm{ns}}(\alpha)$ of
$E$ are used in this part: the learned directions depend on the
noise, so the fixed-subspace bounds of $E$ do not apply to them.
By (C8) and the triangle inequality, for every $m$,
\[
\bigl\| x_0^{(m)} - \hat S_{A_m} \hat c^{(m)} \bigr\|
\;\le\; \bigl\| x^{(m)} - \hat S_{A_m} \hat c^{(m)} \bigr\|
+ \| \varepsilon^{(m)} \|
\;\le\; \varepsilon_{\mathrm{fit}}
+ \varepsilon_{\mathrm{ns}}(\alpha)
\;=\; \bar\varepsilon .
\]

\emph{Step 2 (tight cardinalities).} (C11) and (C9) are the
hypotheses of Lemma~\ref{lem:thm2-count}(i), so
$|A_m| = |J_m|$ for every $m$.

\emph{Step 3 (anchoring).} Fix active $i$ and let $(m, m')$ be
the pair provided by Conjecture~R, so that $(J_m, A_m)$ and
$(J_{m'}, A_{m'})$ are $\gamma_{\mathrm{c}}$-spanned. For
$(J_m, A_m)$, with $\varkappa = |J_m| = |A_m|$ (Step 2): the
witnessing samples obey the Step-1 fit, the spark condition gives
$\sigma_{\min}( S_{J_m} ) \ge s_{\min}$, and the first condition
of (\ref{eq:thm2-wrob}) gives
$\sqrt{\varkappa}\, \bar\varepsilon \le \sqrt{k}\, \bar\varepsilon
< s_{\min} \gamma_{\mathrm{c}}$. Lemma~\ref{lem:thm2-serve} (with
$s = s_{\min}$, $\gamma = \gamma_{\mathrm{c}}$) therefore yields,
for every $j \in A_m$,
\[
\dist\bigl( \hat s_j, V_{J_m} \bigr)
\;\le\;
\frac{\varkappa\, \bar\varepsilon}
{s_{\min} \gamma_{\mathrm{c}}
- \sqrt{\varkappa}\, \bar\varepsilon}
\;\le\;
\frac{k\, \bar\varepsilon}
{s_{\min} \gamma_{\mathrm{c}} - \sqrt{k}\, \bar\varepsilon}
\;=\; f ,
\]
the second inequality because the bound is nondecreasing in
$\varkappa$ and $\varkappa \le k$; likewise
$\dist( \hat s_j, V_{J_{m'}} ) \le f$ for every $j \in A_{m'}$.

Set $U = A_m \cup A_{m'}$ and $W = V_{J_m \cup J_{m'}}$. Every
column indexed by $U$ lies within $f$ of $W$ (as
$V_{J_m}, V_{J_{m'}} \subseteq W$), and independence of the true
profiles on $J_m \cup J_{m'}$ (spark condition,
$|J_m \cup J_{m'}| \le 2k$) with Lemma~\ref{lem:thm2-spanint}
gives $V_{J_m} \cap V_{J_{m'}} = \mathrm{span}(s_i)$ and
\[
\dim W \;=\; |J_m| + |J_{m'}| - 1 \;=:\; D .
\]
By (C10), $\sigma_{\min}( \hat S_U ) \ge \hat s_{\min}$
($|U| \le 2k$), and the second condition of (\ref{eq:thm2-wrob})
gives $\sqrt{|U|}\, f \le \sqrt{2k}\, f < \hat s_{\min}$;
Lemma~\ref{lem:thm2-dim} with $\eta = f$ yields $|U| \le D$.
Inclusion--exclusion with Step 2 then gives
\[
| A_m \cap A_{m'} |
\;=\; |A_m| + |A_{m'}| - |U|
\;\ge\; |J_m| + |J_{m'}| - D
\;=\; 1 .
\]
Pick $j^{\ast} \in A_m \cap A_{m'}$; it lies within $f$ of both
$V_{J_m}$ and $V_{J_{m'}}$, so Lemma~\ref{lem:thm2-pinch}
(with $\delta_1 = \delta_2 = f$) gives
\[
\dist\bigl( \hat s_{j^{\ast}},\, \mathrm{span}(s_i) \bigr)
\;\le\; f + \sqrt{2k}\, \frac{2f}{s_{\min}}
\;=\; f' .
\]
Set $\pi(i) = j^{\ast}$.

\emph{Step 4 (injectivity).} Suppose $\pi(i) = \pi(i') = j$ with
$i \ne i'$. Step 3 provides scalars $a, a'$ with
$\| \hat s_j - a\, s_i \| \le f'$ and
$\| \hat s_j - a'\, s_{i'} \| \le f'$, so
$\| a\, s_i - a'\, s_{i'} \| \le 2 f'$, and the spark condition
on the pair $\{i, i'\}$ gives
\[
s_{\min} \sqrt{a^2 + a'^2} \;\le\; 2 f' ,
\qquad \text{hence} \qquad
|a| \;\le\; \frac{2 f'}{s_{\min}} .
\]
Then, by the third condition of (\ref{eq:thm2-wrob}),
\[
1 \;=\; \| \hat s_j \|
\;\le\; |a| + \| \hat s_j - a\, s_i \|
\;\le\; \frac{2 f'}{s_{\min}} + f'
\;=\; f' \Bigl( 1 + \frac{2}{s_{\min}} \Bigr)
\;<\; 1 ,
\]
a contradiction. Hence $\pi$ is injective.

\emph{Step 5 (supports by floor exclusion).} Fix $m$. For
$i \in J_m$ write the orthogonal decomposition
\[
\hat s_{\pi(i)} \;=\; a_i s_i + e_i ,
\qquad e_i \perp s_i ,
\qquad
\| e_i \| = \dist\bigl( \hat s_{\pi(i)}, \mathrm{span}(s_i)
\bigr) \le f' ,
\]
so $a_i^2 = 1 - \| e_i \|^2$ and
$|a_i| \ge \sqrt{1 - f'^2} \ge 1 - f' > 0$ (the third condition
of (\ref{eq:thm2-wrob}) forces $f' < 1$). The sign of $a_i$ is
fixed only in Step 6 and is not used here. Substituting
$s_i = ( \hat s_{\pi(i)} - e_i ) / a_i$,
\[
x_0^{(m)} \;=\; \hat S_{\pi(J_m)}\, \tilde c + r ,
\qquad
\tilde c_i = \frac{c^{(m)}_i}{a_i} ,
\qquad
r = - \sum_{i \in J_m} \tilde c_i\, e_i ,
\qquad
\| r \| \;\le\; \frac{k\, c_{\max} f'}{1 - f'} \;=:\; \bar r .
\]
With Step 1 and the triangle inequality,
\[
\bigl\| \hat S_{A_m} \hat c^{(m)}
- \hat S_{\pi(J_m)}\, \tilde c \bigr\|
\;\le\; \bar\varepsilon + \bar r .
\]
The difference is $\hat S_B v$ on $B = A_m \cup \pi(J_m)$,
$|B| \le 2k$ (Steps 2 and 4), where $v$ carries
$\hat c^{(m)}_j$ on $A_m \setminus \pi(J_m)$,
$\hat c^{(m)}_j - \tilde c_{\pi^{-1}(j)}$ on the intersection,
and $- \tilde c_{\pi^{-1}(j)}$ on $\pi(J_m) \setminus A_m$. By
(C10),
\[
\| v \| \;\le\; \frac{\bar\varepsilon + \bar r}{\hat s_{\min}} .
\]
If some $j \in A_m \setminus \pi(J_m)$ existed, its entry would
give, by the (C8) floor and the fifth condition of
(\ref{eq:thm2-wrob}),
\[
\vartheta \;\le\; \hat c^{(m)}_j \;=\; |v_j| \;\le\; \| v \|
\;\le\; \frac{\bar\varepsilon + k\, c_{\max} f' / (1 - f')}
{\hat s_{\min}}
\;<\; \vartheta ,
\]
a contradiction. Hence $A_m \subseteq \pi(J_m)$, and
$|A_m| = |J_m| = |\pi(J_m)|$ (Steps 2 and 4) forces
$A_m = \pi(J_m)$: part (ii). Taking unions, by injectivity,
\[
\bigcup_m A_m \;=\; \pi\bigl( J_{\mathrm{act}} \bigr) ,
\qquad
\Bigl| \bigcup_m A_m \Bigr| \;=\; K^{\ast} :
\]
part (iii).

\emph{Step 6 (coefficients and signs).} Fix $m$. Substituting
$A_m = \pi(J_m)$ and $\hat s_{\pi(i)} = a_i s_i + e_i$ into the
Step-1 fit,
\[
x_0^{(m)} - \hat S_{A_m} \hat c^{(m)}
\;=\; S_{J_m}\bigl( c^{(m)} - \tilde a \bigr)
- \sum_{i \in J_m} \hat c^{(m)}_{\pi(i)}\, e_i ,
\qquad
\tilde a_i = a_i\, \hat c^{(m)}_{\pi(i)} .
\]
By Step 1, the triangle inequality, Cauchy--Schwarz, and
Lemma~\ref{lem:thm2-crude},
\[
\bigl\| S_{J_m}\bigl( c^{(m)} - \tilde a \bigr) \bigr\|
\;\le\; \bar\varepsilon
+ \sum_{i \in J_m} \hat c^{(m)}_{\pi(i)}\, \| e_i \|
\;\le\; \bar\varepsilon
+ \sqrt{k}\, \bigl\| \hat c^{(m)} \bigr\|\, f'
\;\le\; \bar\varepsilon + \sqrt{k}\, \hat C f' ,
\]
so the spark condition on $J_m$ gives, entrywise,
\[
\bigl| c^{(m)}_i - a_i\, \hat c^{(m)}_{\pi(i)} \bigr|
\;\le\; \bigl\| c^{(m)} - \tilde a \bigr\|
\;\le\; \frac{\bar\varepsilon + \sqrt{k}\, \hat C f'}{s_{\min}}
\;=\; \kappa_{\mathrm{c}} .
\]
By the fourth condition of (\ref{eq:thm2-wrob}),
\[
a_i\, \hat c^{(m)}_{\pi(i)}
\;\ge\; c^{(m)}_i - \kappa_{\mathrm{c}}
\;\ge\; c_{\min} - \kappa_{\mathrm{c}}
\;>\; 0 ,
\]
and $\hat c^{(m)}_{\pi(i)} \ge \vartheta > 0$, so $a_i > 0$; with
Step 5, $a_i \in [1 - f', 1]$, and
$\langle \hat s_{\pi(i)}, s_i \rangle = a_i > 0$ together with
Step 3 proves part (i). Finally, using
$\hat c^{(m)}_{\pi(i)} \le \| \hat c^{(m)} \| \le \hat C$
(Lemma~\ref{lem:thm2-crude}) and $1 - a_i \le f'$,
\[
\bigl| \hat c^{(m)}_{\pi(i)} - c^{(m)}_i \bigr|
\;\le\; ( 1 - a_i )\, \hat c^{(m)}_{\pi(i)}
+ \bigl| a_i\, \hat c^{(m)}_{\pi(i)} - c^{(m)}_i \bigr|
\;\le\; \hat C f' + \kappa_{\mathrm{c}} :
\]
part (iv).
\end{proof}

\begin{proof}[Proof of Corollary~1]
\emph{Step 1 (the retained pool inherits the spark condition).}
When the pool is the recovered one, part (b)(i) gives
$\hat s_{\pi(i)} = s_i$ exactly for every
$i \in J_{\mathrm{act}}$, so the retained columns are the true
active profiles; when the pool is a known component library this
holds by definition. Every $B \subseteq J_{\mathrm{act}}$ with
$|B| \le 2k$ is a subset of the full true pool, so
$\sigma_{\min}( S_B ) \ge s_{\min}$ by
Definition~\ref{def:spark}: the retained pool satisfies the spark
condition with the same margin.

\emph{Step 2 (part (a) at $M = 1$).} The fresh sample obeys the
model of part (a) with $M = 1$ over the retained pool, and the
stated conditions on $(A, \hat c)$ are exactly (C1)--(C2) at
$M = 1$ (there, $\sum_m |A_m| = |A|$ and
$\sum_m |J_m| = |J|$). The window (\ref{eq:thm2-wfix}) holds with
the $M = 1$ levels by hypothesis, so part (a) applies: with
probability at least $1 - \alpha$ over $\varepsilon$, every such
decoding satisfies
\[
A \;=\; J ,
\qquad
\| \hat c - c \|
\;\le\;
\frac{\varepsilon_{\mathrm{fit}}
+ \varepsilon^{(1)}_{\mathrm{prj}}(\alpha)}{s_{\min}} .
\qedhere
\]
\end{proof}

\medskip
\noindent
Three remarks close the appendix. \emph{What each part consumes.}
Part (a) nowhere uses nonnegativity of the candidate
coefficients: the spark condition and parsimony alone carry the
known-pool result, which therefore holds against
sign-unconstrained decoders. Part (b) consumes less than its
statement offers: its proof uses only positivity of the used
coefficients ($c_{\min}$ enters nowhere) and only linear
independence of the subsets named in (C5) and
Definition~\ref{def:spark} (neither $\hat s_{\min}$ nor
$s_{\min}$ enters quantitatively); positivity acts exactly once,
in Step 4, to fix signs, and without it part (b)(i) weakens to
$\hat s_{\pi(i)} = \pm s_i$. The quantitative margins return in
part (c), where every bound is explicit in $\bar\varepsilon$.
Part (c) also pays a $\sqrt{d}$-order noise constant through
$\varepsilon_{\mathrm{ns}}(\alpha)$: the learned directions
depend on the noise, so the fixed-direction projection bounds
that sharpen part (a) (where detection costs only
$\varepsilon_{\mathrm{prj}}(\alpha)$, of order
$\sigma \sqrt{k \log N + \log M}$) are unavailable there.

\medskip
\noindent
\emph{The role of (C7) and (C5).} Without (C7), recovery fails
outright: with $N \ge M$ pool slots, the candidate pool
$\hat s_m = x_0^{(m)} / \| x_0^{(m)} \|$ with $A_m = \{m\}$
satisfies (C3) and (C4) at total usage $M \le \sum_m |J_m|$ yet
recovers nothing; (C7) excludes exactly this compression of
several true components into one bespoke slot. A counting
argument makes (C7) plausible rather than proving it: a subspace
of dimension below $|J|$ can contain the coefficient vectors of
at most $|J| - 1$ same-pattern samples in general position,
against at most $\sum_{r \le k} \binom{N}{r}$ candidate subsets,
so sufficient per-pattern multiplicity rules compression out;
(C7) is stated here as an assumption. Condition (C5) also
protects the component count, not only the geometry: solutions
that represent overlapping true spans by rotated bases, or trade
usage through a single combined profile inside a
two-dimensional span, place more unit columns inside a span than
its dimension and are excluded by (C5) through
Lemmas~\ref{lem:thm2-tight} and \ref{lem:thm2-anchor}.

\medskip
\noindent
\emph{Status of Conjecture~R.} Three proven facts calibrate the
conjecture. Its mechanism is unconditional:
Lemma~\ref{lem:thm2-serve} shows that out-of-span errors cannot
cancel across $\varkappa$ coefficient-diverse samples. Its
$1$-sparse case is closed: Lemma~\ref{lem:thm2-k1} derives its
conclusion from the certificate alone, so part (c) is
unconditional in the $1$-sparse regime (the spark condition
applied to a singleton gives $s_{\min} \le 1$, so the first
condition of (\ref{eq:thm2-wrob}) at $\gamma_{\mathrm{c}} = c_{\min}$ implies
the lemma's hypothesis $\bar\varepsilon < c_{\min}$). And in the
noiseless limit its role vanishes: at $\bar\varepsilon = 0$,
exact span membership follows from tight usage alone
(Lemma~\ref{lem:thm2-tight}), $f \to 0$, and the conclusions of
part (c) reduce to those of part (b). What remains open is purely
combinatorial: ruling out adaptive fragmentation, the
configuration in which, for some active component, the samples of
every singleton-intersection pattern are split among candidate
sets so that no single candidate set accumulates $|J|$
same-pattern samples with $\gamma_{\mathrm{c}}$-independent
coefficients. A pigeonhole over the at most $N^k$ candidate sets
produces recurring sets whenever patterns recur; the missing step
is that a recurring set's coefficient matrix inherits genericity
even though the solver assigns samples to candidate sets after
seeing the data. Alternatively, the conjecture can be bypassed
operationally: its conclusion is checkable on a returned
solution, and adding it to the certificate trades the open lemma
for a runtime check (\S\ref{sec:theory-frontier}).

\medskip
\noindent
\emph{What this appendix establishes.} The three parts return the
answer promised at the head of the appendix: success is now
recognizable. A returned solution can be certified as the truth
with no claim about the search that produced it. With a known
pool, the certificate and one noise event yield every support,
the exact component count, and noise-floor concentrations (part
(a)). With a learned pool, the same interface delivers the
components themselves, up to the relabeling the data provably
cannot resolve (part (b)), and persists under noise up to one
named combinatorial statement whose conclusion is itself
checkable on the solution (part (c)). Corollary~1 turns recovery
into reuse: a certified pool decodes fresh samples one at a time.
Everything downstream stands on this quantification over
certificate-passing solutions: the solver section builds an
instrument whose objective pursues the certificate, the
experiments measure the quantities the certificate names, and
Theorem~3 (Appendix~\ref{app:thm3}) takes over where
finite-sample certification ends, stating what accumulating
samples must deliver.

\section{Formal Statement and Proof of Theorem~3}
\label{app:thm3}

This appendix proves Theorem~3 in full. The theorem states what
accumulating samples deliver, and why the framework's asymptotic
target is the generative process rather than the per-sample
decodings. Certification (Theorem~2) is a finite-sample notion:
it covers the $M$ samples of a fixed dataset. Part (a) shows this
finiteness is intrinsic, not a defect of the certificate: at
fixed SNR, as $M$ grows, no estimator (the pool known, the latent
law supplied, unlimited computation) recovers every sample's
support; all-sample correctness has probability decaying at the
rate $(1-q)^M$, and the noise norm exceeds any fixed tolerance on
infinitely many samples. The claim is made because of the paper's
object of inference: samples are instantiations of the process,
and under noise the instantiations cannot all be solved at any
number of samples, so what can be acquired exactly in the limit
is the process, not its instantiations. Part (b) states that it
is acquired, exactly: estimated by near-minimizing a penalized
minimum-distance criterion (the model's data law matched to the
empirical measure, with the expected usage entering at vanishing
weight as a tie-breaker), the generative process converges almost
surely, in the quotient metric that identifies processes up to
relabeling, to the true process, and the estimated component
count is eventually exactly correct.

Part (a) is proved in three steps: noise recurrence (Step~1),
posterior factorization across samples (Step~2), and strict
positivity of the per-sample posterior error mass $q$ (Step~3).
Part (b) rests on three lemmas, organized in three layers, and
runs the compactness scheme of Wald-type consistency proofs
\citep{wald1949note,pollard1981strong}.
Layer I (Lemma~\ref{lem:thm3-quotient}) constructs the space of
processes modulo relabeling: the map sending a process to the law
of its anonymous configurations separates exactly the relabeling
classes and makes the quotient a compact metric space.
Layer II (Lemma~\ref{lem:thm3-decode}) shows the data law
determines the latent structure: known noise divides out, and a
pool decodes its own mixtures uniquely, with a measurable
decoder. Layer III (Lemma~\ref{lem:thm3-popid}) is the population
identification theorem: any process that reproduces the true
noiseless law at no greater expected usage is the true process up
to relabeling; ``population'' here and below means: at the level
of the exact laws rather than a finite sample. The proof
of part (b) then runs in five steps (uniform control, value
comparison, limit points, identification, quotient convergence)
followed by a four-step count argument that reads the component
count off the quotient limit through small balls around the true
component profiles.

\paragraph{Notation.}
The notation of Appendices~\ref{app:prop1} and \ref{app:thm1}
remains in force. New objects:
\begin{center}
\begin{tabular}{@{}ll@{}}
\toprule
$\varphi$ & the noise law of $\varepsilon$ on $\R^d$; in part (a),
$\mathcal{N}(0, \sigma^2 I_d)$\\
$\Omega_k$ & $= \bigsqcup_{A \subseteq [N],\, |A| \le k}
\{A\} \times [c_{\min}, c_{\max}]^{A}$, the latent space\\
$\mathcal{S}$ & pools: $S \in (\mathbb{S}^{d-1})^N$ with
$\sigma_{\min}(S_B) \ge s_{\min}$ for all $|B| \le 2k$\\
$G = (S, \mu)$ & a generative process: pool $S \in \mathcal{S}$,
latent law $\mu \in \mathcal{P}(\Omega_k)$\\
$\Pi$ & $= \mathcal{S} \times \mathcal{P}(\Omega_k)$, the class of
processes; $G^{\ast} = (S^{\ast}, \mu^{\ast})$ the truth\\
$\Lambda_S$ & mixing map $(A, c) \mapsto \sum_{j \in A} c_j S_j$\\
$\nu_G$ & $= (\Lambda_S)_{\#}\mu$, the noiseless law of the
process\\
$G_x$ & $= \nu_G \ast \varphi$, the data law\\
$U(G)$, $\rho_j(G)$ & expected usage $\E_{\mu}|A|$; activation
probability $\mu(j \in A)$\\
$\mathrm{act}(G)$, $K^{\ast}$ & active set
$\{j : \rho_j(G) > 0\}$; $K^{\ast} = |\mathrm{act}(G^{\ast})|$, the
true count\\
$\approx$, $[G]$ & relabeling equivalence (below); the class of $G$
in $\Pi/\!\approx$\\
$\Xi$, $T(G)$ & anonymous configuration
$\{(S_j, c_j) : j \in A\}$; its law under $\mu$\\
$\beta$ & bounded-Lipschitz metric between laws\\
$d_G$ & $= \beta\bigl(T(\cdot), T(\cdot)\bigr)$, the quotient
metric on $\Pi/\!\approx$\\
$\mathbb{D}_M$, $\beta_M$ & empirical measure of
$x^{(1)}, \dots, x^{(M)}$; $\beta_M = \beta(\mathbb{D}_M,
G^{\ast}_x)$\\
$F_M$, $\hat G_M$ & criterion
$\beta(G_x, \mathbb{D}_M) + \lambda_M U(G)$; an
$\varepsilon_M$-minimizer of $F_M$\\
$\hat N_M$ & $= \#\{ j : \rho_j(\hat G_M) > \rho_{\min}/2 \}$, the
thresholded component count\\
\bottomrule
\end{tabular}
\end{center}
Eight conventions and facts. First, the gMCR model
(Eq.~\ref{eq:model}) with the spark condition
(Definition~\ref{def:spark}) is exactly sampling from a process
$G^{\ast} \in \Pi$: draw $(J, c) \sim \mu^{\ast}$, observe
$x = \Lambda_{S^{\ast}}(J, c) + \varepsilon$; the observed samples
are i.i.d.\ with law $G^{\ast}_x$. Second, $\Omega_k$ is a compact
metric space whose pieces $\{A\} \times [c_{\min}, c_{\max}]^A$ are
clopen; $\mathcal{S}$ is compact ($\sigma_{\min}(S_B)$ is
continuous in $S$, so $\mathcal{S}$ is a closed subset of a product
of spheres); $\mathcal{P}(\Omega_k)$ is compact metrizable in the
weak topology by Prokhorov's theorem
\citep[Theorems~II.6.2 and~II.6.4]{parthasarathy1967probability}; hence $\Pi$ is compact
metrizable. The functionals $U$ and $\rho_j$ are weakly continuous,
because $|A|$ and $\mathbf{1}[j \in A]$ are bounded and continuous
on $\Omega_k$ (locally constant on the clopen pieces). Third, the
relabeling equivalence: the symmetric group $\mathrm{Sym}(N)$ acts
on latents by $R_\pi(A, c) = (\pi(A),\, c \circ \pi^{-1})$, with
pushforward $\pi_{\#}\mu = (R_\pi)_{\#}\mu$, and
\[
G \approx G' \;:\iff\; \exists\, \pi \in \mathrm{Sym}(N):\ \
\pi_{\#}\mu = \mu' \ \ \text{and} \ \ S'_{\pi(j)} = S_j
\ \ \text{for every } j \in \mathrm{act}(G) .
\]
This is an equivalence relation: the identity witnesses
reflexivity, and the identity
$\rho_{\pi(j)}(G') = \rho_j(G)$ (immediate from
$\pi_{\#}\mu = \mu'$) shows $\pi$ maps $\mathrm{act}(G)$ onto
$\mathrm{act}(G')$, so $\pi^{-1}$ and composites witness symmetry
and transitivity. Fourth, the configuration space: on
$X := \mathbb{S}^{d-1} \times [c_{\min}, c_{\max}]$ with the metric
$d\bigl((s,c), (s',c')\bigr) = \max\{\|s - s'\|,\, |c - c'|\}$, let
$\mathcal{K}_k$ be the space of subsets of $X$ of cardinality at
most $k$: the nonempty ones under the Hausdorff metric, the empty
configuration adjoined as an isolated point. $\mathcal{K}_k$ is a
compact metric space, and $T(G)$ is the pushforward of $\mu$ under
the assembly map $(A, c) \mapsto \{(S_j, c_j) : j \in A\}$
(continuous on each clopen piece for fixed $S$). Fifth, $\beta$
denotes the bounded-Lipschitz metric,
$\beta(\mu, \nu) = \sup\{\, |\int f\, d\mu - \int f\, d\nu| :
\|f\|_\infty + \mathrm{Lip}(f) \le 1 \,\}$; on a separable metric
space it metrizes weak convergence \citep[\S 11.3]{dudley2002real}.
Sixth, part (a): an estimator is a measurable map
$(x^{(1)}, \dots, x^{(M)}) \mapsto (\hat A_1, \dots, \hat A_M)$
into supports (the known pool $S$ may enter as a parameter), the
posterior $\Pr(J = A \mid x)$ is taken under the joint law of one
draw $(J, c, \varepsilon)$ of the model, and
\[
q \;:=\; 1 - \E\Bigl[\, \max_{A}\, \Pr\bigl(J = A \mid x\bigr)
\Bigr] .
\]
Seventh, two shorthands of Lemma~\ref{lem:thm3-popid}: the
\emph{class spark} condition is the spark condition that every
pool in $\mathcal{S}$ satisfies by membership in the configuration
space; the \emph{true spark} condition is the same condition for
the true pool $S^{\ast}$. Eighth, the symbol $K^{\ast}$:
Appendix~\ref{app:thm2} uses the dataset-level count
$|\bigcup_m J_m|$, while here $K^{\ast} = |\mathrm{act}(G^{\ast})|$
is the process-level count; the two agree almost surely for all
sufficiently large $M$, since every $j \in \mathrm{act}(G^{\ast})$
has $\rho_j(G^{\ast}) > 0$ and therefore eventually appears in
some support (second Borel--Cantelli), while components outside
$\mathrm{act}(G^{\ast})$ never appear. Each appendix uses the
count of its own setting.

\paragraph{Standing assumptions of part (b).}
Throughout part (b) the observed samples are i.i.d.\ with law
$G^{\ast}_x$ for a true process $G^{\ast} \in \Pi$; membership in
$\Pi$ is the model (Eq.~\ref{eq:model}) together with the spark
condition (Definition~\ref{def:spark}) on the true pool. Four
further assumptions:
\begin{itemize}[leftmargin=*]
\item \textbf{(D1) Known noise.} The noise law $\varphi$ is known
and its characteristic function $\widehat{\varphi}$ vanishes
nowhere. Gaussian noise qualifies, and the assumption is a
boundary rather than a convenience: with fully unknown noise the
concentration law is not identifiable, since concentration spread
and noise can be exchanged along a component ray.
\item \textbf{(D2) Richness.} For every
$i \in \mathrm{act}(G^{\ast})$ there are supports $B_1, B_2$ with
$\mu^{\ast}\bigl(\{B_l\} \times [c_{\min}, c_{\max}]^{B_l}\bigr)
> 0$ for $l = 1, 2$ and $B_1 \cap B_2 = \{i\}$. Two supports
overlapping in exactly one component pin that component profile to
a single pool slot (Lemma~\ref{lem:thm3-popid}, Step~3).
\item \textbf{(D3) Smooth concentrations.} For every support $A$
of positive $\mu^{\ast}$-mass, the conditional law
$\mu^{\ast}(\cdot \mid A)$ of the concentrations is absolutely
continuous on $[c_{\min}, c_{\max}]^{A}$. This is what makes every
proper-subspace event null in Lemma~\ref{lem:thm3-popid}.
\item \textbf{(D4) Schedule and near-minimization.} The estimators
$\hat G_M$ are measurable with
$F_M(\hat G_M) \le \inf_{\Pi} F_M + \varepsilon_M$, where
$\lambda_M \downarrow 0$, $\varepsilon_M \downarrow 0$, and almost
surely $(\beta_M + \varepsilon_M)/\lambda_M \to 0$.
Near-minimization over the nonconvex $\Pi$ is an assumption about
what the run achieved, not a certified runtime property; the
schedule condition is nonvacuous, since $\beta \le W_1$ and the
concentration bounds of \citet[Theorem~2]{fournier2015rate} combined with
Borel--Cantelli give an almost-sure rate
$\beta_M = O(M^{-\gamma})$ for an explicit $\gamma(d) > 0$, so
any $\lambda_M = M^{-a}$ with $0 < a < \gamma$ and
$\varepsilon_M = o(\lambda_M)$ qualifies.
\end{itemize}

\paragraph{Theorem~3, restated.}
\emph{(a)} Fix $\sigma > 0$ and a pool of unit-norm component
profiles $S$. Suppose the samples follow Eq.~\ref{eq:model} with
$\varepsilon^{(m)} \sim \mathcal{N}(0, \sigma^2 I_d)$ i.i.d., the
latent pairs $(J_m, c^{(m)})$ i.i.d.\ under a law that assigns
positive probability to at least two distinct supports, and latents
independent of the noise. Then:

\emph{(i)} for every fixed $\epsilon > 0$, almost surely
$\|\varepsilon^{(m)}\| > \epsilon$ for infinitely many $m$;

\emph{(ii)} for every $M$ and every measurable map from
$(x^{(1)}, \dots, x^{(M)})$ to supports
$(\hat A_1, \dots, \hat A_M)$,
\[
\Pr\bigl( \hat A_m = J_m \ \text{for all } m \le M \bigr)
\;\le\; (1 - q)^{M} ,
\qquad q \in (0, 1) ,
\]
where $q$ (Notation) depends only on $S$, the latent law, and
$\sigma$, and not on $M$.

\emph{(b)} Under the standing assumptions (D1)--(D4), with
\[
\rho_{\min} \;:=\; \min_{i \in \mathrm{act}(G^{\ast})}
\rho_i(G^{\ast}) \;>\; 0 :
\qquad
d_G\bigl( [\hat G_M],\, [G^{\ast}] \bigr) \;\longrightarrow\; 0
\quad \text{almost surely,}
\]
and almost surely $\hat N_M = K^{\ast}$ for all sufficiently large
$M$.

\begin{proof}[Proof of part (a)]
\emph{Step 1 (noise recurrence).} Fix $\epsilon > 0$ and set
\[
p_0 \;:=\; \Pr\bigl( \|\varepsilon\| > \epsilon \bigr) \;>\; 0 ,
\]
which is positive because $\|\varepsilon\|^2 / \sigma^2$ is a
$\chi^2_d$ variable, whose support is all of $[0, \infty)$. The
events $\{ \|\varepsilon^{(m)}\| > \epsilon \}$ are independent
with constant probability $p_0$, so
\[
\sum_{m=1}^{\infty}
\Pr\bigl( \|\varepsilon^{(m)}\| > \epsilon \bigr)
\;=\; \sum_{m=1}^{\infty} p_0 \;=\; \infty ,
\]
and the second Borel--Cantelli lemma
\citep[\S 2.3]{durrett2019probability} gives
$\|\varepsilon^{(m)}\| > \epsilon$ for infinitely many $m$, almost
surely. This proves (i).

\emph{Step 2 (posterior factorization).} Write
$\mathcal{D} = (x^{(1)}, \dots, x^{(M)})$. The triples
$(J_m, c^{(m)}, \varepsilon^{(m)})$ are independent across $m$ and
$x^{(m)}$ is a function of the $m$-th triple alone, so
conditionally on $\mathcal{D}$ the supports $J_1, \dots, J_M$ are
independent, with
\[
\Pr\bigl( J_m = A \mid \mathcal{D} \bigr)
\;=\; \Pr\bigl( J_m = A \mid x^{(m)} \bigr)
\qquad \text{for every } A .
\]
Each $\hat A_m$ is $\mathcal{D}$-measurable, hence a constant
given $\mathcal{D}$, and therefore
\begin{align*}
\Pr\bigl( \hat A_m = J_m \ \forall m \le M \bigr)
&= \E\Bigl[ \Pr\bigl( J_1 = \hat A_1, \dots, J_M = \hat A_M
\bigm| \mathcal{D} \bigr) \Bigr]\\
&= \E\Bigl[ \textstyle\prod_{m=1}^{M}
\Pr\bigl( J_m = \hat A_m \bigm| x^{(m)} \bigr) \Bigr]\\
&\le \E\Bigl[ \textstyle\prod_{m=1}^{M}
\max_{A} \Pr\bigl( J = A \bigm| x^{(m)} \bigr) \Bigr]\\
&= \prod_{m=1}^{M}
\E\Bigl[ \max_{A} \Pr\bigl( J = A \bigm| x^{(m)} \bigr) \Bigr]
\;=\; (1 - q)^{M} ,
\end{align*}
by, in order: the tower property; conditional independence of the
supports given $\mathcal{D}$ (display above); bounding each factor
by the maximum over supports; independence of the $x^{(m)}$; their
identical distribution and the definition of $q$.

\emph{Step 3 ($q$ lies in $(0,1)$).} The supports range over the
finite family $\mathcal{A} = \{ A \subseteq [N] : |A| \le k \}$,
and the posteriors sum to one, so pointwise
\[
\max_{A} \Pr\bigl( J = A \mid x \bigr)
\;\ge\; \frac{1}{|\mathcal{A}|} ,
\qquad \text{hence} \qquad
q \;\le\; 1 - \frac{1}{|\mathcal{A}|} \;<\; 1 .
\]
For $q > 0$: pick distinct supports $A_1 \ne A_2$ with prior
masses $p_l := \Pr(J = A_l) > 0$. Conditionally on $J = A_l$, the
sample $x = S_{A_l} c + \varepsilon$ has density
\[
f_l(x) \;=\; \int \varphi_\sigma\bigl( x - S_{A_l} c \bigr)\,
d\mathrm{Law}\bigl( c \mid J = A_l \bigr) \;>\; 0
\qquad \text{for every } x \in \R^d ,
\]
where $\mathrm{Law}( c \mid J = A_l )$ denotes the conditional
distribution of the concentration vector given the event
$J = A_l$ (a probability measure on the coefficient space): the
integral averages the Gaussian density over the concentrations the
model produces with support $A_l$, and is strictly positive
because $\varphi_\sigma$ is strictly positive everywhere. The marginal
density is $f = \sum_{A} \Pr(J = A)\, f_A$, and Bayes' rule gives,
for every $x$,
\[
\Pr\bigl( J = A_l \mid x \bigr)
\;=\; \frac{ p_l\, f_l(x) }{ f(x) } \;>\; 0 ,
\qquad l = 1, 2 .
\]
Any maximizer of $A \mapsto \Pr(J = A \mid x)$ differs from at
least one of $A_1, A_2$, and the posteriors are nonnegative and
sum to one, so for every $x$,
\[
\max_{A} \Pr\bigl( J = A \mid x \bigr)
\;\le\; 1 - \min\Bigl\{ \Pr\bigl( J = A_1 \mid x \bigr),\
\Pr\bigl( J = A_2 \mid x \bigr) \Bigr\}
\;<\; 1 .
\]
The variable $1 - \max_A \Pr(J = A \mid x)$ is a measurable
function of $x$ (a finite maximum over the posterior kernel's
components), bounded in $[0,1]$, and strictly positive; its
expectation is therefore strictly positive (a nonnegative variable
with zero expectation vanishes almost surely):
\[
q \;=\; \E\Bigl[ 1 - \max_{A} \Pr\bigl( J = A \mid x \bigr)
\Bigr] \;>\; 0 .
\]
Finally $q$ is defined from the law of a single draw
$(J, c, \varepsilon)$ and the pool, so it does not depend on $M$.
This proves (ii).
\end{proof}

\noindent
The remaining material proves part (b), in three layers and an
assembly: Layer I constructs the space in which the limit lives,
Layer II shows the observed law determines the latent structure,
Layer III identifies the process from its population law, and the
assembly runs the compactness scheme and reads off the count.

\subsection*{Layer I: the space of processes
(Lemma \ref{lem:thm3-quotient})}

Convergence up to relabeling needs a space in which it is plain
convergence. This layer builds it: anonymous configurations
separate exactly the relabeling classes, and the resulting
quotient is a compact metric space.

\begin{lemma}[Quotient metric]
\label{lem:thm3-quotient}
$T$ is continuous and $\approx$-invariant, and $T(G) = T(G')$ if
and only if $G \approx G'$. Consequently
$d_G\bigl([G], [G']\bigr) := \beta\bigl(T(G), T(G')\bigr)$ is a
metric on $\Pi/\!\approx$, the space $(\Pi/\!\approx,\, d_G)$ is
compact, and $d_G$ metrizes the quotient topology.
\end{lemma}

\begin{proof}
\emph{Step 1 (slot separation).} Fix $G = (S, \mu) \in \Pi$ and
$j \ne j'$. The Gram matrix of the pair $(S_j, S_{j'})$ is
$\bigl(\begin{smallmatrix} 1 & t \\ t & 1 \end{smallmatrix}\bigr)$
with $t = \langle S_j, S_{j'} \rangle$, with eigenvalues
$1 \pm t$; the spark condition at $|B| = 2$ gives
$\sigma_{\min}(S_{\{j, j'\}})^2 = 1 - |t| \ge s_{\min}^2$, hence
\[
\| S_j - S_{j'} \|^2 \;=\; 2 - 2t \;\ge\; 2\, s_{\min}^2
\;=:\; \Delta^2 ,
\qquad \Delta = \sqrt{2}\, s_{\min} \;>\; 0 .
\]
Every process in $\Pi$ therefore has pairwise
$\Delta$-separated component profiles. In particular the points of
$\Xi = \{(S_j, c_j) : j \in A\}$ have pairwise distinct first
coordinates, $|\Xi| = |A|$, and for fixed $S$ the assembly map
$(A, c) \mapsto \Xi$ is injective on $\Omega_k$.

\emph{Step 2 (invariance).} First,
\[
\mu\bigl( A \not\subseteq \mathrm{act}(G) \bigr)
\;\le\; \sum_{j \notin \mathrm{act}(G)} \mu( j \in A )
\;=\; \sum_{j \notin \mathrm{act}(G)} \rho_j(G) \;=\; 0 ,
\]
so $A \subseteq \mathrm{act}(G)$ for $\mu$-a.e.\ latent. Let
$G \approx G'$ with witness $\pi$. For any latent $(A, c)$ with
$A \subseteq \mathrm{act}(G)$, the configuration of the relabeled
latent under the pool $S'$ is
\[
\bigl\{ \bigl( S'_{j'},\, (c \circ \pi^{-1})_{j'} \bigr) :
j' \in \pi(A) \bigr\}
\;=\; \bigl\{ \bigl( S'_{\pi(j)},\, c_j \bigr) : j \in A \bigr\}
\;=\; \bigl\{ ( S_j, c_j ) : j \in A \bigr\} ,
\]
the configuration of $(A, c)$ under $S$, using
$S'_{\pi(j)} = S_j$ on $\mathrm{act}(G)$. Pushing
$\mu' = \pi_{\#}\mu$ through the assembly therefore gives
$T(G') = T(G)$.

\emph{Step 3 (continuity).} Let $G_n = (S_n, \mu_n) \to
G = (S, \mu)$ in $\Pi$ and let $f$ on $\mathcal{K}_k$ satisfy
$\|f\|_\infty + \mathrm{Lip}(f) \le 1$. Then
\begin{align*}
\Bigl| \int f\, dT(G_n) - \int f\, dT(G) \Bigr|
\;\le\;&
\sup_{(A, c) \in \Omega_k}
\bigl| f\bigl( \Xi_{S_n}(A, c) \bigr)
- f\bigl( \Xi_{S}(A, c) \bigr) \bigr|\\
&+
\Bigl| \int (f \circ \Xi_S)\, d\mu_n
- \int (f \circ \Xi_S)\, d\mu \Bigr| ,
\end{align*}
where $\Xi_S$ denotes the assembly map under $S$. The first term
is controlled by the Hausdorff distance: matching the points of
the two configurations slot by slot,
\[
d_H\bigl( \Xi_{S_n}(A, c),\, \Xi_S(A, c) \bigr)
\;\le\; \max_{j \in A} \| S_{n, j} - S_j \|
\;\longrightarrow\; 0
\qquad \text{uniformly over } \Omega_k ,
\]
so the supremum tends to $0$ ($f$ is Lipschitz). The second term
tends to $0$ by weak convergence $\mu_n \rightharpoonup \mu$,
because $f \circ \Xi_S$ is bounded and continuous on $\Omega_k$
(continuous on each clopen piece). Taking the supremum over such
$f$: $\beta\bigl( T(G_n), T(G) \bigr) \to 0$.

\emph{Step 4 ($T$ determines the class).} Suppose
$T(G) = T(G')$ with $G = (S, \mu)$, $G' = (S', \mu')$. For a unit
vector $s$, put $K_s := \{s\} \times [c_{\min}, c_{\max}]$, a
compact subset of $X$. By Step~1 at most one slot carries the
profile $s$, so
\[
T(G)\bigl( \{ \Xi : \Xi \cap K_s \ne \emptyset \} \bigr)
\;=\; \mu\bigl( \exists\, j \in A : S_j = s \bigr) ,
\]
which equals $\rho_j(G)$ when $s = S_j$ for the (unique) slot $j$,
and $0$ when no slot carries $s$.
The set $\{ S_j : j \in \mathrm{act}(G) \}$ is therefore
recoverable from $T(G)$ as the profiles carried with positive
probability, together with their activation masses. Equality of
the two laws gives
$\{ S_j : j \in \mathrm{act}(G) \}
= \{ S'_{j'} : j' \in \mathrm{act}(G') \}$ as subsets of the
sphere, so there is a bijection
$\pi : \mathrm{act}(G) \to \mathrm{act}(G')$ with
$S'_{\pi(j)} = S_j$, unique by pairwise distinctness on each side;
extend $\pi$ to a permutation of $[N]$ arbitrarily.

Next, the match-back map. For $B \subseteq \mathrm{act}(G)$
define
\[
D_B \;:=\; \Bigl\{ \Xi \in \mathcal{K}_k :\
\Xi \cap K_{S_j} \ne \emptyset \ \forall j \in B, \ \
\Xi \subseteq \textstyle\bigcup_{j \in B} K_{S_j}, \ \
|\Xi| = |B| \Bigr\} .
\]
Each defining condition is Borel in the Hausdorff topology on
$\mathcal{K}_k$ \citep[cf.\ \S 4.F]{kechris1995classical}:
$\{ \Xi : \Xi \cap K \ne \emptyset \}$ is closed for compact $K$,
$\{ \Xi : \Xi \subseteq F \}$ is closed for closed $F$ (its
complement is $\{ \Xi : \Xi \cap F^{c} \ne \emptyset \}$, open
since $F^c$ is open), and $\{ \Xi : |\Xi| \le n \}$ is closed. The
sets $D_B$ are pairwise disjoint, and a configuration in $D_B$
carries exactly one point over each $S_j$, $j \in B$. On
$\{ \Xi : \Xi \cap K_{S_j} \ne \emptyset \}$ define
\[
g_j(\Xi) \;:=\; \min\bigl\{ t \in [c_{\min}, c_{\max}] :
(S_j, t) \in \Xi \bigr\} ;
\]
$g_j$ is Borel, because
$\{ \Xi : g_j(\Xi) \le a \}
= \{ \Xi : \Xi \cap (\{S_j\} \times [c_{\min}, a]) \ne
\emptyset \}$ is closed. The match-back map
\[
m(\Xi) \;:=\; \bigl( B,\ ( g_j(\Xi) )_{j \in B} \bigr)
\qquad \text{on } D_B , \quad B \subseteq \mathrm{act}(G) ,
\]
is Borel on $\bigcup_B D_B$, and on the $\mu$-full event
$\{ A \subseteq \mathrm{act}(G) \}$ it inverts the assembly:
$m\bigl( \Xi_S(A, c) \bigr) = (A, c)$, since the assembled
configuration lies in $D_A$ and $g_j$ reads off $c_j$. Since
$T(G)\bigl( \bigcup_B D_B \bigr)
= \mu\bigl( A \subseteq \mathrm{act}(G) \bigr) = 1$ (Step~2),
\[
\mu \;=\; m_{\#}\, T(G) .
\]
Build $m'$ from $S'$ over $\mathrm{act}(G')$ the same way. A
configuration $\Xi \in D_B$ has first coordinates
$\{ S_j : j \in B \} = \{ S'_{\pi(j)} : j \in B \}$, so it lies in
the $S'$-set $D'_{\pi(B)}$, and $g'_{\pi(j)}(\Xi) = g_j(\Xi)$ (the
same slice of $\Xi$); hence $m' = R_\pi \circ m$ on
$\bigcup_B D_B$, and
\[
\mu' \;=\; m'_{\#}\, T(G')
\;=\; m'_{\#}\, T(G)
\;=\; (R_\pi)_{\#}\, m_{\#}\, T(G)
\;=\; \pi_{\#}\, \mu .
\]
Together with $S'_{\pi(j)} = S_j$ on $\mathrm{act}(G)$ this is
precisely $G \approx G'$. With Step~2, $T(G) = T(G')$ if and only
if $G \approx G'$.

\emph{Step 5 (metric).} $d_G$ is well defined on classes by
Step~2, symmetric with the triangle inequality inherited from
$\beta$, and $d_G([G], [G']) = 0$ if and only if $T(G) = T(G')$
($\beta$ is a metric) if and only if $G \approx G'$ (Step~4), that
is, $[G] = [G']$.

\emph{Step 6 (compactness and the quotient topology).} The map
$[G] \mapsto T(G)$ is a well-defined isometry from
$(\Pi/\!\approx, d_G)$ onto $(T(\Pi), \beta)$; $T(\Pi)$ is compact
as the continuous image (Step~3) of the compact $\Pi$, so
$(\Pi/\!\approx, d_G)$ is compact. The quotient topology is
compact as the image of $\Pi$ under the (continuous) quotient map,
and the identity from $\Pi/\!\approx$ with the quotient topology
to $(\Pi/\!\approx, d_G)$ is continuous ($T$ is continuous and
constant on classes, so $d_G$-open sets pull back to open sets); a
continuous bijection from a compact space to a Hausdorff space is
a homeomorphism, so $d_G$ metrizes the quotient topology.
\end{proof}

\subsection*{Layer II: from the data law to the latent law
(Lemma \ref{lem:thm3-decode})}

This layer shows the observation determines the structure: the
known noise divides out of the data law, and a pool reads its own
mixtures uniquely, with a measurable decoder; the pool and the
noiseless law together determine the latent law.

\begin{lemma}[From the data law to the latent law]
\label{lem:thm3-decode}
Let $S \in \mathcal{S}$. Then:
\emph{(i)} under (D1), if $\nu, \nu'$ are finite Borel measures
on $\R^d$ with $\nu \ast \varphi = \nu' \ast \varphi$, then
$\nu = \nu'$; in particular $G_x = G'_x$ implies
$\nu_G = \nu_{G'}$;
\emph{(ii)} $\Lambda_S$ is injective on $\Omega_k$;
\emph{(iii)} $\Lambda_S$ is a homeomorphism onto its range
$R_S := \Lambda_S(\Omega_k)$, and $\Lambda_S^{-1} : R_S \to
\Omega_k$ is Borel, as are the images $\Lambda_S(B)$ of Borel
sets $B \subseteq \Omega_k$;
\emph{(iv)} for every $G = (S, \mu) \in \Pi$,
\[
\mu \;=\; \bigl( \Lambda_S^{-1} \bigr)_{\#}\, \nu_G :
\]
the pool and the noiseless law determine the latent law.
\end{lemma}

\begin{proof}
\emph{Step 1 (Fourier factorization).} For every $t \in \R^d$,
the characteristic function of a convolution is the product:
\[
\widehat{\nu \ast \varphi}\,(t)
\;=\; \widehat{\nu}(t)\; \widehat{\varphi}(t) ,
\qquad
\widehat{\nu' \ast \varphi}\,(t)
\;=\; \widehat{\nu'}(t)\; \widehat{\varphi}(t) .
\]

\emph{Step 2 (division).} The hypothesis of part (i) equates the
left sides, and $\widehat{\varphi}(t) \ne 0$ for every $t$ by
(D1), so
\[
\widehat{\nu}(t) \;=\; \widehat{\nu'}(t)
\qquad \text{for every } t \in \R^d .
\]

\emph{Step 3 (uniqueness; part (i)).} Finite Borel measures on
$\R^d$ with equal characteristic functions coincide
\citep[\S 9.5]{dudley2002real}. Hence $\nu = \nu'$; applying this
to $\nu_G, \nu_{G'}$ gives the second claim of part (i).

\emph{Step 4 (injectivity; part (ii)).} Suppose
$\Lambda_S(A, c) = \Lambda_S(A', c')$. Define $v$ on
$A \cup A'$ (with $|A \cup A'| \le 2k$) coordinate-wise, exactly
as in Step~1 of Appendix~\ref{app:prop1}:
\[
v_j =
\begin{cases}
c_j, & j \in A \setminus A',\\
c_j - c'_j, & j \in A \cap A',\\
-\,c'_j, & j \in A' \setminus A.
\end{cases}
\]
Then $S_{A \cup A'}\, v = \Lambda_S(A, c) - \Lambda_S(A', c') = 0$,
and the spark condition gives
\[
0 \;=\; \| S_{A \cup A'}\, v \|
\;\ge\; s_{\min}\, \|v\| ,
\qquad \text{hence} \qquad v = 0 .
\]
On $A \setminus A'$ and $A' \setminus A$ the entries of $v$ have
magnitude at least $c_{\min} > 0$, so both differences are empty:
$A = A'$; and then $c - c' = v = 0$.

\emph{Step 5 (Borel inverse; part (iii)).} $\Lambda_S$ is
continuous on $\Omega_k$ (linear in $c$ on each clopen piece),
$\Omega_k$ is compact, and $\R^d$ is Hausdorff, so $\Lambda_S$ is
a closed map and, being injective by Step~4, a homeomorphism onto
$R_S$; its inverse is continuous on $R_S$, hence Borel. (In
general, an injective Borel map between standard Borel spaces has
Borel image and Borel inverse by the Lusin--Suslin theorem,
\citealp[Theorem 15.1]{kechris1995classical}; compactness makes
the elementary route sufficient here.) For Borel
$B \subseteq \Omega_k$, the image
$\Lambda_S(B) = (\Lambda_S^{-1})^{-1}(B)$ is Borel in the compact
$R_S$, hence Borel in $\R^d$.

\emph{Step 6 (measure determination; part (iv)).} For every Borel
$B \subseteq \Omega_k$, using $\nu_G = (\Lambda_S)_{\#}\mu$ and
injectivity,
\[
\bigl( \Lambda_S^{-1} \bigr)_{\#} \nu_G\, (B)
\;=\; \nu_G\bigl( \Lambda_S(B) \bigr)
\;=\; \mu\bigl( \Lambda_S^{-1}\bigl( \Lambda_S(B) \bigr) \bigr)
\;=\; \mu(B) .
\]
\end{proof}

\subsection*{Layer III: population identification
(Lemma \ref{lem:thm3-popid})}

This layer is the identification engine: a usage-minimal
reproduction of the true noiseless law is the true process up to
relabeling. At the population level the assumptions lighten
rather than accumulate: the coefficient-genericity and
no-compression conditions of the finite-sample theory become
theorems here, because absolutely continuous concentration laws
(D3) make every proper-subspace event null. Wherever linear
independence of profiles is used below, it is supplied by the
class or true spark condition on index sets of at most $2k$
elements.

\begin{lemma}[Population identification]
\label{lem:thm3-popid}
Let $G^{\ast} \in \Pi$ obey (D2)--(D3), and let
$G = (S, \mu) \in \Pi$ satisfy
\[
\nu_G \;=\; \nu_{G^{\ast}}
\qquad \text{and} \qquad
U(G) \;\le\; U(G^{\ast}) .
\]
Then $G \approx G^{\ast}$.
\end{lemma}

\begin{proof}
Write $\nu := \nu_{G^{\ast}} = \nu_G$ and
$V_B := \mathrm{span}( s^{\ast}_i : i \in B )$ for
$B \subseteq [N]$.

\emph{Step 0 (decoding both pools; the degenerate case).} By
Lemma~\ref{lem:thm3-decode} applied to $S$ and to $S^{\ast}$, for
every $x$ in the $\nu$-full sets $R_S$ and $R_{S^{\ast}}$ there
are unique decodings
\[
x \;=\; \Lambda_S\bigl( \alpha(x),\, c(x) \bigr)
\;=\; \Lambda_{S^{\ast}}\bigl( \alpha^{\ast}(x),\, c^{\ast}(x)
\bigr) ,
\]
with Borel maps $\alpha, c, \alpha^{\ast}, c^{\ast}$; the change
of variables $\nu = (\Lambda_S)_{\#}\mu$ gives
\[
\E_\nu |\alpha| \;=\; \E_\mu |A| \;=\; U(G) ,
\qquad
\E_\nu |\alpha^{\ast}| \;=\; U(G^{\ast}) ,
\]
and, for any Borel set $\Gamma \subseteq \Omega_k$,
$\nu\bigl( (\alpha, c) \in \Gamma \bigr) = \mu(\Gamma)$, likewise for the
starred pair with $\mu^{\ast}$. If
$\mathrm{act}(G^{\ast}) = \emptyset$ then
$U(G^{\ast}) = 0$, so $U(G) = \E_\mu |A| = 0$, both $\mu$ and
$\mu^{\ast}$ put full mass on the empty support, and the identity
permutation witnesses $G \approx G^{\ast}$. Assume henceforth
$\mathrm{act}(G^{\ast}) \ne \emptyset$; by (D2) some support of
positive mass is nonempty, so $k \ge 1$.

\emph{Step 1 (full-rank service is automatic).} Fix a pair of
supports $(A, B)$ and suppose
$W_{A,B} := \mathrm{span}(S_A) \cap V_B \subsetneq V_B$. On the
event $\{ \alpha = A,\ \alpha^{\ast} = B \}$ the decodings place
\[
x \;\in\; \mathrm{span}\bigl( S_{\alpha(x)} \bigr)
\cap V_{\alpha^{\ast}(x)}
\;=\; W_{A, B} .
\]
The event $\{ \alpha^{\ast} = B \}$ is, up to $\nu$-null sets, the
image of the latent event $\{ A' = B \}$, so conditioning on it,
$x = S^{\ast}_B\, c$ with $c \sim \mu^{\ast}(\cdot \mid B)$
absolutely continuous by (D3); the preimage
$(S^{\ast}_B)^{-1}( W_{A,B} )$ is a proper linear subspace of
$\R^{B}$ (the map $S^{\ast}_B$ is injective by the true spark
condition, and $W_{A,B} \subsetneq V_B$), hence Lebesgue-null, so
\[
\nu\bigl( \alpha = A,\ \alpha^{\ast} = B \bigr)
\;\le\;
\mu^{\ast}\bigl( A' = B \bigr) \cdot
\mu^{\ast}\bigl( c \in (S^{\ast}_B)^{-1}(W_{A,B}) \bigm| B \bigr)
\;=\; 0 .
\]
There are finitely many pairs $(A, B)$, so for $\nu$-a.e.\ $x$,
\[
V_{\alpha^{\ast}(x)} \;\subseteq\;
\mathrm{span}\bigl( S_{\alpha(x)} \bigr) .
\]

\emph{Step 2 (tight service, a.e.).} For $\nu$-a.e.\ $x$, by
Step~1 and $\dim V_{B} = |B|$ (true spark),
\[
|\alpha(x)| \;\ge\; \dim \mathrm{span}\bigl( S_{\alpha(x)} \bigr)
\;\ge\; \dim V_{\alpha^{\ast}(x)}
\;=\; |\alpha^{\ast}(x)| .
\]
Integrating against $\nu$ and using the hypothesis,
\[
U(G^{\ast}) \;\ge\; U(G)
\;=\; \E_\nu |\alpha|
\;\ge\; \E_\nu |\alpha^{\ast}|
\;=\; U(G^{\ast}) ,
\]
so $\E_\nu\bigl[ |\alpha| - |\alpha^{\ast}| \bigr] = 0$ with a
$\nu$-a.e.\ nonnegative integrand, forcing
$|\alpha(x)| = |\alpha^{\ast}(x)|$ for $\nu$-a.e.\ $x$. The
columns $S_{\alpha(x)}$ are linearly independent
($|\alpha(x)| \le k \le 2k$, class spark), so
\[
\dim \mathrm{span}\bigl( S_{\alpha(x)} \bigr)
\;=\; |\alpha(x)|
\;=\; |\alpha^{\ast}(x)|
\;=\; \dim V_{\alpha^{\ast}(x)} ,
\]
and with the containment of Step~1 (equal finite dimensions),
\[
\mathrm{span}\bigl( S_{\alpha(x)} \bigr)
\;=\; V_{\alpha^{\ast}(x)}
\qquad \text{for $\nu$-a.e.\ } x .
\]

\emph{Step 3 (anchoring).} Fix $i \in \mathrm{act}(G^{\ast})$ and
the (D2) supports $B_1, B_2$ with $B_1 \cap B_2 = \{i\}$. The
events $\{ \alpha^{\ast} = B_l \}$ have positive $\nu$-mass, so
each contains a point of the a.e.-good set of Steps~1--2: pick
such $x_1, x_2$ and set $A_1 := \alpha(x_1)$,
$A_2 := \alpha(x_2)$; by Step~2,
\[
\mathrm{span}\bigl( S_{A_l} \bigr) = V_{B_l} ,
\qquad
|A_l| = |B_l| , \qquad l = 1, 2 .
\]
The true family on $B_1 \cup B_2$ is linearly independent
($|B_1 \cup B_2| \le 2k$, true spark), and spans over an
independent family intersect exactly along shared indices: a
member of $V_{B_1} \cap V_{B_2}$ has a representation over each
of $B_1$ and $B_2$, the difference of the two representations is
a vanishing combination over $B_1 \cup B_2$, and independence
forces every coefficient off $B_1 \cap B_2 = \{i\}$ to vanish.
Hence
\[
V_{B_1} \cap V_{B_2}
\;=\; \mathrm{span}\bigl( s^{\ast}_i \bigr) ,
\qquad
\dim\bigl( V_{B_1} \cap V_{B_2} \bigr) = 1 .
\]
Set $U := A_1 \cup A_2$, of cardinality
$|U| \le |A_1| + |A_2| = |B_1| + |B_2| \le 2k$, so the columns
$S_U$ are independent (class spark) and, by the dimension formula
for a sum of subspaces,
\[
|U| \;=\; \dim \mathrm{span}( S_U )
\;=\; \dim\bigl( V_{B_1} + V_{B_2} \bigr)
\;=\; |B_1| + |B_2| - 1 .
\]
Inclusion--exclusion then yields exactly one shared index:
\[
|A_1 \cap A_2| \;=\; |A_1| + |A_2| - |U| \;=\; 1 ,
\qquad A_1 \cap A_2 =: \{ j^{\ast} \} .
\]
The shared column lies in both spans, so
\[
S_{j^{\ast}} \;\in\;
\mathrm{span}\bigl( S_{A_1} \bigr) \cap
\mathrm{span}\bigl( S_{A_2} \bigr)
\;=\; V_{B_1} \cap V_{B_2}
\;=\; \mathrm{span}\bigl( s^{\ast}_i \bigr) ,
\]
hence $S_{j^{\ast}} = a\, s^{\ast}_i$ for a scalar $a$, and unit
norms give $|a| = 1$: write
$S_{j^{\ast}} = \sigma_i\, s^{\ast}_i$ with
$\sigma_i \in \{ \pm 1 \}$. The slot is unique: a second pool
column in the line $\mathrm{span}(s^{\ast}_i)$ would form with
$S_{j^{\ast}}$ a linearly dependent pair, contradicting the class
spark at $|B| = 2$; so $j^{\ast}$ does not depend on the chosen
pair of generic points, and we set $\pi(i) := j^{\ast}$. The map
$\pi$ is injective: $\pi(i) = \pi(i')$ with $i \ne i'$ would give
$\sigma_i s^{\ast}_i = \sigma_{i'} s^{\ast}_{i'}$, writing one
member of the independent true pair
$\{ s^{\ast}_i, s^{\ast}_{i'} \}$ as a multiple of the other,
which is impossible.

\emph{Step 4 (signs, supports, and coefficients, a.e.).} Fix a
point $x$ of the a.e.-good set and $i \in \alpha^{\ast}(x)$, and
suppose $\pi(i) \notin \alpha(x)$. The pool subset
$\{ \pi(i) \} \cup \alpha(x)$ has cardinality
$|\alpha(x)| + 1 \le k + 1 \le 2k$, hence its columns are
independent (class spark); yet all of them lie in
$V_{\alpha^{\ast}(x)}$: the columns of $\alpha(x)$ by Step~2, and
$S_{\pi(i)} = \sigma_i s^{\ast}_i \in V_{\alpha^{\ast}(x)}$ since
$i \in \alpha^{\ast}(x)$. An independent family in
$V_{\alpha^{\ast}(x)}$ has at most
$\dim V_{\alpha^{\ast}(x)} = |\alpha^{\ast}(x)| = |\alpha(x)|$
members, so
\[
|\alpha(x)| + 1 \;\le\; |\alpha(x)| ,
\]
absurd. Hence $\pi(\alpha^{\ast}(x)) \subseteq \alpha(x)$, and
$|\pi(\alpha^{\ast}(x))| = |\alpha^{\ast}(x)| = |\alpha(x)|$
($\pi$ injective; Step~2) upgrades the containment to
\[
\alpha(x) \;=\; \pi\bigl( \alpha^{\ast}(x) \bigr) .
\]
Substituting $S_{\pi(i)} = \sigma_i s^{\ast}_i$ into the two
decodings of the same point,
\[
\sum_{i \in \alpha^{\ast}(x)}
\bigl( c(x)_{\pi(i)}\, \sigma_i \bigr)\, s^{\ast}_i
\;=\; x
\;=\; \sum_{i \in \alpha^{\ast}(x)}
c^{\ast}(x)_i\, s^{\ast}_i ,
\]
and coefficient extraction along the independent true family on
$\alpha^{\ast}(x)$ gives
\[
c(x)_{\pi(i)}\, \sigma_i \;=\; c^{\ast}(x)_i
\qquad \text{for every } i \in \alpha^{\ast}(x) .
\]
Both $c(x)_{\pi(i)} \ge c_{\min} > 0$ and
$c^{\ast}(x)_i \ge c_{\min} > 0$ (coefficients live in the box),
so $\sigma_i = +1$ whenever some good point has
$i \in \alpha^{\ast}(x)$; the (D2) event $\{\alpha^{\ast} = B_1\}$
with $B_1 \ni i$ has positive mass, so this holds for every
$i \in \mathrm{act}(G^{\ast})$:
\[
S_{\pi(i)} \;=\; s^{\ast}_i
\qquad \text{for every } i \in \mathrm{act}(G^{\ast}) ,
\]
and $c(x)_{\pi(i)} = c^{\ast}(x)_i$ for $\nu$-a.e.\ $x$.

\emph{Step 5 (assembly).} Extend $\pi$ to a permutation of $[N]$
(arbitrarily on inactive indices; immaterial, because
$\mu^{\ast}( A \subseteq \mathrm{act}(G^{\ast}) ) = 1$ by the
computation of Lemma~\ref{lem:thm3-quotient}, Step~2). On the
$\mu^{\ast}$-full event $\{ A \subseteq \mathrm{act}(G^{\ast})\}$,
Step~4's profile identity gives
\[
\Lambda_S\bigl( R_\pi(A, c) \bigr)
\;=\; \sum_{j \in \pi(A)} ( c \circ \pi^{-1} )_j\, S_j
\;=\; \sum_{i \in A} c_i\, S_{\pi(i)}
\;=\; \sum_{i \in A} c_i\, s^{\ast}_i
\;=\; \Lambda_{S^{\ast}}(A, c) ,
\]
so, by injectivity of $\Lambda_S$
(Lemma~\ref{lem:thm3-decode}(ii)),
$\Lambda_S^{-1} \circ \Lambda_{S^{\ast}} = R_\pi$ holds
$\mu^{\ast}$-almost everywhere. Chaining pushforwards
(Lemma~\ref{lem:thm3-decode}(iv)),
\[
\mu
\;=\; \bigl( \Lambda_S^{-1} \bigr)_{\#} \nu
\;=\; \bigl( \Lambda_S^{-1} \bigr)_{\#}
\bigl( \Lambda_{S^{\ast}} \bigr)_{\#} \mu^{\ast}
\;=\; \bigl( \Lambda_S^{-1} \circ \Lambda_{S^{\ast}}
\bigr)_{\#}\, \mu^{\ast}
\;=\; ( R_\pi )_{\#}\, \mu^{\ast}
\;=\; \pi_{\#}\, \mu^{\ast} .
\]
Together with $S_{\pi(i)} = s^{\ast}_i$ on
$\mathrm{act}(G^{\ast})$, the permutation $\pi$ witnesses
$G^{\ast} \approx G$; the relation is symmetric (Notation), so
$G \approx G^{\ast}$.
\end{proof}

\begin{proof}[Proof of part (b)]
All statements are pathwise. By Varadarajan's theorem
\citep{varadarajan1958convergence}, almost surely
$\mathbb{D}_M \rightharpoonup G^{\ast}_x$, and $\beta$ metrizes
weak convergence, so $\beta_M \to 0$ almost surely; intersect
this event with the almost-sure (D4) event
$(\beta_M + \varepsilon_M)/\lambda_M \to 0$, fix one point of the
intersection, and argue deterministically. Steps~1--5 prove the
quotient convergence; Steps~6--9 prove the count claim.

\emph{Step 1 (uniform control).} For every $G \in \Pi$, by the
triangle inequality for $\beta$,
\[
\bigl| \beta( G_x, \mathbb{D}_M )
- \beta( G_x, G^{\ast}_x ) \bigr|
\;\le\; \beta( \mathbb{D}_M, G^{\ast}_x )
\;=\; \beta_M :
\]
the minimum-distance design makes the uniform law of large numbers
a single line.

\emph{Step 2 (value comparison).} Near-minimization against the
candidate $G^{\ast}$, then Step~1 at $G^{\ast}$
($\beta(G^{\ast}_x, \mathbb{D}_M) \le \beta_M$) and
$U(G^{\ast}) = \E_{\mu^{\ast}}|A| \le k$:
\[
F_M(\hat G_M)
\;\le\; F_M(G^{\ast}) + \varepsilon_M
\;=\; \beta( G^{\ast}_x, \mathbb{D}_M )
+ \lambda_M\, U(G^{\ast}) + \varepsilon_M
\;\le\; \beta_M + \lambda_M k + \varepsilon_M .
\]
Both terms of $F_M$ are nonnegative, so each is bounded by the
whole. The fit term, moved to the true law by Step~1,
\[
\rho(\hat G_M) \;:=\;
\beta\bigl( (\hat G_M)_x,\, G^{\ast}_x \bigr)
\;\le\; \beta\bigl( (\hat G_M)_x,\, \mathbb{D}_M \bigr)
+ \beta_M
\;\le\; 2\beta_M + \lambda_M k + \varepsilon_M
\;\longrightarrow\; 0 ,
\]
and the usage term, divided by $\lambda_M > 0$,
\[
U(\hat G_M)
\;\le\; U(G^{\ast})
+ \frac{ \beta_M + \varepsilon_M }{ \lambda_M }
\;\longrightarrow\; U(G^{\ast})
\]
by the (D4) schedule condition.

\emph{Step 3 (limit points).} The functional
$G \mapsto \rho(G) = \beta(G_x, G^{\ast}_x)$ is continuous on
$\Pi$. Indeed, for $\|f\|_\infty + \mathrm{Lip}(f) \le 1$, the
smoothed function $g(y) := \int f(y + e)\, d\varphi(e)$ satisfies
$\|g\|_\infty + \mathrm{Lip}(g) \le 1$, and
\[
\Bigl| \int f\, dG_x - \int f\, dG'_x \Bigr|
\;=\; \Bigl| \int g\, d\nu_G - \int g\, d\nu_{G'} \Bigr| ,
\qquad \text{hence} \quad
\beta( G_x, G'_x ) \le \beta( \nu_G, \nu_{G'} ) ;
\]
and $G \mapsto \nu_G$ is weakly continuous by the computation of
Lemma~\ref{lem:thm3-quotient}, Step~3, with $\Lambda_S$ in place
of the assembly map and the uniform bound
\[
\bigl\| \Lambda_{S_n}(A, c) - \Lambda_{S}(A, c) \bigr\|
\;\le\; \sum_{j \in A} c_j\, \| S_{n, j} - S_j \|
\;\le\; k\, c_{\max} \max_{j} \| S_{n, j} - S_j \|
\;\longrightarrow\; 0 .
\]
$U$ is continuous (Notation). $\Pi$ is compact, so every
subsequence of $(\hat G_M)$ has a convergent sub-subsequence; for
any subsequential limit $G_\infty$, Step~2 and continuity give
\[
\rho( G_\infty ) \;=\; \lim \rho( \hat G_{M_j} ) \;=\; 0 ,
\qquad
U( G_\infty ) \;=\; \lim U( \hat G_{M_j} )
\;\le\; U( G^{\ast} ) ,
\]
that is, $(G_\infty)_x = G^{\ast}_x$ and
$U(G_\infty) \le U(G^{\ast})$.

\emph{Step 4 (identification).} Lemma~\ref{lem:thm3-decode}(i)
upgrades $(G_\infty)_x = G^{\ast}_x$ to
$\nu_{G_\infty} = \nu_{G^{\ast}}$, and
Lemma~\ref{lem:thm3-popid} then gives
$G_\infty \approx G^{\ast}$: every subsequential limit of
$(\hat G_M)$ lies in the single class $[G^{\ast}]$.

\emph{Step 5 (quotient convergence).} Suppose
$d_G([\hat G_M], [G^{\ast}]) \not\to 0$: some $\eta > 0$ and a
subsequence keep $d_G \ge \eta$. Extract a convergent
sub-subsequence $\hat G_{M_j} \to G_\infty$ (compactness); by
Steps~3--4, $T(G_\infty) = T(G^{\ast})$, and by continuity of $T$
(Lemma~\ref{lem:thm3-quotient}),
\[
d_G\bigl( [\hat G_{M_j}], [G^{\ast}] \bigr)
\;=\; \beta\bigl( T(\hat G_{M_j}),\, T(G^{\ast}) \bigr)
\;\longrightarrow\;
\beta\bigl( T(G_\infty),\, T(G^{\ast}) \bigr) \;=\; 0 ,
\]
contradicting $d_G \ge \eta$. Hence
$d_G([\hat G_M], [G^{\ast}]) \to 0$ on the fixed almost-sure
event, which is the first claim.

\emph{Step 6 (slot separation and the radius).} By
Lemma~\ref{lem:thm3-quotient}, Step~1, every process in $\Pi$ has
pairwise $\Delta$-separated component profiles,
$\Delta = \sqrt{2}\, s_{\min}$; in particular the active true
profiles $s^{\ast}_i$, $i \in \mathrm{act}(G^{\ast})$, are
pairwise $\Delta$-separated ($S^{\ast} \in \mathcal{S}$). Fix a
radius $r$ with
\[
0 \;<\; r \;<\; \frac{\Delta}{2} ,
\qquad \text{so that} \qquad
2r \;<\; \Delta \;\le\;
\min_{\substack{i, i' \in \mathrm{act}(G^{\ast}) \\ i \ne i'}}
\bigl\| s^{\ast}_i - s^{\ast}_{i'} \bigr\| .
\]

\emph{Step 7 (balls are continuity sets).} For each
$i \in \mathrm{act}(G^{\ast})$ put
$K_i := \bar B( s^{\ast}_i, r ) \times [c_{\min}, c_{\max}]$ and
\[
E_i \;:=\; \bigl\{ \Xi \in \mathcal{K}_k :
\Xi \cap K_i \ne \emptyset \bigr\} .
\]
$E_i$ is closed, and its interior contains
$\{ \Xi : \Xi \cap ( B(s^{\ast}_i, r) \times [c_{\min}, c_{\max}]
) \ne \emptyset \}$ (an open set: the open ball times the full
coefficient range is open in $X$), so every boundary
configuration meets the closed ball but not the open one: it
carries a point whose profile is at distance exactly $r$ from
$s^{\ast}_i$. Under $T(G^{\ast})$ every carried profile is some
$s^{\ast}_{i'}$, at distance $0$ or at least $2r > r$ from
$s^{\ast}_i$, so
\[
T(G^{\ast})\bigl( \partial E_i \bigr) \;=\; 0 ,
\qquad
T(G^{\ast})( E_i )
\;=\; \mu^{\ast}( i \in A )
\;=\; \rho_i( G^{\ast} ) .
\]
By Step~5, $T(\hat G_M) \rightharpoonup T(G^{\ast})$ ($\beta$
metrizes weak convergence), so the continuity-set form of the
portmanteau theorem \citep[Theorem 2.1]{billingsley1999convergence}
gives
\[
T(\hat G_M)( E_i )
\;\longrightarrow\; \rho_i( G^{\ast} )
\;\ge\; \rho_{\min}
\qquad \text{for every } i \in \mathrm{act}(G^{\ast}) .
\]

\emph{Step 8 (matched slots).} The ball
$\bar B(s^{\ast}_i, r)$ has diameter $2r < \Delta$, so at most one
slot of $\hat G_M$ has its profile inside it (Step~6); when such a
slot $j$ exists,
\[
T(\hat G_M)( E_i )
\;=\; \hat\mu_M\bigl( \exists\, j' \in A :
S_{j'}(\hat G_M) \in \bar B(s^{\ast}_i, r) \bigr)
\;=\; \hat\mu_M( j \in A )
\;=\; \rho_j( \hat G_M ) ,
\]
and when no such slot exists, $T(\hat G_M)(E_i) = 0$. Since
$T(\hat G_M)(E_i) \to \rho_i(G^{\ast}) \ge \rho_{\min} > 0$,
eventually a (unique) matched slot $j_i(M)$ exists for every
active $i$, with
\[
\rho_{j_i(M)}( \hat G_M )
\;=\; T(\hat G_M)( E_i )
\;>\; \frac{\rho_{\min}}{2}
\qquad \text{eventually.}
\]
The $K^{\ast}$ balls are pairwise disjoint ($2r$ below the minimal
distance between active profiles), so the matched slots are
distinct across $i$: eventually at least $K^{\ast}$ slots clear
the threshold.

\emph{Step 9 (unmatched slots and the verdict).} Let
\[
C \;:=\; \Bigl\{ (s, c) \in X :
\min_{i \in \mathrm{act}(G^{\ast})}
\| s - s^{\ast}_i \| \ge r \Bigr\} ,
\qquad
E_{\mathrm{out}} \;:=\; \bigl\{ \Xi :
\Xi \cap C \ne \emptyset \bigr\} ,
\]
a closed set of configurations. Under $T(G^{\ast})$ every carried
profile is an active $s^{\ast}_i$, at distance $0$ from itself, so
$T(G^{\ast})(E_{\mathrm{out}}) = 0$, and the closed-set half of
the portmanteau theorem
\citep[Theorem 2.1]{billingsley1999convergence} gives
\[
\limsup_{M} \; T(\hat G_M)\bigl( E_{\mathrm{out}} \bigr)
\;\le\; T(G^{\ast})\bigl( E_{\mathrm{out}} \bigr) \;=\; 0 .
\]
Call a slot $j$ of $\hat G_M$ unmatched if its profile is at
distance greater than $r$ from every active $s^{\ast}_i$; then the
event $\{ j \in A \}$ forces $\Xi$ to carry the point
$( S_j(\hat G_M),\, c_j ) \in C$, so
\[
\rho_j( \hat G_M )
\;\le\; T(\hat G_M)\bigl( E_{\mathrm{out}} \bigr)
\;\longrightarrow\; 0 ,
\]
uniformly over unmatched slots: eventually every unmatched slot
has activation below $\rho_{\min}/2$. Every slot is matched to
exactly one active $i$ or unmatched (the balls are disjoint), so
eventually the slots with
$\rho_j(\hat G_M) > \rho_{\min}/2$ are exactly the $K^{\ast}$
matched ones:
\[
\hat N_M \;=\; K^{\ast}
\qquad \text{for all sufficiently large } M ,
\]
on the fixed almost-sure event. This is the second claim.
\end{proof}

\noindent
Two boundaries of part (b), stated exactly. First, the criterion:
identification rests on the distance term alone; the usage term
enters with vanishing weight as a tie-breaker among processes
reproducing the data law, so no claim is made that the true process
optimizes a fit functional at any fixed $M$, and (D4)
near-minimization is an assumption about what the optimization run
achieved, not a property certified at runtime. Second, the count:
the threshold $\rho_{\min}/2$ is defined from the unknown
$G^{\ast}$, so $\hat N_M$ is an oracle-threshold count, stated this
way to isolate the identification content; a data-driven threshold
$\tau_M \to 0$ decaying sufficiently slowly yields the same
conclusion, a separate statement not pursued here. Without any
threshold the raw count of slots with positive activation is only
lower-semicontinuous along $d_G$-convergence, and over-counting by
slots of vanishing activation cannot be excluded: this matches the
empirically observed over-selection at low SNR
(Fig.~\ref{fig:bands}).

\medskip
\noindent
\emph{What this appendix establishes.} The two parts return the
answer promised at the head of the appendix: the object of
inference is the right one. Part (a) shows that per-sample
decoding is a finite-sample notion by necessity: under noise, no
estimator decodes every instantiation, at any data size. Part (b)
shows the process itself is delivered without residue: the
estimated process converges almost surely to the truth up to
relabeling, and the estimated component count is eventually
exactly the true count. Together the theorems close the
framework's account of what can be known from mixtures: existence
(\S\ref{sec:gmcr}), findability (Theorem~1), recognizability at
finite samples (Theorem~2), and the exact asymptotic deliverable
(Theorem~3). The component count the experiments report is the
quantity part (b) proves eventually exact, and reuse of a
recovered process (Corollary~1) is meaningful precisely because
the process is the object that accumulating data pin down.

\section{Formal Statement and Proof of Proposition~2 (Two-Phase
Convergence)}
\label{app:dyn}

This appendix states and proves Proposition~2 in full. Its
subject is the one object about which Theorems~1--3 are
deliberately silent: the running solver itself. During training,
the solver revises which components it selects and the
concentrations it assigns to them; early in a run the selected
component set changes often, and at some point it stops changing.
The practical question is whether a stable selection can be
trusted. Proposition~2 identifies, from quantities the solver
already computes, a condition under which it can: once the
condition holds, the selected components provably remain fixed
for a computable number of steps while the reconstruction error
decreases toward an explicit floor, and a checkpoint taken inside
this period records a stable selection rather than a transient
one.

In the vocabulary of the model, the condition is an
\emph{anchor event}: at some step $t_0$, every component's empirical
energy margin (the dataset mean of the gate's per-sample energy
gap, signed so that positive means agreement with the selection
made at $t_0$) clears a level $\gamma$. Two phases follow. In the
first, the anchor opens a \emph{support-stability window} of
explicit length $T(\gamma)$: for that many steps the selected
support cannot change, for any realization of the data and of the
training noise, because Lipschitz continuity of the energies and
the bounded optimizer step bound how fast margins of size $\gamma$ can decrease. In the second, the objective with the gate frozen at
the anchored support contracts linearly, in expectation, to an
explicit noise floor. Once the objective gap falls below an
explicit basin level $\varepsilon^{\star}$, the small gap forces
every margin positive again, so the support becomes
self-sustaining. The gate re-selects the anchored support with
quantified failure probability, at each fixed time and uniformly
over any finite horizon.

Theorems~1--3 state what a returned solution means; this
proposition explains why the solver is a dependable way to reach
one. What a practitioner observes in a run, selections
stabilizing, the fit descending, the stabilized selection
persisting, is proved here to be the mechanism's expected
behavior: after an observed anchor event, the selected support
holds for an explicit horizon, the objective contracts toward an
explicit floor, and the support re-selects itself with quantified
risk. The guarantee is conditional and finite-horizon by design:
it activates at the anchor (whether a run reaches one is read off
the run itself), it rests on stated assumptions about the local
geometry of the objective, and whether the stabilized support is
the true one is decided by Theorem~\ref{thm:cert}'s certificate,
with the variational content supplied by
Theorem~\ref{thm:ident}(b) through Lemma~\ref{lem:dyn-gate}. The
proposition is the reason the checkpointing rule of
\S\ref{sec:solver-checkpoint} samples a meaningful object: after
an anchor, checkpoints within the window all record the same
support, and the risk of a later support flip is quantified per
unit horizon. Even where the conditions for exact recovery fail,
the run reaches a stable, checkable object rather than a
transient fit.

Three lemmas carry the proofs.
Lemma~\ref{lem:dyn-gate} grounds the margin levels the
assumptions demand and confines all dependence on the sample to
one event: the flip gap of Theorem~\ref{thm:ident}(b) yields
per-sample energy margins at some parameter, under an
expressivity hypothesis (part (i)); an aggregation argument
transfers them to the dataset-mean gate whenever component
prevalences avoid an explicit intermediate band (part (ii)); and
a net argument makes the deviation of the empirical gaps uniform
over the training region (part (iii)), with a scope remark
recording at which scales its constants bind.
Lemma~\ref{lem:dyn-window} proves the certified stability
window. Lemma~\ref{lem:dyn-inside} proves linear contraction of
the frozen objective with bias, down to a noise floor, under the
Polyak--{\L}ojasiewicz inequality (part (i)), converts a small
objective gap into selection margins (part (ii)), bounds the
per-time-point risk (part (iii)), and makes the risk bound
uniform in time by a maximal inequality (part (iv)). The
assembly takes five steps.

\paragraph{Notation.}
The notation of Appendices~\ref{app:prop1}--\ref{app:thm3}
remains in force. New objects of the gate and the window:
\begin{center}
\begin{tabular}{@{}ll@{}}
\toprule
$\theta_t,\ \Theta,\ R,\ D$ & training iterates; the region they
visit, $\Theta \subseteq B(\theta_{\mathrm{ref}}, R)$, a ball\\
& of radius $R$ fixed in advance; $D$ the parameter dimension of
the gate\\
$e^{\mathrm{sel}}_j(x;\theta),\ e^{\mathrm{rej}}_j(x;\theta)$ &
the gate's two selection energies for component $j$ at sample
$x$\\
& (row $j$ of $E$, \S\ref{sec:solver-ebselect})\\
$g_j(x;\theta)$ & $= e^{\mathrm{rej}}_j(x;\theta) -
e^{\mathrm{sel}}_j(x;\theta)$, the per-sample energy gap\\
$\hat\Delta_j(\theta),\ \Delta_j(\theta)$ &
$= \tfrac1M \sum_{m} g_j(x^{(m)};\theta)$ and
$= \mathbb{E}_{x \sim \mathcal{D}}\, g_j(x;\theta)$, the mean
gaps\\
$\hat J_t$ & $= \{ j \in [N] : \hat\Delta_j(\theta_t) > 0 \}$,
the evaluation support at step $t$\\
$J_{\sharp},\ t_0,\ \gamma$ & anchored support
$\hat J_{t_0}$; anchor time; anchor level\\
$m_j(\theta),\ \hat m_j(\theta)$ & signed margins:
$\Delta_j(\theta)$, $\hat\Delta_j(\theta)$ for
$j \in J_{\sharp}$, their negatives\\
& for $j \notin J_{\sharp}$\\
$U,\ L_e,\ L_\Delta$ & energy bound; energy Lipschitz constant;
$L_\Delta = 2 L_e$\\
$G,\ \eta_t,\ \eta$ & update-displacement constant; step sizes
(constant $\eta$ after $t_0$)\\
$T(\gamma)$ & $= \sup\{ T \in \mathbb{N} :
L_\Delta G \sum_{s=t_0}^{t_0+T-1} \eta_s \le \gamma/4 \}$, the
stability horizon\\
$E_{\mathrm{gap}}$ & the concentration event of
Lemma~\ref{lem:dyn-gate}(iii)\\
$\mathcal{F}_t$ & the $\sigma$-algebra of all randomness up to step $t$
(the data included)\\
\bottomrule
\end{tabular}
\end{center}
Objects of the descent analysis and of the margin grounding:
\begin{center}
\begin{tabular}{@{}ll@{}}
\toprule
$f,\ f^{\star},\ \mathcal{X}^{\star},\ \mathcal{U}$ & objective
with the gate frozen at $J_{\sharp}$; its infimum over a\\
& neighborhood $\mathcal{U} \subseteq \Theta$; its minimizer set
in $\mathcal{U}$\\
$L,\ \mu$ & gradient-Lipschitz and Polyak--{\L}ojasiewicz
constants of $f$ on $\mathcal{U}$\\
$\beta,\ \nu^2$ & gradient bias bound; gradient variance bound\\
$\Phi(\eta)$ & $= 2\beta^2/\mu + L \eta (\nu^2 + 2\beta^2)/\mu$,
the noise floor\\
$\gamma^{\star},\ \varepsilon^{\star}$ & minimizer margin level;
$\varepsilon^{\star} = \mu (\gamma^{\star})^2 / (8 L_\Delta^2)$,
the basin level\\
$z_t,\ c_\eta$ & objective gap $f(\theta_t) - f^{\star}$;
per-step injection $\eta \beta^2 + \tfrac{L}{2} \eta^2 (\nu^2 +
2\beta^2)$\\
$T_{\mathrm{ctr}}$ & $= \bigl\lceil \tfrac{2}{\mu\eta}
\log\bigl( 2 (f(\theta_{t_0}) - f^{\star}) / \varepsilon^{\star}
\bigr) \bigr\rceil$, the contraction length\\
$g_{\mathrm{gap}}$ & flip gap of Theorem~\ref{thm:ident}(b):
$\min\{ \lambda - \sigma^2 L_{\alpha_{\mathrm{w}}}^2,\,
(c_{\min} s_{\min} - \sigma L_{\alpha_{\mathrm{w}}})^2 - \lambda
\}$\\
$\theta^{\dagger},\ a$ & margin-realizing parameter
(Lemma~\ref{lem:dyn-gate}(i)); margin level
$a = g_{\mathrm{gap}}/2$\\
$p_j,\ \hat\rho_j$ & prevalence
$\Pr_{x \sim \mathcal{D}}(j \in J(x))$, $J(x)$ the generative\\
& support of $x$; its empirical counterpart\\
\bottomrule
\end{tabular}
\end{center}
The uniform deviation level of the empirical gaps, established in
Lemma~\ref{lem:dyn-gate}(iii), is
\begin{equation}
r_M(\alpha) \;=\;
2U \sqrt{ \frac{2 \bigl( \log\tfrac{2N}{\alpha}
+ D \log\bigl( 3 + \tfrac{2 R L_\Delta \sqrt{M}}{U} \bigr)
\bigr)}{M} }
\;+\; \frac{2U}{\sqrt{M}} .
\label{eq:dyn-rM}
\end{equation}

Four conventions. First, symbol reuse. Within this appendix $D$
is the parameter dimension of the gate network; the data matrix
of Eq.~\ref{eq:1} does not appear. Likewise $\mu$, $\beta$, $U$,
$G$, and $L$ denote the objects of the tables above, unrelated to
the latent law, metric, usage functional, process, and criterion
of Appendix~\ref{app:thm3}; the noise event $E$ of
Appendix~\ref{app:thm2} is not used; and $T$, $T(\gamma)$,
$T_{\mathrm{ctr}}$ are step counts, not index sets. The letter
$g$ follows its subscript: $g_j(x;\theta)$, with component
subscript, is a gate output, and $g_t$, with time subscript, is
the stochastic update direction of (E6); the deterministic
gradient is always written $\nabla f$. Second,
observability. A hat marks a quantity computable from the data
and the current iterate at run time ($\hat\Delta_j$, $\hat m_j$,
$\hat J_t$, $\hat\rho_j$); unhatted counterparts are population
quantities. The anchor event is a condition on hatted quantities
only, and $T(\gamma)$ is computable from architecture-level
constants and the step schedule; by contrast $\gamma^{\star}$,
$\mu$, $L$, $\beta$, $\nu^2$, $f^{\star}$, and therefore
$\varepsilon^{\star}$, $\Phi(\eta)$, and $T_{\mathrm{ctr}}$, are
properties of the objective and its geometry, not computable at
run time. Third, the gap sign matches the gate: the evaluation
gate selects component $j$ exactly when the selected state has
the lower energy, so positive $g_j$ favors selection, and
$\hat m_j(\theta_t) > 0$ for all $j$ implies
$\hat J_t = J_{\sharp}$ (ties $\hat\Delta_j = 0$ break the
converse; only this direction is used). Fourth, sums over an
empty index range are zero, so $T(\gamma) \ge 0$ always;
$T(\gamma) = \infty$ occurs for summable step schedules and only
strengthens every conclusion, and for non-constant schedules
$T(\gamma)$ depends on $t_0$.

Seven standing assumptions.
\begin{itemize}[leftmargin=*]
\item \emph{(E1)} (sampling) $x^{(1)}, \dots, x^{(M)}$ are
i.i.d.\ draws from a distribution $\mathcal{D}$ on $\R^d$.
\item \emph{(E2)} (bounded, Lipschitz energies) There are finite
$U, L_e > 0$ and a ball $B(\theta_{\mathrm{ref}}, R)$ whose
center and radius are fixed in advance, independently of the data
and of the training noise, such that all training iterates lie in
$\Theta \subseteq B(\theta_{\mathrm{ref}}, R)$ surely, and for
all $j \in [N]$, both energy states
$s \in \{\mathrm{sel}, \mathrm{rej}\}$, all $x$ in the support of
$\mathcal{D}$, and all $\theta, \theta' \in \Theta$,
\[
\bigl| e^{s}_j(x;\theta) \bigr| \le U,
\qquad
\bigl| e^{s}_j(x;\theta) - e^{s}_j(x;\theta') \bigr|
\le L_e \| \theta - \theta' \| .
\]
Consequently $|g_j| \le 2U$ and $g_j(x;\cdot)$,
$\hat\Delta_j(\cdot)$, $\Delta_j(\cdot)$ are
$L_\Delta$-Lipschitz on $\Theta$ with $L_\Delta = 2 L_e$.
\item \emph{(E3)} (bounded update displacement)
$\| \theta_{t+1} - \theta_t \| \le \eta_t\, G$ for every training
step $t$.
\item \emph{(E4)} (anchor certificate) There exist a time $t_0$
and a level $\gamma > 0$ with
$\hat m_j(\theta_{t_0}) \ge \gamma$ for all $j \in [N]$, the
margins signed relative to $J_{\sharp} = \hat J_{t_0}$.
\item \emph{(E5)} (geometry after the anchor) The objective $f$
with the gate frozen at $J_{\sharp}$ (reconstruction on the
selected components plus the smooth regularizers of
Eq.~\ref{eq:9}; the usage penalty is constant once the gate is
frozen and is omitted) has $L$-Lipschitz gradient on a
neighborhood $\mathcal{U} \subseteq \Theta$ containing the
post-anchor iterates, satisfies there the
Polyak--{\L}ojasiewicz inequality
$\tfrac12 \| \nabla f(\theta) \|^2 \ge
\mu (f(\theta) - f^{\star})$ with $\mu > 0$ and
$\mathcal{X}^{\star} \ne \emptyset$, and satisfies the
quadratic-growth bound
$f(\theta) - f^{\star} \ge \tfrac{\mu}{2}
\dist(\theta, \mathcal{X}^{\star})^2$ on $\mathcal{U}$.
\item \emph{(E6)} (stochastic gradient model after the anchor)
For $t \ge t_0$ the update is
$\theta_{t+1} = \theta_t - \eta_t g_t$ with
\[
\mathbb{E}[ g_t \mid \mathcal{F}_t ] = \nabla f(\theta_t) + b_t,
\qquad
\| b_t \| \le \beta,
\qquad
\mathbb{E}\bigl[ \| g_t - \mathbb{E}[g_t \mid \mathcal{F}_t] \|^2
\,\big|\, \mathcal{F}_t \bigr] \le \nu^2 .
\]
\item \emph{(E7)} (margin regularity of minimizers) There is
$\gamma^{\star} > 0$ such that every
$\bar\theta \in \mathcal{X}^{\star}$ satisfies
$m_j(\bar\theta) \ge \gamma^{\star}$ for all $j \in [N]$.
\end{itemize}
The seven conditions differ in status: some hold by
construction, some are observable during a run, and some are
assumptions about the run. The comments below state which is
which. First, (E4) is the only condition on the run
that is observable: $\hat\Delta_j(\theta_t)$ is computable from
the forward passes the solver already performs. The implemented
checkpointing records $R^2$, nMSE, and usage rather than margin
means (Appendix~\ref{app:b.2}), so realizing the anchor
certificate takes one additional logged quantity; (E4) is then a
logging predicate, not an unverifiable assumption. The analysis does not
prove that training reaches the anchor; it proves what follows
once the anchor is observed.

Second, (E1) is a modeling
assumption, and it is the only distributional one: no tail
condition on the data is needed anywhere in this appendix,
because the boundedness in (E2) already yields Hoeffding-type
concentration regardless of the data law; the gMCR model of
Eq.~\ref{eq:model} enters only through the grounding results of
Lemma~\ref{lem:dyn-gate}.

Third, (E2) and (E3) are verifiable from the
architecture: the gate is a three-layer $\tanh$ network, so its
outputs are bounded and Lipschitz on any bounded weight set, with
the constants controlled in practice by the energy regularizer
$\lambda_{\mathrm{se}} \| E \|_2^2$ and by weight decay; and an
Adam-type update moves each coordinate by at most $\eta_t$ (its
stabilized update ratio is bounded by one), so (E3) holds
deterministically with $G = \sqrt{D}$; weight decay enlarges $G$
slightly.

Fourth,
(E5)--(E7) are assumptions about the run, in force for all
$t \ge t_0$: the analysis models post-anchor training as
stochastic descent on the frozen objective throughout, an
idealization of the implemented sampler whose recognized cost is
the bias term $\beta$ in (E6). That bias is the gap between the
relaxed-gate gradient at temperature $\tau > 0$ and the gradient
of $f$; it vanishes as $\tau \to 0$. $\nu^2$ collects mini-batch,
Gumbel-resampling, and energy-exploration noise. In the global
setting the Polyak--{\L}ojasiewicz inequality implies quadratic
growth \citep[Theorem~2]{karimi2016linear}; on a neighborhood the
implication additionally requires $\mathcal{U}$ to contain the
relevant sublevel set, so (E5) hypothesizes the quadratic-growth
bound directly.

Fifth, the gate itself is idealized: the
evaluation support $\hat J_t$ reads the sign of dataset-mean
gaps, whereas the implemented gate thresholds per-sample
selection probabilities at $\tau_{\mathrm{thres}}$, which at
$\tau_{\mathrm{eval}} = 0.01$ is a per-sample energy-gap floor of
about $0.14$ (Appendix~\ref{app:gate}).
Lemma~\ref{lem:dyn-gate} states
exactly what grounds this idealization; whether training reaches
the grounding parameter remains assumed.

\paragraph{Proposition~2, restated.}
Fix $\alpha \in (0,1)$ and $p \in (0, \tfrac12]$. Suppose
(E1)--(E3) hold; suppose (E4) holds with anchor level $\gamma$ at
time $t_0$ and anchored support $J_{\sharp} = \hat J_{t_0}$;
suppose (E5)--(E7) hold for all $t \ge t_0$, with every iterate
$\theta_t$, $t \ge t_0$, and every segment
$[\theta_t, \theta_{t+1}]$ contained in $\mathcal{U}$, and with a
constant step $\eta \le \tfrac{1}{2L}$ for $t \ge t_0$. All
expectations and probabilities over the training randomness below
are conditional on $\mathcal{F}_{t_0}$. Suppose the compatibility
conditions
\begin{equation}
r_M(\alpha) \;\le\; \frac{\gamma}{4},
\label{eq:dyn-c1}
\end{equation}
\begin{equation}
r_M(\alpha) \;\le\; \frac{\gamma^{\star}}{4},
\label{eq:dyn-c2}
\end{equation}
\begin{equation}
\Phi(\eta) \;\le\; p\, \varepsilon^{\star},
\label{eq:dyn-c3}
\end{equation}
\begin{equation}
T(\gamma) \;\ge\; T_{\mathrm{ctr}},
\qquad \text{with } T_{\mathrm{ctr}} = 0 \text{ when }
f(\theta_{t_0}) - f^{\star} \le \varepsilon^{\star}/2 .
\label{eq:dyn-c4}
\end{equation}
Let $E_{\mathrm{gap}}$ be the event of
Lemma~\ref{lem:dyn-gate}(iii) at level $\alpha$, so
$\Pr(E_{\mathrm{gap}}) \ge 1 - \alpha$ over the sample. Then:

\emph{(i) (Certified block.)} Deterministically, for every realization of
the sample and of the training noise,
\[
\hat J_t = J_{\sharp}
\qquad \text{and} \qquad
\hat m_j(\theta_t) \ge \frac{3\gamma}{4}
\qquad
\text{for all } j \in [N],\;
t \in [t_0,\; t_0 + T(\gamma)] ;
\]
on $E_{\mathrm{gap}}$, under (\ref{eq:dyn-c1}), also
$m_j(\theta_t) \ge \gamma/2$ for all such $j, t$.

\emph{(ii) (Contraction.)} For $0 \le k \le T(\gamma)$,
\[
\mathbb{E}\bigl[ f(\theta_{t_0+k}) - f^{\star} \bigr]
\;\le\;
\Bigl( 1 - \frac{\mu\eta}{2} \Bigr)^{k}
\bigl( f(\theta_{t_0}) - f^{\star} \bigr) + \Phi(\eta),
\]
and in particular
$\mathbb{E}[ f(\theta_{t_0+T_{\mathrm{ctr}}}) - f^{\star} ]
\le (\tfrac12 + p)\, \varepsilon^{\star}
\le \varepsilon^{\star}$.

\emph{(iii) (Self-sustaining support.)} For every fixed
$t \ge t_0 + T_{\mathrm{ctr}}$,
\[
\Pr\bigl( f(\theta_t) - f^{\star} > \varepsilon^{\star} \bigr)
\;\le\;
p + \frac12
\Bigl( 1 - \frac{\mu\eta}{2} \Bigr)^{t - t_0 - T_{\mathrm{ctr}}},
\]
and on every sample realization in $E_{\mathrm{gap}}$, under
(\ref{eq:dyn-c2}), the same bound holds for
$\Pr\bigl( \hat J_t \ne J_{\sharp} \bigr)$.

\emph{(iv) (Uniform permanence.)} For every horizon
$T \in \mathbb{N}$,
\[
\Pr\Bigl( \exists\, t \in
[\, t_0 + T_{\mathrm{ctr}},\; t_0 + T_{\mathrm{ctr}} + T \,] :
f(\theta_t) - f^{\star} \ge \varepsilon^{\star} \Bigr)
\;\le\;
\frac12 + p + p\, \frac{\mu\eta}{2}\, T ,
\]
and on every sample realization in $E_{\mathrm{gap}}$, under
(\ref{eq:dyn-c2}), the same bound holds for
$\Pr\bigl( \exists\, t \in
[\, t_0 + T_{\mathrm{ctr}},\; t_0 + T_{\mathrm{ctr}} + T \,] :
\hat J_t \ne J_{\sharp} \bigr)$. For every
$\delta \in (p, 1)$, replacing $T_{\mathrm{ctr}}$ by
$T_{\mathrm{ctr}}(\delta) = \bigl\lceil \tfrac{2}{\mu\eta}
\log_{+}\bigl( (f(\theta_{t_0}) - f^{\star}) /
((\delta - p)\, \varepsilon^{\star}) \bigr) \bigr\rceil$, with
$\log_{+} = \max\{ \log, 0 \}$,
replaces $\tfrac12 + p$ by $\delta$.

\bigskip
\noindent
Lemma~\ref{lem:dyn-gate}
grounds the margin levels and confines the sample randomness;
Lemma~\ref{lem:dyn-window} proves part (i);
Lemma~\ref{lem:dyn-inside} proves parts (ii)--(iv); the assembly
closes.

The first lemma connects the dynamics to the identifiable target
of Theorem~\ref{thm:ident} and confines all dependence on the
sample to one event: the flip gap of Theorem~\ref{thm:ident}(b)
(Appendix~\ref{app:thm1})
produces per-sample energy margins at some parameter
$\theta^{\dagger}$ under an expressivity hypothesis, and
prevalence separation transfers them to the dataset-mean gate
that defines $\hat J_t$, so the levels demanded by (E4) and (E7)
are attainable, not merely posited; whether training reaches
such a parameter is exactly the anchor event, and is not proved.
The uniformity of the deviation bound over the whole training
region is a necessity, not a refinement: the iterates $\theta_t$
are functions of the same $M$ samples that define
$\hat\Delta_j$, so concentration applied at $\theta_t$ pointwise
would be circular, and a union bound over time would tie the
sample size to the window length; a net argument over $\Theta$
removes both defects at once, with a bound that holds at every
point of every trajectory contained in $\Theta$.

\begin{lemma}[The empirical gate and the data]
\label{lem:dyn-gate}
\emph{(i) (From flip gap to per-sample margins.)}
Assume the gMCR model (Eq.~\ref{eq:model}) with the spark
condition (Definition~\ref{def:spark}). Fix
$\alpha_{\mathrm{w}} \in (0,1)$, set
$L_{\alpha_{\mathrm{w}}} = \sqrt{2 \log(6NM/\alpha_{\mathrm{w}})}$,
and assume the SNR condition and window of
Theorem~\ref{thm:ident}(b):
$c_{\min} s_{\min} > 2 \sigma L_{\alpha_{\mathrm{w}}}$ and
$\sigma^2 L_{\alpha_{\mathrm{w}}}^2 < \lambda <
( c_{\min} s_{\min} - \sigma L_{\alpha_{\mathrm{w}}} )^2$, so that
$g_{\mathrm{gap}} > 0$. Define, for each sample $m$ and each
$j \in [N]$, the ideal signed gaps
\[
\Delta^{\mathrm{idl}}_j\bigl( x^{(m)} \bigr) \;=\;
\begin{cases}
+\bigl[\, \ell_\lambda\bigl( J_m \setminus \{j\};\, x^{(m)}
\bigr) - \ell_\lambda\bigl( J_m;\, x^{(m)} \bigr) \,\bigr],
& j \in J_m,\\[3pt]
-\bigl[\, \ell_\lambda\bigl( J_m \cup \{j\};\, x^{(m)}
\bigr) - \ell_\lambda\bigl( J_m;\, x^{(m)} \bigr) \,\bigr],
& j \notin J_m,
\end{cases}
\]
with $\ell_\lambda$ as in Eq.~\ref{eq:penobj}. On the event of
Theorem~\ref{thm:ident}(b), of probability at least
$1 - \alpha_{\mathrm{w}}$ over the noise:
\emph{(a)}
$\Delta^{\mathrm{idl}}_j( x^{(m)} ) \ge g_{\mathrm{gap}}$ for
$j \in J_m$ and
$\Delta^{\mathrm{idl}}_j( x^{(m)} ) \le -g_{\mathrm{gap}}$ for
$j \notin J_m$, for every $m$;
\emph{(b)} if some $\theta^{\dagger} \in \Theta$ satisfies the
expressivity hypothesis
\begin{equation}
\bigl| g_j\bigl( x^{(m)};\, \theta^{\dagger} \bigr) -
\Delta^{\mathrm{idl}}_j\bigl( x^{(m)} \bigr) \bigr|
\;\le\; \frac{g_{\mathrm{gap}}}{2}
\qquad \text{for every } m \in [M],\; j \in [N],
\label{eq:dyn-e1}
\end{equation}
then
$g_j( x^{(m)}; \theta^{\dagger} ) \ge g_{\mathrm{gap}}/2$ for
$j \in J_m$ and
$g_j( x^{(m)}; \theta^{\dagger} ) \le -g_{\mathrm{gap}}/2$ for
$j \notin J_m$, for every $m$.

\emph{(ii) (Mean-gate aggregation.)}
In the setting of part (i), with (E1) and
(E2), write $a = g_{\mathrm{gap}}/2$ and let
$p_j = \Pr_{x \sim \mathcal{D}}( j \in J(x) )$ be the prevalence
of component $j$, where $J(x)$ is the generative support of $x$.
Fix $\gamma > 0$ and $\kappa > 0$, and suppose every $j \in [N]$
satisfies either
\begin{equation}
p_j \;\ge\; \frac{2U + \gamma}{a + 2U} + \kappa
\quad (\text{the prevalent set } S^{\star})
\qquad \text{or} \qquad
p_j \;\le\; \frac{a - \gamma}{a + 2U} - \kappa .
\label{eq:dyn-ps}
\end{equation}
The prevalent alternative is satisfiable precisely when
$\gamma + \kappa( a + 2U ) \le a$. Then with probability at least
$1 - \alpha_{\mathrm{w}} - 2N \exp( -2M\kappa^2 )$ over the
joint draw of the samples and their noise,
\[
\hat\Delta_j( \theta^{\dagger} ) \ge \gamma
\quad ( j \in S^{\star} ),
\qquad
\hat\Delta_j( \theta^{\dagger} ) \le -\gamma
\quad ( j \notin S^{\star} ) ;
\]
in particular the evaluation gate at $\theta^{\dagger}$ returns
exactly $S^{\star}$, and the anchor certificate (E4) holds at
$\theta^{\dagger}$ with level $\gamma$ and reference support
$S^{\star}$, i.e., (E4) is satisfied at any step whose iterate
is $\theta^{\dagger}$.

\emph{(iii) (Uniform deviation of empirical gaps.)}
Let (E1) and (E2) hold. For any $\rho \in (0, R]$ and
$\alpha \in (0,1)$, with probability at least $1 - \alpha$ over
the sample,
\[
\sup_{\theta \in \Theta}\; \max_{j \in [N]}\;
\bigl| \hat\Delta_j(\theta) - \Delta_j(\theta) \bigr|
\;\le\;
2U \sqrt{ \frac{2 \bigl( \log\tfrac{2N}{\alpha}
+ D \log\bigl( 1 + \tfrac{2R}{\rho} \bigr) \bigr)}{M} }
\;+\; 2 L_\Delta\, \rho .
\]
The choice $\rho^{\star} = \min\{ R,\; U / ( L_\Delta \sqrt{M} )
\}$ bounds the right side by $r_M(\alpha)$ of
Eq.~\ref{eq:dyn-rM}. $E_{\mathrm{gap}}$ denotes the event of the
display at $\rho = \rho^{\star}$.
\end{lemma}

\begin{proof}
\emph{Part (i).}

\emph{Step 1 (gap signs).} On the event of
Theorem~\ref{thm:ident}(b), every selection obtained from $J_m$
by adding or removing one component costs at least
$g_{\mathrm{gap}}$:
\[
\ell_\lambda\bigl( A;\, x^{(m)} \bigr)
- \ell_\lambda\bigl( J_m;\, x^{(m)} \bigr)
\;\ge\; g_{\mathrm{gap}}
\qquad \text{whenever } | A \,\triangle\, J_m | = 1 .
\]
For $j \in J_m$ the set $A = J_m \setminus \{j\}$ satisfies
$| A \,\triangle\, J_m | = 1$, so
\[
\Delta^{\mathrm{idl}}_j\bigl( x^{(m)} \bigr)
\;=\;
\ell_\lambda\bigl( J_m \setminus \{j\};\, x^{(m)} \bigr)
- \ell_\lambda\bigl( J_m;\, x^{(m)} \bigr)
\;\ge\; g_{\mathrm{gap}} ;
\]
for $j \notin J_m$ the set $A = J_m \cup \{j\}$ satisfies
$| A \,\triangle\, J_m | = 1$, so
\[
\Delta^{\mathrm{idl}}_j\bigl( x^{(m)} \bigr)
\;=\;
-\bigl[\, \ell_\lambda\bigl( J_m \cup \{j\};\, x^{(m)} \bigr)
- \ell_\lambda\bigl( J_m;\, x^{(m)} \bigr) \,\bigr]
\;\le\; -g_{\mathrm{gap}} .
\]

\emph{Step 2 (realization).} For $j \in J_m$, by the triangle
inequality, (\ref{eq:dyn-e1}), and Step 1,
\[
g_j\bigl( x^{(m)};\, \theta^{\dagger} \bigr)
\;\ge\;
\Delta^{\mathrm{idl}}_j\bigl( x^{(m)} \bigr)
- \frac{g_{\mathrm{gap}}}{2}
\;\ge\;
g_{\mathrm{gap}} - \frac{g_{\mathrm{gap}}}{2}
\;=\; \frac{g_{\mathrm{gap}}}{2} ,
\]
and for $j \notin J_m$, by the same three facts,
\[
g_j\bigl( x^{(m)};\, \theta^{\dagger} \bigr)
\;\le\;
\Delta^{\mathrm{idl}}_j\bigl( x^{(m)} \bigr)
+ \frac{g_{\mathrm{gap}}}{2}
\;\le\;
- g_{\mathrm{gap}} + \frac{g_{\mathrm{gap}}}{2}
\;=\; - \frac{g_{\mathrm{gap}}}{2} .
\]

\emph{Part (ii).}
Write $\chi^{(m)}_j = \mathbf{1}[\, j \in J_m \,]$ and
$\hat\rho_j = \tfrac1M \sum_m \chi^{(m)}_j$.

\emph{Step 3 (pointwise sandwich).} On the event of part (i), for every $m$ and $j$,
\[
( a + 2U )\, \chi^{(m)}_j - 2U
\;\le\;
g_j\bigl( x^{(m)};\, \theta^{\dagger} \bigr)
\;\le\;
( a + 2U )\, \chi^{(m)}_j - a .
\]
Both cases check directly. For $\chi^{(m)}_j = 1$ the claim
reads
\[
a \;\le\; g_j\bigl( x^{(m)};\, \theta^{\dagger} \bigr)
\;\le\; 2U ,
\]
the left side by part (i)(b) and the right by (E2); for
$\chi^{(m)}_j = 0$ it reads
\[
-2U \;\le\; g_j\bigl( x^{(m)};\, \theta^{\dagger} \bigr)
\;\le\; -a ,
\]
the left side by (E2) and the right by part (i)(b).
Averaging over $m$,
\[
\hat\rho_j\, ( a + 2U ) - 2U
\;\le\;
\hat\Delta_j( \theta^{\dagger} )
\;\le\;
\hat\rho_j\, ( a + 2U ) - a .
\]

\emph{Step 4 (deterministic gate).} If
$\hat\rho_j \ge ( 2U + \gamma ) / ( a + 2U )$, then by Step 3
\[
\hat\Delta_j( \theta^{\dagger} )
\;\ge\;
\hat\rho_j\, ( a + 2U ) - 2U
\;\ge\;
( 2U + \gamma ) - 2U
\;=\; \gamma ,
\]
and if $\hat\rho_j \le ( a - \gamma ) / ( a + 2U )$, then by
Step 3
\[
\hat\Delta_j( \theta^{\dagger} )
\;\le\;
\hat\rho_j\, ( a + 2U ) - a
\;\le\;
( a - \gamma ) - a
\;=\; -\gamma .
\]

\emph{Step 5 (prevalence concentration).} For fixed $j$ the
indicators $\chi^{(m)}_j$, $m \in [M]$, are i.i.d.\
$\{0,1\}$-valued with mean $p_j$ by (E1), so Hoeffding's
inequality \citep[Theorem~2]{hoeffding1963probability} gives
\[
\Pr\bigl( | \hat\rho_j - p_j | \ge \kappa \bigr)
\;\le\; 2 \exp( -2M\kappa^2 ) .
\]

\emph{Step 6 (union bound).} Off the union of the $N$ events of
Step 5 and the complement of the event of part (i) (the failure
probabilities add),
$| \hat\rho_j - p_j | < \kappa$ for every $j$ simultaneously.
For $j \in S^{\star}$, by (\ref{eq:dyn-ps}),
\[
\hat\rho_j
\;>\;
p_j - \kappa
\;\ge\;
\frac{2U + \gamma}{a + 2U} ,
\]
and for $j \notin S^{\star}$, by (\ref{eq:dyn-ps}),
\[
\hat\rho_j
\;<\;
p_j + \kappa
\;\le\;
\frac{a - \gamma}{a + 2U} ,
\]
so Step 4 applies to every $j$. Since
$\gamma > 0$, the sign pattern of
$\hat\Delta_j( \theta^{\dagger} )$ is exactly $S^{\star}$, which
is the gate identity, and the margins clear $\gamma$, which is
(E4) at $\theta^{\dagger}$.

\emph{Part (iii).}

\emph{Step 7 (net).} Fix a $\rho$-net $\mathcal{N}_\rho$ of
$B(\theta_{\mathrm{ref}}, R)$ with
\[
| \mathcal{N}_\rho | \;\le\;
\Bigl( 1 + \frac{2R}{\rho} \Bigr)^{D}
\]
\citep[\S 4.2]{vershynin2018high}. The net is fixed in advance,
because the ball is data-independent by (E2).

\emph{Step 8 (tail at one net point).} Fix
$( j, \bar\theta ) \in [N] \times \mathcal{N}_\rho$. By (E2),
$g_j( x; \bar\theta ) \in [-2U, 2U]$; by (E1) the summands of
$\hat\Delta_j( \bar\theta )$ are i.i.d.; and $\bar\theta$ is a
fixed, data-independent point. Hoeffding's inequality
\citep[Theorem~2]{hoeffding1963probability} therefore applies:
\[
\Pr\Bigl( \bigl| \hat\Delta_j( \bar\theta )
- \Delta_j( \bar\theta ) \bigr| \ge u \Bigr)
\;\le\;
2 \exp\Bigl( - \frac{2 M u^2}{(4U)^2} \Bigr)
\;=\;
2 \exp\Bigl( - \frac{M u^2}{8 U^2} \Bigr) .
\]

\emph{Step 9 (union bound and level).} Set
\[
u \;=\; 2U \sqrt{ \frac{2 \bigl( \log\tfrac{2N}{\alpha}
+ \log | \mathcal{N}_\rho | \bigr)}{M} } ,
\qquad \text{so} \qquad
2 \exp\Bigl( - \frac{M u^2}{8U^2} \Bigr)
\;=\; \frac{\alpha}{N | \mathcal{N}_\rho |} ,
\]
and a union bound over the $N | \mathcal{N}_\rho |$ pairs makes
the total failure probability at most $\alpha$.

\emph{Step 10 (Lipschitz transfer).} On the complementary event,
every $\theta \in \Theta$ has a net point $\bar\theta$ with
$\| \theta - \bar\theta \| \le \rho$, and since both
$\hat\Delta_j$ and $\Delta_j$ are $L_\Delta$-Lipschitz on
$\Theta$ by (E2) (the empirical bound holds surely for every
realization of the sample),
\[
\bigl| \hat\Delta_j(\theta) - \Delta_j(\theta) \bigr|
\;\le\;
\bigl| \hat\Delta_j(\theta) - \hat\Delta_j(\bar\theta) \bigr|
+ \bigl| \hat\Delta_j(\bar\theta) - \Delta_j(\bar\theta) \bigr|
+ \bigl| \Delta_j(\bar\theta) - \Delta_j(\theta) \bigr|
\;\le\;
u + 2 L_\Delta \rho .
\]

\emph{Step 11 (canonical choice).} If
$\rho^{\star} = U / ( L_\Delta \sqrt{M} ) \le R$, the transfer
term and the net argument satisfy
\[
2 L_\Delta \rho^{\star} \;=\; \frac{2U}{\sqrt{M}} ,
\qquad
1 + \frac{2R}{\rho^{\star}}
\;=\;
1 + \frac{2 R L_\Delta \sqrt{M}}{U}
\;\le\;
3 + \frac{2 R L_\Delta \sqrt{M}}{U} .
\]
If $\rho^{\star} = R$, then $R \le U / ( L_\Delta \sqrt{M} )$,
so
\[
2 L_\Delta \rho^{\star}
\;=\;
2 L_\Delta R
\;\le\;
\frac{2U}{\sqrt{M}} ,
\qquad
1 + \frac{2R}{\rho^{\star}}
\;=\;
3
\;\le\;
3 + \frac{2 R L_\Delta \sqrt{M}}{U} .
\]
In both regimes the right side of the display is at most
$r_M(\alpha)$ of Eq.~\ref{eq:dyn-rM}.
\end{proof}

\noindent
Two remarks on the hypothesis (\ref{eq:dyn-e1}). First, the
approximation target in (\ref{eq:dyn-e1}) is data-defined, not
chosen: the ideal gaps are computed at the true supports $J_m$,
and by Theorem~\ref{thm:ident}(a) and
Proposition~\ref{prop:welldef} the true supports are the unique
minimal selections of the noiseless samples. Second,
(\ref{eq:dyn-e1}) is an expressivity assumption on the gate
network, stated at the observed samples because
Lemma~\ref{lem:dyn-gate}(ii) applies the assumption at exactly
those samples. Under this assumption, part (i) turns the
existence of margins from a postulate into a consequence of
identifiability plus expressivity, and fixes the scale at which
margins can be demanded, namely $g_{\mathrm{gap}}/2$.

\noindent
\emph{What the aggregation grounds, and what it does not.} Under
(\ref{eq:dyn-ps}) the per-sample margins ground the dataset-mean
gate with an explicit finite-sample failure probability: the
mean-gate idealization of $\hat J_t$ is discharged rather than
assumed. What the aggregation does not ground is equally
explicit. As $\gamma \to 0$ the band excluded by
(\ref{eq:dyn-ps}) tends to
$\bigl( g_{\mathrm{gap}} / ( g_{\mathrm{gap}} + 4U ),\;
4U / ( g_{\mathrm{gap}} + 4U ) \bigr)$, a wide band when
$U \gg g_{\mathrm{gap}}$, and a component whose prevalence falls
inside the band genuinely has a mean gate of ambiguous sign: a
real component active in few samples mixes a few positive
per-sample margins with many negative ones. For such a component
the mean gate is the wrong object, not a gap in the analysis,
and the implemented gate is the better object: the implemented
gate thresholds the per-sample selection probability at
$\tau_{\mathrm{thres}} = 0.9999994$, which at
$\tau_{\mathrm{eval}} = 0.01$ is a per-sample energy-gap floor of
about $0.14$ (Appendix~\ref{app:gate}), and per-sample
selection is the semantics that Lemma~\ref{lem:dyn-gate}(i)
grounds without any prevalence condition. Finally, the same
sandwich taken in expectation reads
$m_j( \theta^{\dagger} ) \ge \gamma + \kappa ( a + 2U )$ for
$j \in S^{\star}$ under (\ref{eq:dyn-ps}), which is the scale
demanded of $\gamma^{\star}$ in (E7): the expectation sandwich
calibrates the scale of $\gamma^{\star}$. The calibration is not
a certificate for $\gamma^{\star}$, because it treats the
per-sample margins as holding for $\mathcal{D}$-almost every
sample, while Lemma~\ref{lem:dyn-gate}(i) provides them only on
a finite-sample noise event.

\medskip
\noindent
\emph{Scope of the constants (the $D \log$ term).} The parameter
count $D$ is a property of the solver's gate network, not of the
inference problem: the pool, the supports, and the
concentrations live in $\R^d$, and no statement of Theorems~1--3
involves $D$. $D$ enters in exactly one place, the uniform
concentration over the parameter ball in part (iii), where the
$D \log(\cdot)$ factor counts the net that covers the full
parameter ball, a deliberately coarse account of the search
space rather than a measured property of the gate network. For
the implemented architecture $D$ is the full parameter count, of
order $10^4$ to $10^6$, while the experiments use $M$ between
$10^3$ and $10^4$. At these scales the conditions
(\ref{eq:dyn-c1})--(\ref{eq:dyn-c2}) fail numerically:
$r_M(\alpha)$ exceeds $U/2$, while any usable margin has
$\gamma \le 2U$ because each $\hat\Delta_j$ is an average of
$[-2U, 2U]$-valued terms, so the requirement
$r_M(\alpha) \le \gamma/4 \le U/2$ cannot hold. The failure is
confined accordingly. The population rider of part (i) and the
support conversions of parts (iii)--(iv) are, at experimental
scale, conditional on hypotheses that fail numerically; the sure
window of part (i) and the objective-gap bounds of parts
(ii)--(iv) contain no $r_M$ and stand as stated. The bound is
read for the mechanism and the direction of the $M$-dependence:
larger $M$ tightens every margin uniformly over $\Theta$, the
mechanism behind the faster stabilization observed at $8N$
versus $4N$ samples (\S\ref{sec:exp}); the bound does not give
certified constants at experimental scale.

The second lemma proves part (i) of the proposition: after the
observed anchor, Lipschitz continuity of the energies and the
bounded optimizer step alone confine every margin for an
explicit number of steps. The window is sure, needing neither
probability nor sample size; the concentration event enters only
to upgrade empirical margins to population ones.

\begin{lemma}[The anchor window]
\label{lem:dyn-window}
Let (E2)--(E4) hold with anchor level $\gamma$ at time $t_0$, and
recall the stability horizon
\begin{equation}
T(\gamma) \;=\; \sup\Bigl\{ T \in \mathbb{N} \;:\;
L_\Delta\, G \sum_{s = t_0}^{t_0 + T - 1} \eta_s
\;\le\; \frac{\gamma}{4} \Bigr\}
\qquad \Bigl( \text{for constant } \eta:\ \
T(\gamma) = \Bigl\lfloor \frac{\gamma}{4 L_\Delta G \eta}
\Bigr\rfloor \Bigr).
\label{eq:dyn-Tgamma}
\end{equation}
Then:
\emph{(i) (Margins persist.)} Deterministically, for every realization of
the sample and of the training noise,
\[
\hat m_j( \theta_t ) \ge \frac{3\gamma}{4}
\qquad
\text{for all } j \in [N],\;
t \in [\, t_0,\; t_0 + T(\gamma) \,] .
\]
\emph{(ii) (Constant support.)} On the same realizations,
$\hat J_t = J_{\sharp}$ for all
$t \in [\, t_0,\; t_0 + T(\gamma) \,]$.
\emph{(iii) (Population rider.)} If in addition (E1) and
(\ref{eq:dyn-c1}) hold, then on $E_{\mathrm{gap}}$ also
$m_j( \theta_t ) \ge \gamma/2$ for all such $j$ and $t$.
\end{lemma}

\begin{proof}
\emph{Step 1 (path length).} By the triangle inequality and
(E3),
\[
\| \theta_t - \theta_{t_0} \|
\;\le\;
\sum_{s = t_0}^{t-1} \| \theta_{s+1} - \theta_s \|
\;\le\;
G \sum_{s = t_0}^{t-1} \eta_s .
\]

\emph{Step 2 (Lipschitz).} By (E2), $\Delta_j$ and
$\hat\Delta_j$ are $L_\Delta$-Lipschitz on $\Theta$ (the
empirical statement holds surely), so Step 1 gives, for any
$t_0 \le t$,
\[
\max_{j \in [N]}\;
\bigl| \Delta_j( \theta_t ) - \Delta_j( \theta_{t_0} ) \bigr|
\;\le\;
L_\Delta\, G \sum_{s = t_0}^{t-1} \eta_s ,
\]
and the same bound holds for $\hat\Delta_j$, for every
realization of the sample.

\emph{Step 3 (sure empirical route).} By (E4),
$\hat m_j( \theta_{t_0} ) \ge \gamma$ for every $j$.
The bound of Step 2 applies to the empirical gaps for every
sample realization, so for $t \in [t_0,\, t_0 + T(\gamma)]$, by
(\ref{eq:dyn-Tgamma}),
\[
\hat m_j( \theta_t )
\;\ge\;
\hat m_j( \theta_{t_0} ) - L_\Delta G
\sum_{s = t_0}^{t-1} \eta_s
\;\ge\;
\gamma - \frac{\gamma}{4}
\;=\; \frac{3\gamma}{4}
\;>\; 0 .
\]
This is part (i).

\emph{Step 4 (support equality).} Positivity of every signed
empirical margin is exactly the statement that the sign pattern
of $\hat\Delta_j( \theta_t )$ agrees with $J_{\sharp}$ (the
gap-sign convention of the Notation paragraph), that is,
$\hat J_t = J_{\sharp}$, which is
part (ii).

\emph{Step 5 (population rider).} On $E_{\mathrm{gap}}$, by
Lemma~\ref{lem:dyn-gate}(iii) and (\ref{eq:dyn-c1}), for every
$\theta \in \Theta$ simultaneously,
\[
\bigl| \hat\Delta_j(\theta) - \Delta_j(\theta) \bigr|
\;\le\;
r_M(\alpha)
\;\le\;
\frac{\gamma}{4} ;
\]
in particular the bound holds at
every iterate of the trajectory, with no independence between
$\theta_t$ and the data required. Hence
\[
m_j( \theta_t )
\;\ge\;
\hat m_j( \theta_t ) - \frac{\gamma}{4}
\;\ge\;
\frac{3\gamma}{4} - \frac{\gamma}{4}
\;=\; \frac{\gamma}{2} .
\qedhere
\]
\end{proof}

\noindent
The rest of the proposition takes place inside this block: the
descent of part (ii) runs on it, and the permanence of parts
(iii)--(iv) starts from the state it certifies. The two routes
of the lemma serve separate needs of that program. The sure
route involves no probability: $T(\gamma)$ is computed from the
Lipschitz constant $L_\Delta$, the displacement bound $G$, and
the step sizes alone, so an observed anchor certifies a
support-stable window whose length is set by run parameters, at
any sample size. The population rider upgrades the empirical
margins to margins for the latent law on $E_{\mathrm{gap}}$; the
sample-size condition (\ref{eq:dyn-c1}) enters the proposition
only there. The two hypotheses respond to separate controls of a
run: a larger $M$ reduces $r_M(\alpha)$, easing
(\ref{eq:dyn-c1}); a smaller step or a smaller $L_\Delta G$
lengthens $T(\gamma)$. Both controls are observable in a run.

On the certified block the evaluation support is constant, so
the objective the checkpoint sees reduces to $f$, the objective
with the gate frozen at $J_{\sharp}$; the third lemma proves
that $f$ contracts linearly, in expectation, to an explicit
noise floor, the standard Polyak--{\L}ojasiewicz descent
\citep[Theorems~1 and~4]{karimi2016linear} carried out with a bias term, the
recognized cost of training through a relaxed gate at
$\tau > 0$. It then supplies the mechanism of self-sustenance:
quadratic growth converts a small objective gap into parameter
proximity to the minimizer set, the minimizer margins of (E7)
and Lipschitz continuity convert proximity into margin
positivity, so a small gap forces the gate to re-select
$J_{\sharp}$ with no separate assumption, and Markov's
inequality bounds the risk at each fixed time. The per-time
bound cannot be summed over a long horizon, a union over $T$
time points costing $\approx T p$; treating the whole horizon
as a single event, by compensating the objective gap into a
supermartingale and applying a maximal inequality, makes the
failure probability grow only at the noise-injection rate per
step.

\begin{lemma}[Descent, basin, and permanence]
\label{lem:dyn-inside}
\emph{(i) (One-step contraction with bias and noise.)}
Let (E5) and (E6) hold, let $\eta_t \le \tfrac{1}{2L}$, and
suppose the segment $[ \theta_t, \theta_{t+1} ]$ lies in
$\mathcal{U}$ almost surely given $\mathcal{F}_t$. Then
\begin{equation}
\mathbb{E}\bigl[ f( \theta_{t+1} ) - f^{\star} \,\big|\, \mathcal{F}_t
\bigr]
\;\le\;
\Bigl( 1 - \frac{\mu \eta_t}{2} \Bigr)
\bigl( f( \theta_t ) - f^{\star} \bigr)
\;+\; \eta_t \beta^2
\;+\; \frac{L}{2} \eta_t^2 \bigl( \nu^2 + 2 \beta^2 \bigr) .
\label{eq:dyn-onestep}
\end{equation}
With a constant step $\eta \le \tfrac{1}{2L}$, unrolling from
$t_0$ gives, for every $k \ge 0$,
\begin{equation}
\mathbb{E}\bigl[ f( \theta_{t_0 + k} ) - f^{\star} \bigr]
\;\le\;
\Bigl( 1 - \frac{\mu \eta}{2} \Bigr)^{k}
\bigl( f( \theta_{t_0} ) - f^{\star} \bigr)
\;+\; \Phi(\eta) ,
\qquad
\Phi(\eta) = \frac{2 \beta^2}{\mu}
+ \frac{L \eta ( \nu^2 + 2 \beta^2 )}{\mu} .
\label{eq:dyn-unrolled}
\end{equation}

\emph{(ii) (Margin from objective gap.)}
Let (E2), (E5), and (E7) hold, and recall
$\varepsilon^{\star} = \mu ( \gamma^{\star} )^2 /
( 8 L_\Delta^2 )$. Then every $\theta \in \mathcal{U}$ with
$f(\theta) - f^{\star} \le \varepsilon^{\star}$ satisfies
\[
m_j( \theta ) \;\ge\; \frac{\gamma^{\star}}{2}
\qquad \text{for all } j \in [N] .
\]
If moreover (E1) and (\ref{eq:dyn-c2}) hold, then on
$E_{\mathrm{gap}}$ additionally
$\hat m_j( \theta ) \ge \gamma^{\star}/4 > 0$, so the evaluation
gate at $\theta$ selects $J_{\sharp}$.

\emph{(iii) (Per-time-point permanence.)}
In the setting of parts (i) and (ii), with constant step $\eta \le \tfrac{1}{2L}$,
fix $p \in (0, 1)$ and suppose (\ref{eq:dyn-c3}),
$\Phi(\eta) \le p\, \varepsilon^{\star}$. Let $t_1 \ge t_0$ be a
time with
$\mathbb{E}[ f( \theta_{t_1} ) - f^{\star} ] \le
\varepsilon^{\star}/2$. Then for every fixed $t \ge t_1$, with
iterates in $\mathcal{U}$,
\[
\Pr\bigl( f( \theta_t ) - f^{\star} > \varepsilon^{\star} \bigr)
\;\le\;
\frac12 \Bigl( 1 - \frac{\mu\eta}{2} \Bigr)^{t - t_1} + p ,
\]
and on every sample realization in $E_{\mathrm{gap}}$, under
(\ref{eq:dyn-c2}), the same bound holds for
$\Pr( \hat J_t \ne J_{\sharp} )$. The probabilities are over the
training randomness only; the data is
$\mathcal{F}_{t_1}$-measurable, so the bound and the entry
condition are both read per data realization.

\emph{(iv) (Uniform-in-time permanence.)}
In the setting of parts (i) and (ii), with constant step $\eta \le \tfrac{1}{2L}$,
write $z_t = f( \theta_t ) - f^{\star} \ge 0$ and
\[
c_\eta \;=\; \eta \beta^2
+ \frac{L}{2} \eta^2 \bigl( \nu^2 + 2 \beta^2 \bigr)
\;=\; \frac{\mu\eta}{2}\, \Phi(\eta) ,
\]
the per-step injection of (\ref{eq:dyn-onestep}). Then for every
time $t_1 \ge t_0$, every horizon $T \in \mathbb{N}$, and every
level $\varepsilon > 0$, with iterates in $\mathcal{U}$,
\begin{equation}
\Pr\Bigl( \exists\, t \in [\, t_1,\; t_1 + T \,] :\;
z_t \ge \varepsilon \Bigr)
\;\le\;
\frac{ \mathbb{E}[ z_{t_1} ] + T c_\eta }{ \varepsilon }
.
\label{eq:dyn-uperm}
\end{equation}
Under (\ref{eq:dyn-c3}) (applied below at the basin level
$\varepsilon^{\star}$) the right side is at most
$\mathbb{E}[ z_{t_1} ] / \varepsilon^{\star} +
p\, \tfrac{\mu\eta}{2}\, T$. On the complementary event every
$t$ in the horizon has gap below $\varepsilon^{\star}$, so on
every sample realization in $E_{\mathrm{gap}}$, under
(\ref{eq:dyn-c2}), part (ii) yields
\[
\hat J_t = J_{\sharp}
\qquad \text{and} \qquad
\hat m_j( \theta_t ) \ge \frac{\gamma^{\star}}{4}
\qquad
\text{for all } t \in [\, t_1,\; t_1 + T \,]
\text{ simultaneously.}
\]
(Expectations conditional on $\mathcal{F}_{t_0}$; conditioning on
$\mathcal{F}_{t_1}$ instead replaces $\mathbb{E}[ z_{t_1} ]$ by the
observed gap $z_{t_1}$.)
\end{lemma}

\begin{proof}
\emph{Part (i).}

\emph{Step 1 (smoothness).} By the $L$-Lipschitz gradient of
(E5) and the quadratic upper bound it implies along the segment
\citep[Lem.~1.2.3]{nesterov2018lectures},
\[
f( \theta_{t+1} )
\;\le\;
f( \theta_t )
- \eta_t \bigl\langle \nabla f( \theta_t ),\, g_t \bigr\rangle
+ \frac{L}{2} \eta_t^2 \| g_t \|^2 .
\]

\emph{Step 2 (inner product with bias).} Taking
$\mathbb{E}[\,\cdot \mid \mathcal{F}_t]$ and writing
$\nabla f = \nabla f( \theta_t )$, by (E6) and Cauchy--Schwarz,
\[
\mathbb{E}\bigl[ \langle \nabla f, g_t \rangle \,\big|\, \mathcal{F}_t
\bigr]
\;=\;
\| \nabla f \|^2 + \langle \nabla f, b_t \rangle
\;\ge\;
\| \nabla f \|^2 - \beta \| \nabla f \| ,
\]
and since $0 \le \bigl( \tfrac12 \| \nabla f \| - \beta \bigr)^2$
expands to the bound on the left below,
\[
\beta \| \nabla f \|
\;\le\;
\frac14 \| \nabla f \|^2 + \beta^2 ,
\qquad \text{so} \qquad
\mathbb{E}\bigl[ \langle \nabla f, g_t \rangle \,\big|\,
\mathcal{F}_t \bigr]
\;\ge\;
\frac34 \| \nabla f \|^2 - \beta^2 .
\]

\emph{Step 3 (quadratic term).} By the variance bound of (E6)
and $\| a + b \|^2 \le 2 \| a \|^2 + 2 \| b \|^2$,
\[
\mathbb{E}\bigl[ \| g_t \|^2 \,\big|\, \mathcal{F}_t \bigr]
\;\le\;
\nu^2 + \| \nabla f + b_t \|^2
\;\le\;
\nu^2 + 2 \| \nabla f \|^2 + 2 \beta^2 .
\]

\emph{Step 4 (collect).} Substituting Steps 2 and 3 into Step 1,
\[
\mathbb{E}\bigl[ f( \theta_{t+1} ) \,\big|\, \mathcal{F}_t \bigr]
\;\le\;
f( \theta_t )
- \eta_t \Bigl( \frac34 - L \eta_t \Bigr) \| \nabla f \|^2
+ \eta_t \beta^2
+ \frac{L}{2} \eta_t^2 \bigl( \nu^2 + 2 \beta^2 \bigr) .
\]

\emph{Step 5 (Polyak--{\L}ojasiewicz).} With
$\eta_t \le \tfrac{1}{2L}$,
\[
\frac34 - L \eta_t
\;\ge\;
\frac34 - \frac12
\;=\;
\frac14 ,
\]
and the inequality of (E5) gives
\[
- \frac{\eta_t}{4} \| \nabla f \|^2
\;\le\;
- \frac{\mu \eta_t}{2} \bigl( f( \theta_t ) - f^{\star} \bigr)
\]
\citep[Theorem~4]{karimi2016linear}. Subtracting $f^{\star}$ from both
sides of Step 4 yields (\ref{eq:dyn-onestep}).

\emph{Step 6 (unroll).} Take total expectations in
(\ref{eq:dyn-onestep}) and iterate. The constants satisfy
$\mu \le L$: $L$-smoothness gives
$\| \nabla f(\theta) \|^2 \le
2L \bigl( f(\theta) - f^{\star} \bigr)$ (the descent lemma
evaluated at a step of length $1/L$), the
Polyak--{\L}ojasiewicz inequality gives
$\| \nabla f(\theta) \|^2 \ge
2\mu \bigl( f(\theta) - f^{\star} \bigr)$, and dividing the
two bounds at any $\theta$ with $f(\theta) > f^{\star}$ yields
$\mu \le L$ (if no such $\theta$ exists, every bound below is
trivial). So with $\eta \le \tfrac{1}{2L}$,
\[
\frac{\mu\eta}{2}
\;\le\;
\frac{\mu}{4L}
\;\le\;
\frac14 ,
\qquad \text{hence} \qquad
1 - \frac{\mu\eta}{2} \in [0, 1) .
\]
The accumulated injections are bounded by the geometric series
\[
\sum_{i \ge 0} \Bigl( 1 - \frac{\mu\eta}{2} \Bigr)^i
\Bigl( \eta \beta^2 + \frac{L}{2} \eta^2 ( \nu^2 + 2\beta^2 )
\Bigr)
\;\le\;
\frac{2}{\mu\eta}
\Bigl( \eta \beta^2 + \frac{L}{2} \eta^2 ( \nu^2 + 2\beta^2 )
\Bigr)
\;=\; \Phi(\eta) ,
\]
which is (\ref{eq:dyn-unrolled}).

\emph{Part (ii).}

\emph{Step 7 (proximity).} By the quadratic-growth bound of
(E5),
\[
\dist\bigl( \theta, \mathcal{X}^{\star} \bigr)
\;\le\;
\sqrt{ \frac{2 ( f(\theta) - f^{\star} )}{\mu} }
\;\le\;
\sqrt{ \frac{2 \varepsilon^{\star}}{\mu} }
\;=\;
\frac{\gamma^{\star}}{2 L_\Delta} .
\]

\emph{Step 8 (margin transfer).} Fix $\epsilon > 0$ and pick
$\bar\theta \in \mathcal{X}^{\star}$ with
$\| \theta - \bar\theta \| \le
\dist( \theta, \mathcal{X}^{\star} ) + \epsilon$. By (E7) and
the Lipschitz property of (E2),
\[
m_j( \theta )
\;\ge\;
m_j( \bar\theta ) - L_\Delta \| \theta - \bar\theta \|
\;\ge\;
\gamma^{\star} - L_\Delta
\Bigl( \frac{\gamma^{\star}}{2 L_\Delta} + \epsilon \Bigr)
\;=\;
\frac{\gamma^{\star}}{2} - L_\Delta \epsilon ,
\]
and letting $\epsilon \downarrow 0$ gives
$m_j( \theta ) \ge \gamma^{\star}/2$.

\emph{Step 9 (empirical rider).} On $E_{\mathrm{gap}}$ with
(\ref{eq:dyn-c2}),
\[
\hat m_j( \theta )
\;\ge\;
m_j( \theta ) - r_M(\alpha)
\;\ge\;
\frac{\gamma^{\star}}{2} - \frac{\gamma^{\star}}{4}
\;=\;
\frac{\gamma^{\star}}{4}
\;>\; 0 ,
\]
and positivity of every signed empirical margin is the gate
identity (the gap-sign convention of the Notation paragraph).

\emph{Part (iii).}

\emph{Step 10 (unroll from $t_1$).} By
(\ref{eq:dyn-unrolled}) started at $t_1$ and the entry
condition,
\[
\mathbb{E}\bigl[ f( \theta_t ) - f^{\star} \bigr]
\;\le\;
\Bigl( 1 - \frac{\mu\eta}{2} \Bigr)^{t - t_1}
\frac{\varepsilon^{\star}}{2}
+ \Phi(\eta) .
\]

\emph{Step 11 (Markov).} By Markov's inequality at level
$\varepsilon^{\star}$ and (\ref{eq:dyn-c3}),
\[
\Pr\bigl( f( \theta_t ) - f^{\star} > \varepsilon^{\star} \bigr)
\;\le\;
\frac{ ( 1 - \tfrac{\mu\eta}{2} )^{t - t_1}\,
\tfrac{\varepsilon^{\star}}{2}
+ p\, \varepsilon^{\star} }{ \varepsilon^{\star} }
\;=\;
\frac12 \Bigl( 1 - \frac{\mu\eta}{2} \Bigr)^{t - t_1} + p .
\]

\emph{Step 12 (conversion).} On the complementary event
$\{ f( \theta_t ) - f^{\star} \le \varepsilon^{\star} \}$,
intersected with $E_{\mathrm{gap}}$,
part (ii) gives $\hat J_t = J_{\sharp}$; the
support-failure event is therefore contained in the
gap-exceedance event, on $E_{\mathrm{gap}}$.

\emph{Part (iv).}

\emph{Step 13 (drift bound).} Since $z_t \ge 0$ and
$1 - \tfrac{\mu\eta}{2} \le 1$, the one-step recursion
(\ref{eq:dyn-onestep}) implies
\[
\mathbb{E}\bigl[ z_{t+1} \,\big|\, \mathcal{F}_t \bigr]
\;\le\;
\Bigl( 1 - \frac{\mu\eta}{2} \Bigr) z_t + c_\eta
\;\le\;
z_t + c_\eta .
\]

\emph{Step 14 (compensated supermartingale).} Define, for
$t \in [\, t_1,\; t_1 + T \,]$,
\[
Y_t \;=\; z_t + \bigl( T - ( t - t_1 ) \bigr)\, c_\eta ,
\]
the gap plus a deterministic compensator that absorbs the
remaining injections in advance. By Step 13,
\[
\mathbb{E}\bigl[ Y_{t+1} \,\big|\, \mathcal{F}_t \bigr]
\;=\;
\mathbb{E}\bigl[ z_{t+1} \,\big|\, \mathcal{F}_t \bigr]
+ \bigl( T - ( t + 1 - t_1 ) \bigr) c_\eta
\;\le\;
z_t + c_\eta
+ \bigl( T - ( t - t_1 ) \bigr) c_\eta - c_\eta
\;=\; Y_t ,
\]
so $( Y_t )$ is a supermartingale on the horizon, and
$Y_t \ge z_t \ge 0$ there.

\emph{Step 15 (maximal inequality).} Let
$\tau = \min\{ t \ge t_1 : Y_t \ge \varepsilon^{\star} \}$, with
$\tau = \infty$ if no such $t$ exists in the horizon. For
$t_1 \le n < t_1 + T$,
\[
Y_{\tau \wedge (n+1)} - Y_{\tau \wedge n}
\;=\;
\mathbf{1}[\, \tau > n \,]
\bigl( Y_{n+1} - Y_n \bigr) ,
\]
and since $\{ \tau > n \} \in \mathcal{F}_n$, taking
$\mathbb{E}[\,\cdot \mid \mathcal{F}_n]$ and using Step 14,
\[
\mathbb{E}\bigl[ Y_{\tau \wedge (n+1)} \,\big|\, \mathcal{F}_n \bigr]
- Y_{\tau \wedge n}
\;=\;
\mathbf{1}[\, \tau > n \,]
\Bigl( \mathbb{E}\bigl[ Y_{n+1} \,\big|\, \mathcal{F}_n \bigr] - Y_n
\Bigr)
\;\le\; 0 .
\]
Iterating over $n$ and taking total expectations,
\[
\mathbb{E}\bigl[ Y_{\tau \wedge ( t_1 + T )} \bigr]
\;\le\;
\mathbb{E}\bigl[ Y_{t_1} \bigr] .
\]
On $\{ \tau \le t_1 + T \}$,
$Y_{\tau \wedge ( t_1 + T )} = Y_{\tau} \ge
\varepsilon^{\star}$, and $Y \ge 0$ everywhere on the horizon,
so
\[
\varepsilon^{\star} \,
\Pr\bigl( \tau \le t_1 + T \bigr)
\;\le\;
\mathbb{E}\bigl[ Y_{\tau \wedge ( t_1 + T )} \bigr]
\;\le\;
\mathbb{E}\bigl[ Y_{t_1} \bigr] ,
\]
the maximal inequality of \citet{ville1939etude} for nonnegative
supermartingales, bounding the running maximum by the initial
expectation (in its modern time-uniform form,
\citealp[Lemma~1]{howard2020timeuniform}). This is the step that replaces
the per-time-point Markov bound of part (iii): the
submartingale inequality of Doob bounds by the terminal value and
cannot deliver uniformity from an initial condition.

\emph{Step 16 (assembly).} On the horizon $Y_t \ge z_t$, so the
$z$-exceedance event is contained in
$\{ \tau \le t_1 + T \}$, and
\[
\mathbb{E}\bigl[ Y_{t_1} \bigr]
\;=\;
\mathbb{E}\bigl[ z_{t_1} \bigr] + T c_\eta ,
\]
which is (\ref{eq:dyn-uperm}). Under (\ref{eq:dyn-c3}),
\[
\frac{T c_\eta}{\varepsilon^{\star}}
\;=\;
\frac{\mu\eta}{2}\, \frac{T\, \Phi(\eta)}{\varepsilon^{\star}}
\;\le\;
p\, \frac{\mu\eta}{2}\, T .
\]
The support statement follows by
applying part (ii) at every $t$ of the
complementary event, on $E_{\mathrm{gap}}$.
\end{proof}

\noindent
With the three lemmas in place, the proposition follows by
composition, and each step of the proof completes one stage of
the solver's stabilization. Step 1 sets the conditioning: every
statement is read from the run's log at the anchor. Step 2
gives part (i): by Lemma~\ref{lem:dyn-window}, bounded
optimizer steps cannot move any margin across zero for
$T(\gamma)$ steps, so the selection holds fixed. Step 3 gives
part (ii): with the selection frozen, the objective the run
descends is a single smooth function, and
Lemma~\ref{lem:dyn-inside}(i) contracts it to its noise floor,
with (\ref{eq:dyn-c4}) guaranteeing that the contraction
finishes inside the window. Steps 4 and 5 give parts (iii) and
(iv): the contracted gap bounds the flip risk at each fixed
time through Markov's inequality,
Lemma~\ref{lem:dyn-inside}(ii) converts a small gap into
re-selection of $J_{\sharp}$, and
Lemma~\ref{lem:dyn-inside}(iv), applied at the entry time,
treats the whole horizon as a single event. Every population
statement holds on the concentration event $E_{\mathrm{gap}}$
of Lemma~\ref{lem:dyn-gate}.

The probabilistic language of the proof describes an ordinary
training run of the EB-gMCR solver of \S\ref{sec:solver}; the
run in every statement below is that solver running. The symbol
$\mathcal{F}_t$, defined in the notation table, stands for the
training log of the run up to step $t$: the dataset together
with every random choice the solver has made through step $t$.
A quantity is $\mathcal{F}_{t_0}$-measurable when the log up to
the anchor time determines it; the iterate $\theta_{t_0}$ and
the anchored support $J_{\sharp}$ are the two examples the
proof uses. Conditioning on $\mathcal{F}_{t_0}$ fixes the
observed history and lets only the randomness after the anchor
vary, so every probability below is a statement about how the
solver's run continues, given what its log already shows at the
anchor.

\begin{proof}[Proof of Proposition~2]
\emph{Step 1 (conditioning).} All expectations and probabilities
over the training randomness are conditional on $\mathcal{F}_{t_0}$. The
anchor time is a stopping time, but the anchor event and
$f( \theta_{t_0} ) - f^{\star}$ are $\mathcal{F}_{t_0}$-measurable;
decomposing over the events $\{ t_0 = s \} \in \mathcal{F}_s$ and
iterating on each, the one-step recursion of
Lemma~\ref{lem:dyn-inside}(i) applies conditionally on
$\mathcal{F}_{t_0}$, and conditions
(\ref{eq:dyn-c3})--(\ref{eq:dyn-c4}) are
$\mathcal{F}_{t_0}$-measurable. The conditioning is suppressed in the
notation below. The sample is $\mathcal{F}_{t_0}$-measurable as well, so
every statement made on $E_{\mathrm{gap}}$ is a statement per
sample realization.

\emph{Step 2 (part (i)).} This is
Lemma~\ref{lem:dyn-window} verbatim: the block support and the
empirical margins hold surely, and the population margins follow
on $E_{\mathrm{gap}}$ under (\ref{eq:dyn-c1}).

\emph{Step 3 (part (ii)).} On the certified block the evaluation
support is constant by part (i), the usage penalty is constant
once the gate is frozen, and Lemma~\ref{lem:dyn-inside}(i) applies
to $f$; unrolling (\ref{eq:dyn-unrolled}) for
$0 \le k \le T(\gamma)$ gives the display, and
(\ref{eq:dyn-c4}) guarantees $T_{\mathrm{ctr}} \le T(\gamma)$,
so the contraction completes within the block. By the definition
of $T_{\mathrm{ctr}}$ and $\log( 1 - x ) \le -x$ for
$x \in [0, 1)$,
\[
\Bigl( 1 - \frac{\mu\eta}{2} \Bigr)^{T_{\mathrm{ctr}}}
\bigl( f( \theta_{t_0} ) - f^{\star} \bigr)
\;\le\;
\exp\Bigl( - \frac{\mu\eta}{2}\, T_{\mathrm{ctr}} \Bigr)
\bigl( f( \theta_{t_0} ) - f^{\star} \bigr)
\;\le\;
\frac{\varepsilon^{\star}}{2} ,
\]
so, by (\ref{eq:dyn-c3}) and $p \le \tfrac12$,
\[
\mathbb{E}\bigl[ f( \theta_{t_0 + T_{\mathrm{ctr}}} )
- f^{\star} \bigr]
\;\le\;
\frac{\varepsilon^{\star}}{2} + \Phi(\eta)
\;\le\;
\Bigl( \frac12 + p \Bigr) \varepsilon^{\star}
\;\le\;
\varepsilon^{\star} .
\]
This is exactly the in-particular bound of part (ii).

\emph{Step 4 (part (iii)).} Lemma~\ref{lem:dyn-inside}(iii)
cannot be applied at $t_1 = t_0 + T_{\mathrm{ctr}}$, because
Step 3 delivers entry level $(\tfrac12 + p)\varepsilon^{\star}$
rather than the $\varepsilon^{\star}/2$ its hypothesis requires;
instead unroll (\ref{eq:dyn-unrolled}) once
from $t_0$, without restarting: for
$t \ge t_0 + T_{\mathrm{ctr}}$, by Step 3 and
(\ref{eq:dyn-c3}),
\[
\mathbb{E}\bigl[ f( \theta_t ) - f^{\star} \bigr]
\;\le\;
\Bigl( 1 - \frac{\mu\eta}{2} \Bigr)^{t - t_0}
\bigl( f( \theta_{t_0} ) - f^{\star} \bigr) + \Phi(\eta)
\;\le\;
\Bigl( 1 - \frac{\mu\eta}{2}
\Bigr)^{t - t_0 - T_{\mathrm{ctr}}}
\frac{\varepsilon^{\star}}{2}
+ p\, \varepsilon^{\star} ,
\]
and Markov's inequality at level $\varepsilon^{\star}$ gives the
gap bound of part (iii), per sample realization. On realizations
in $E_{\mathrm{gap}}$, under (\ref{eq:dyn-c2}),
Lemma~\ref{lem:dyn-inside}(ii) converts
$\{ f( \theta_t ) - f^{\star} \le \varepsilon^{\star} \}$ into
$\{ \hat J_t = J_{\sharp} \}$, which is the support form.

\emph{Step 5 (part (iv)).} Apply
Lemma~\ref{lem:dyn-inside}(iv) at $t_1 = t_0 + T_{\mathrm{ctr}}$. By
Step 3 the entry satisfies
$\mathbb{E}[ z_{t_1} ] \le ( \tfrac12 + p )\,
\varepsilon^{\star}$ (the floor does not contract away, whence
the $\tfrac12 + p$), and
$T c_\eta / \varepsilon^{\star} \le p\, \tfrac{\mu\eta}{2}\, T$
by (\ref{eq:dyn-c3}). The right side of (\ref{eq:dyn-uperm}) at
level $\varepsilon^{\star}$ is therefore at most
$\tfrac12 + p + p\, \tfrac{\mu\eta}{2}\, T$, which is exactly the
bound of part (iv); on every sample realization in
$E_{\mathrm{gap}}$, under (\ref{eq:dyn-c2}), the support form
over the same horizon is the conversion stated in
Lemma~\ref{lem:dyn-inside}(iv). With the deeper entry
$T_{\mathrm{ctr}}(\delta)$, by the argument of Step 3,
\[
\Bigl( 1 - \frac{\mu\eta}{2} \Bigr)^{T_{\mathrm{ctr}}(\delta)}
\bigl( f( \theta_{t_0} ) - f^{\star} \bigr)
\;\le\;
( \delta - p )\, \varepsilon^{\star} ,
\]
so, adding the floor and using (\ref{eq:dyn-c3}),
\[
\mathbb{E}\bigl[ z_{t_0 + T_{\mathrm{ctr}}(\delta)} \bigr]
\;\le\;
( \delta - p )\, \varepsilon^{\star} + \Phi(\eta)
\;\le\;
\delta\, \varepsilon^{\star} ,
\]
replacing $\tfrac12 + p$ by $\delta$.
\end{proof}

\medskip
\noindent
\emph{Concluding remarks.} Taking the proposition part by part:
\begin{itemize}[leftmargin=*]
\item \emph{Part (i), the window.} The conclusion is sure, at
any sample size, and it transfers to the implemented solver as
proved: Lemma~\ref{lem:dyn-window} consumes only the Lipschitz
continuity of (E2) and the bounded update of (E3), and the
normalized AdamW step satisfies (E3) by construction.
\item \emph{Part (ii), the contraction.} The rate is proved for
the SGD-form update of (E6); its AdamW counterpart is open. The
exploration noise of Eq.~\ref{eq:5} sits inside (E6), reaching the
parameters through the gradient with its effect carried by
$\beta$ and $\nu^2$, and annealing $\tau$ reduces the bias
$\beta$ and with it the floor $\Phi(\eta)$, matching the
implemented schedule. Smoothness of $f$ is idealized: the ReLU
layers of the concentration network break it at kink sets, a softplus reparametrization restores it at a larger $L$, and the window needs only Lipschitz continuity. The analysis also holds the usage weight and the temperature fixed after the anchor, while the implementation continues to update both; the window is unaffected, and the contraction rate inherits the open status it has under AdamW.
\item \emph{Parts (iii) and (iv), permanence.} After the
contraction, the per-time failure probability decays
geometrically to $p$, and the uniform bound accrues failure at
$p$ per $2/(\mu\eta)$ steps rather than $p$ per time point
queried, an improvement by the factor $2/(\mu\eta) \gg 1$; its
leading term reflects entering the basin only
$\varepsilon^{\star}/2$-deep in expectation. The linear growth
in $T$ is sharp: a Gaussian noise model consistent with (E6)
leaves the basin $\{ z_t \le \varepsilon^{\star} \}$ with
per-step probability bounded away from zero, so any uniform
bound must grow with the horizon. The support forms consume the
concentration conditions, whose status at implemented scale is
the scope remark after Lemma~\ref{lem:dyn-gate} (a norm-based
or energy-head covering would reduce the $D \log(\cdot)$
factor; none is shown sufficient here), and containment of
long-horizon segments in $\mathcal{U}$ remains assumed, with
projection onto a convex $\mathcal{U}$
\citep[Theorem~5]{karimi2016linear} or truncation of the
injected noise, which only reduces $\nu^2$, the enforcing
modifications.
\end{itemize}

\noindent
Two facts stand outside the parts. Nothing here proves that
training reaches the anchor: the selection dynamics under the
scheduled $\lambda_t$, the annealed $\tau_t$, and the Gumbel
sampling are not analyzed. And whether the anchored support is
the true one is a variational fact, not a dynamical one: inside
the SNR window of Theorem~\ref{thm:ident}(b) the true selection
is the strict optimum with gap $g_{\mathrm{gap}}$, and
Lemma~\ref{lem:dyn-gate} shows margins at the demanded scale
exist at some parameter, but whether the run's $J_{\sharp}$ is
that support remains an assumption about the run.

\medskip
\noindent
\emph{What this appendix establishes.} After an observed
anchor, every checkpoint taken within the certified window
records the same support (part (i)); by the end of the window
the frozen objective has contracted to its noise floor (part
(ii)); afterwards the support maintains itself, at a risk
quantified per time point (part (iii)) and per horizon (part
(iv)). No other result of the paper consumes this proposition.
The reading of the mechanism, its scope at implemented scale,
and its consequences for the MCR task are stated in
\S\ref{sec:solver-convergence}.

\section{A Worked Toy: The Model, Identifiability, and Reuse}
\label{app:toy}

This appendix instantiates the objects of \S\ref{sec:gmcr} on small
numbers and demonstrates, at toy scale, that each model condition
is necessary. The toy is deliberately \emph{separable} (components
also appear alone), a regime in which recovery is easy by
inspection; such data do not need the machinery of this paper,
whose target is the regime with no pure samples, a large pool, and
noise. The demonstrations below exhibit the two ways
identification fails when a condition is dropped, each ending with
an explicit statement of the boundary.

\paragraph{Setup: the model on numbers.}
Pool of two unit-norm component profiles in $\R^4$:
\[
s_1 = [0.6,\, 0.8,\, 0,\, 0], \qquad
s_2 = [0,\, 0,\, 0.8,\, 0.6],
\]
orthogonal, so $\sigma_{\min}$ of the pair is $1$. Fix the toy's
model constants: $c_{\min} = 1$, $c_{\max} = 10$, and spark margin
$s_{\min} = 1/2$, which the pool satisfies with room to spare.
Three noiseless samples with $k = 2$:
\begin{align*}
x_1 &= 5\, s_1 = [3,4,0,0], & J_1 &= \{1\},\\
x_2 &= 5\, s_2 = [0,0,4,3], & J_2 &= \{2\},\\
x_3 &= 5\, s_1 + 5\, s_2 = [3,4,4,3], & J_3 &= \{1,2\}.
\end{align*}
The concentration matrix is $[[5,0],[0,5],[5,5]]$, the occurrence
matrix $\delta$ has rows $[1,0], [0,1], [1,1]$, and the total
usage is $4$. Proposition~\ref{prop:welldef} applies to each sample: for
instance, $x_3$ has exactly one representation over the pool using
at most two components with concentrations at least $c_{\min}$.

\paragraph{Why occurrence diversity is necessary.}
Drop $x_1$ and learn the pool from $x_2, x_3$ alone; the inverse
problem then has infinitely many admissible solutions. A coarser
pool spends one fresh profile on $x_3$:
\[
s' = \frac{x_3}{\|x_3\|} \approx [0.42,\, 0.57,\, 0.57,\, 0.42],
\qquad
x_2 = 5\, s_2, \qquad x_3 = \sqrt{50}\, s' ,
\]
with total usage $2$ against the truth's $3$ on these samples, and
spark margin $\approx 0.54 \ge s_{\min}$. A whole family matches
the truth's usage: writing $a \in [c_{\min}, 5] = [1,5]$ for the
amount of $x_3$ attributed to $s_2$,
\[
r(a) = x_3 - a\, s_2, \qquad
\hat s_1(a) = \frac{r(a)}{\|r(a)\|}, \qquad
x_2 = 5\, s_2, \qquad
x_3 = \|r(a)\|\, \hat s_1(a) + a\, s_2 ,
\]
each member exact, with usage $3$, concentrations inside
$[c_{\min}, c_{\max}]$, and spark margin at least
$0.6 \ge s_{\min}$. The endpoint $a = 5$ is the truth; every other
$a$ is a wrong pool, for instance
\[
a = 2.5: \qquad
\hat s_1(2.5) \approx [0.54,\, 0.72,\, 0.36,\, 0.27], \qquad
x_3 = 5.59\, \hat s_1(2.5) + 2.5\, s_2 .
\]
Standard regularization selects wrong members with certainty:
\begin{align*}
\ell_1 &: \quad \sum |c| = 5 + \|r(a)\| + a ,
& &\text{minimum at } a = 1 ;\\
\ell_2 &: \quad \sum c^2 = 25 + \|r(a)\|^2 + a^2 = 75 - 10a + 2a^2 ,
& &\text{minimum at } a = 2.5 ,
\end{align*}
so an $\ell_1$ penalty picks the $a = 1$ endpoint, and an
$\ell_2$ penalty picks exactly the wrong pool displayed above. A
magnitude penalty encodes a preference among solutions; preference
is not information. From these two samples, no method can be
guaranteed to select the truth.

\paragraph{One further proportion restores identifiability.}
Restore $x_1$. Since
$\hat s_1(a) = (5 s_1 + (5-a) s_2)/\|r(a)\|$, the only
representation of $x_1$ over a family pool is
\[
x_1 = \|r(a)\|\, \hat s_1(a) - (5 - a)\, s_2 ,
\]
whose $s_2$-coefficient is negative for every $a \ne 5$:
inadmissible. A third profile cannot help: every sample lies in the
plane spanned by $s_1$ and $s_2$, a profile serving $x_1$ must lie
in that plane, and three unit profiles in one plane are linearly
dependent, violating Definition~\ref{def:spark}. With $x_1$
present, the truth is the unique admissible solution. One sample in
a new proportion eliminates infinitely many impostors, and more
samples in more proportions prune more; this is the occurrence
condition of \S\ref{sec:theory}, and it is why the approach gains
rather than loses from scale.

\paragraph{Why the spark condition is necessary.}
With all three samples, merging is cheaper than the truth:
\[
x_1 = 5\, s_1, \qquad x_2 = 5\, s_2, \qquad
x_3 = \sqrt{50}\, s_3 \quad \text{with } s_3 = x_3/\|x_3\| = s' ,
\]
at total usage $3 < 4$. But $s_3$ is a combination of $s_1$ and
$s_2$, so
\[
\sigma_{\min}\bigl( [\, s_1 \ \ s_2 \ \ s_3 \,] \bigr) = 0 ,
\]
and the merged pool violates Definition~\ref{def:spark}. Usage
minimality without the spark condition fails, and the spark
condition without usage minimality leaves the family above;
jointly, and in this example uniquely, they select the truth. The
general statement is the subject of \S\ref{sec:theory}.

\paragraph{Reuse.}
New samples from the same generative process, $x_4 = 10\, s_1$ and
$x_5 = 10\, s_2$, are decoded with the pool in hand by
Proposition~\ref{prop:welldef} applied per sample: supports
$\{1\}, \{2\}$, concentrations $10$. Matrix-factorization methods
must refactor the enlarged data matrix, and the recomputed factors
can drift; the process view reads the new samples off.

\paragraph{Physical coordinates.}
At the correct component count, and only there, the recovered
profiles match the physical constituents up to relabeling, and the
concentration entries are amounts of real constituents rather than
coordinates of a fit. Too few components merge constituents; too
many split them. Component-count recovery is therefore an
interpretability claim, not only an accuracy claim.

\newpage
\section{Gate Numerics and Energy Margins}
\label{app:gate}

This appendix records the numerical settings of the selection
gate of \S\ref{sec:solver-ebselect} and the margin they induce,
the quantity the convergence analysis of
Appendix~\ref{app:dyn} references.

\begin{figure}[t]
\centering
\includegraphics[width=\textwidth]{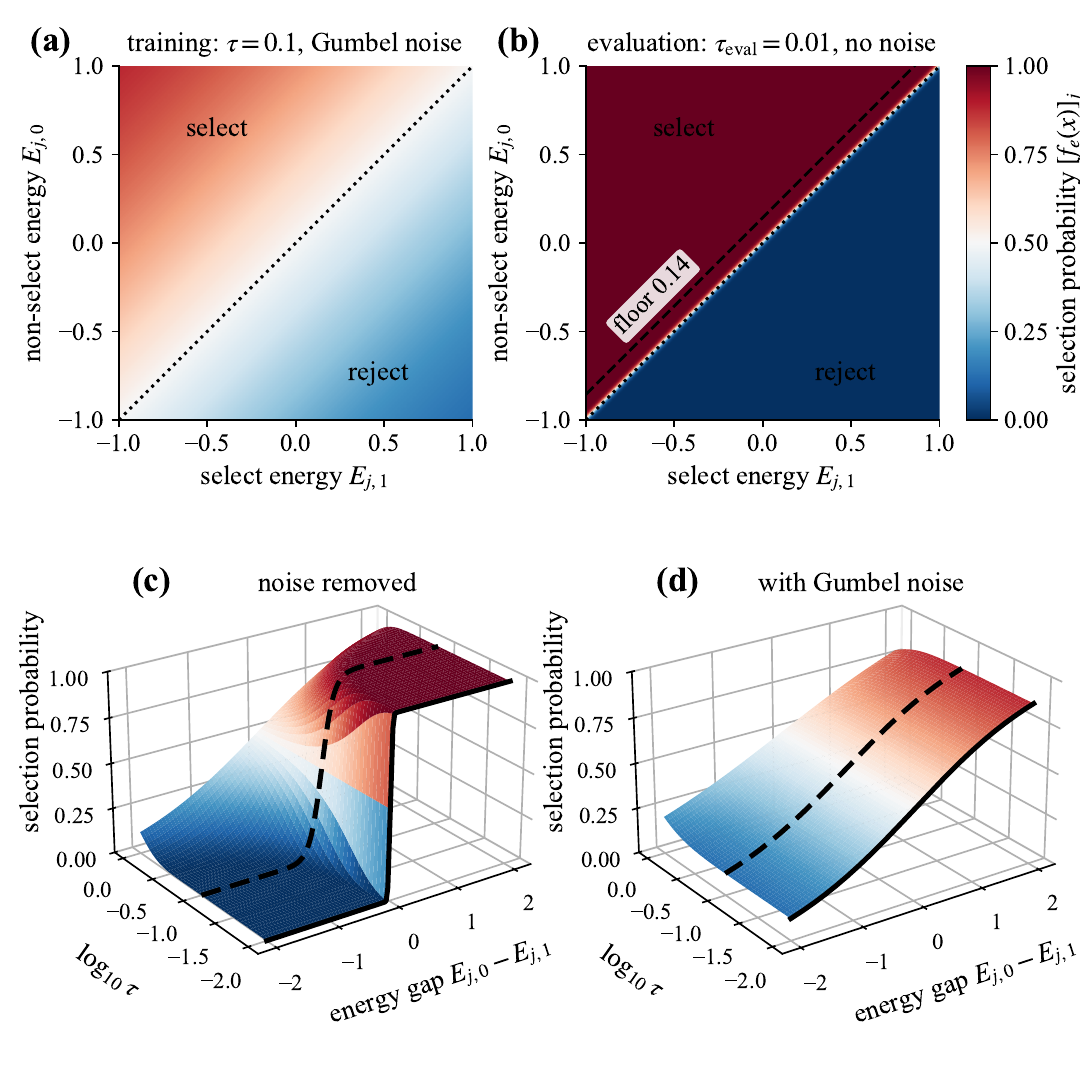}
\caption{Decision surface of the selection gate for one
component. (a) Training: expected selection probability at
$\tau = 0.1$ under Gumbel noise, over the select energy
$E_{j,1}$ and non-select energy $E_{j,0}$; the dotted diagonal
$E_{j,0} = E_{j,1}$ is the level set of probability $1/2$.
(b) Evaluation temperature
$\tau_{\mathrm{eval}} = 0.01$, noise-free surface;
the dotted diagonal is again the level set of probability
$1/2$, and the dashed line is the level set of the same
surface at probability $\tau_{\mathrm{thres}}$, the decision
boundary $E_{j,0} - E_{j,1} = 0.14$ of Eq.~\ref{eq:gate-floor}. (c, d) Selection
probability over the energy gap and $\log_{10} \tau$, without
(c) and with (d) Gumbel noise; the black curves are the cuts at
$\tau_{\mathrm{eval}} = 0.01$ (solid) and $\tau = 0.1$ (dashed).
Without noise the ramp sharpens as $\tau$ decreases; with noise
it saturates at unit width.}
\label{fig:gate}
\end{figure}

\paragraph{Threshold and floor.} At evaluation the gate runs at
$\tau_{\mathrm{eval}} = 0.01$, and component $j$ is selected
when $[f_e(x)]_j > \tau_{\mathrm{thres}} = 0.9999994$, a
five-sigma convention. By Eq.~\ref{eq:gumbel} at the noise-free point
$Z_{j,1} = Z_{j,0}$, the condition is an energy-gap floor:
\begin{equation}
E_{j,0}(x) - E_{j,1}(x)
\;\ge\;
\tau_{\mathrm{eval}} \operatorname{logit}( \tau_{\mathrm{thres}} )
\;\approx\; 0.14 .
\label{eq:gate-floor}
\end{equation}
A component is therefore selected only when rejecting it costs
at least $0.14$ more energy than accepting it, an observable per-sample margin. The reparameterization noise
is still sampled at evaluation; for a gate whose gaps lie well
beyond one energy unit from the floor, as a trained gate's do,
the sampled decision coincides with the noise-free one. The decision depends on the pair $(E_{j,1}, E_{j,0})$ only
through the gap: adding one constant to both energies leaves
Eq.~\ref{eq:gumbel} unchanged, which is one free direction per
component (in Figure~\ref{fig:gate} the surfaces are constant
along the diagonal). The loss term
$\lambda_{\mathrm{se}} \| E \|_2^2$ with
$\lambda_{\mathrm{se}} = 10^{-10}$ removes this freedom, driving
the common shift to zero so that the pair settles at
$\pm$ half the gap, and keeps the energies bounded, so the floor
is a fixed fraction of the attainable gap rather than a
vanishing one.

\paragraph{Decision surface.} Figure~\ref{fig:gate} shows the
gate's decision as a function of the two energies of one
component, and how the decision changes with temperature.
During training (panel (a), $\tau = 0.1$ with Gumbel noise), the
expected selection probability is a ramp across the diagonal
$E_{j,0} = E_{j,1}$ about one energy unit wide. At evaluation
(panel (b), $\tau_{\mathrm{eval}} = 0.01$, noise-free surface) the
ramp collapses to a step of width $\tau_{\mathrm{eval}}$ along
the diagonal, and the decision line is the level set of that
surface at probability $\tau_{\mathrm{thres}}$: the dashed line
offset by $0.14$ from the dotted diagonal on which the
probability is $1/2$. The visible ramp, from probability $0.01$
to $0.99$, is only about $0.05$ wide in the gap, so the
five-sigma threshold places the boundary about three ramp
widths inside the saturated region; a component is selected
only when its probability is within $6 \times 10^{-7}$ of one. Panels (c) and (d) vary the
temperature. With the noise removed (c), the ramp sharpens
without limit as $\tau$ decreases, its width proportional to
$\tau$. With Gumbel noise present (d), the ramp saturates at
unit width as $\tau \to 0$: the difference of two
$\mathrm{Gumbel}(0,1)$ draws is logistic with unit scale, so the
expected probability tends to $\varsigma( E_{j,0} - E_{j,1} )$
whatever the temperature. Annealing $\tau$ during training
therefore sharpens the gate only down to about one energy unit,
which keeps every component whose gap lies within that unit
trainable; the hard decision of evaluation comes from removing
the noise, not from the temperature alone. A margin demanded
of the anchor event of Appendix~\ref{app:dyn} is a margin on
this gap.

\section{Reproducibility Details}
\label{app:repro}

This appendix records the constants and procedures behind the
solver of \S\ref{sec:solver}: the training configuration, the
checkpoint rule of \S\ref{sec:solver-checkpoint}, the termination
rule, the decoding procedure used in training and in reuse, and
the training loop as pseudocode (Algorithm~\ref{alg:1}).
Table~\ref{tab:params} collects every constant. All values are
fixed across every experiment; none is tuned per dataset. The
pool size $N$ is an experimental setting and is stated with each
experiment in \S\ref{sec:exp}.

\begin{table}[h]
\centering
\caption{EB-gMCR constants. All values are fixed across
experiments (no per-dataset tuning).}
\label{tab:params}
\small
\begin{tabular}{@{}llp{6.9cm}@{}}
\toprule
Parameter & Value & Role \\
\midrule
\multicolumn{3}{@{}l}{\textit{Gate (Eq.~\ref{eq:gumbel}; Appendix~\ref{app:gate})}} \\
$\tau_0$ & $1$ & Initial training temperature \\
$\tau_{\min}$ & $0.4$ & Floor of the temperature schedule \\
Temperature step & $0.01$ per $100$ epochs & Linear decrease of $\tau$ \\
$\tau_{\mathrm{eval}}$ & $0.01$ & Evaluation temperature \\
$\tau_{\text{thres}}$ & $0.9999994$ & Selection threshold at evaluation (five-sigma convention) \\
Exploration noise scale $s_{\mathrm{ex}}$ & $0.1 \sqrt{32} \approx 0.57$ & Energy perturbation (Eq.~\ref{eq:5}), training only \\
\midrule
\multicolumn{3}{@{}l}{\textit{Loss (Eq.~\ref{eq:9})}} \\
$\lambda$ & scheduled & Usage weight; zero until $\mathrm{nMSE} < 0.005$ \\
$(\beta_\lambda,\ \rho)$ & $(0.95,\ 1)$ & Smoothing factor and target ratio of the $\lambda$ schedule \\
$\lambda_{\text{se}}$ & $10^{-10}$ & Energy regularizer \\
$\lambda_{\text{amb}}$ & $10^{-10}$ & Ambiguity penalty (Eq.~\ref{eq:7}); RBF kernel, $\sigma = 1$ \\
Membership threshold & $0.5$ & Gate probability above which a component enters the ambiguity set \\
\midrule
\multicolumn{3}{@{}l}{\textit{Optimization}} \\
Optimizer & AdamW & lr $5\times 10^{-4}$, $\beta = (0.9, 0.995)$, weight decay $10^{-3}$, $\epsilon = 10^{-8}$ \\
Batch size & $64$ & Shuffled minibatches \\
Epoch cap & none & Training ends at the termination rule \\
\midrule
\multicolumn{3}{@{}l}{\textit{Checkpointing and termination (\S\ref{sec:solver-checkpoint})}} \\
$R^2$ bands & width $0.005$ & Lower edge $0.97$ (synthetic), $0.95$ (real data); upper edge $1$ \\
nMSE tolerance & $0.01$ & Entry into the reconstructing regime \\
Oscillation window & $100$ epochs & Rolling mean of nMSE compared with the tolerance \\
Energy ratio & $0.25$ & Termination threshold $\bar E_t / \bar E_{\mathrm{init}}$ \\
\bottomrule
\end{tabular}
\end{table}
\label{app:b}

\subsection{Training Configuration}
\label{app:b.1}

\begin{itemize}
  \item \textbf{Optimizer.} AdamW \citep{loshchilov2019adamw} with learning rate $5\times 10^{-4}$, $\beta_1=0.9$, $\beta_2=0.995$, weight decay $10^{-3}$, $\epsilon = 10^{-8}$; minibatches of $64$ samples, reshuffled every epoch. No epoch cap is set; a run ends at the termination rule of Appendix~\ref{app:b.3}.
  \item \textbf{Temperature schedule.} Training starts at $\tau_0 = 1$. After every $100$ epochs, $\tau$ decreases by $0.01$, with floor $\tau_{\min} = 0.4$. The gate of Eq.~\ref{eq:gumbel} is used in its soft form during training; no straight-through hardening is applied. Evaluation runs at $\tau_{\mathrm{eval}} = 0.01$ and selects component $j$ when $[f_e(x)]_j > \tau_{\mathrm{thres}}$ (Appendix~\ref{app:gate}).
  \item \textbf{Energy perturbation.} The exploration noise of Eq.~\ref{eq:5} is Gaussian with scale $s_{\mathrm{ex}} = 0.1 \sqrt{32} \approx 0.57$ per energy entry (the implementation draws it as $32$ accumulated steps of scale $0.1$; a constant shift applied to both energies of a pair cancels in the two-way softmax and is omitted). It is applied during training only; at evaluation the energies enter the gate unperturbed.
  \item \textbf{Architecture.} $f_e$: three linear layers with tanh activations, hidden width $4N$, output $N \times 2$. $f_c$: three linear layers with ReLU activations, hidden width $2N$, output passed through an absolute value. $f_s$: a learnable $N \times d$ matrix, passed through an absolute value and normalized to unit $\ell_2$ norm per row. Weights of $f_e$ are initialized from $\mathcal{N}(0, 0.05^2)$, all other weights from $\mathcal{N}(0, 0.01^2)$, and the profiles from $|\mathcal{N}(0, 0.1^2)|$.
  \item \textbf{Loss.} Eq.~\ref{eq:9} is evaluated per minibatch $X$. The fit term is the mean squared error over all entries of $X - \hat X$. The usage term is $\bar C(X)$, the average of $C(x)$ over the minibatch. The energy term is the mean of the squared perturbed energies (Eq.~\ref{eq:9} writes the squared norm; the mean rescales $\lambda_{\mathrm{se}}$ by the number of entries) over the minibatch, all components, and both channels. The ambiguity term is the RBF kernel $K(\mathbf{s}_j, \mathbf{s}_k) = \exp(-\|\mathbf{s}_j - \mathbf{s}_k\|^2 / 2)$ averaged over the unordered pairs of the set of components whose training gate probability exceeds $0.5$ for at least one sample of the minibatch.
  \item \textbf{Usage weight schedule.} The weight $\lambda$ is updated once per epoch from the evaluation-mode metrics of Appendix~\ref{app:b.2}. While $\mathrm{nMSE} \ge 0.005$ the target is zero. Once $\mathrm{nMSE} < 0.005$, the target is $\lambda_{\mathrm{target}} = \rho\, \mathrm{MSE}\, N / \bar E_t$, with $\rho = 1$ and $\bar E_t$ the mean selection energy of the evaluation pass (Appendix~\ref{app:b.3}): the usage weight is the reconstruction error measured in units of the current selection energy per component, so the parsimony pressure grows as the gate reaches lower energies and changes smoothly rather than jumping when a component flips. The weight follows the target through an exponential moving average, $\lambda \gets 0.95\, \lambda + 0.05\, \lambda_{\mathrm{target}}$, starting from $\lambda = 0$.
\end{itemize}

\subsection{Checkpointing}
\label{app:b.2}

\begin{itemize}
  \item \textbf{Metrics.} At the end of every epoch the model is run in evaluation mode over the training set. With $\bar{D}$ the matrix of column means,
  $R^2 = 1 - \lVert D - \hat{D}\rVert_F^2 / \lVert D - \bar{D}\rVert_F^2$ and
  $\mathrm{nMSE} = \lVert D - \hat{D}\rVert_F^2 / \lVert D \rVert_F^2$.
  The active set $A$ is the set of components selected for at least one sample, and $|A|$ its size.
  \item \textbf{Bands.} Consecutive intervals of width $0.005$ in $R^2$, closed on the left: $[0.970, 0.975), \ldots, [0.995, 1.000)$ for the synthetic experiments, and from $[0.950, 0.955)$ for the real-data experiments. One checkpoint is retained per band.
  \item \textbf{Rule.} When the epoch's $R^2$ falls in band $b$: if $b$ holds no checkpoint, store the current model; otherwise replace the stored model when the current $|A|$ is smaller, or when $|A|$ is equal and the current $R^2$ is not lower. All retained checkpoints are written out at termination.
\end{itemize}

\subsection{Termination and Oscillation Detector}
\label{app:b.3}

Let $\bar E_t$ denote the mean selection energy at epoch $t$, averaged over samples, components, and both channels in the per-epoch evaluation pass over the training data (no held-out set is required; when an evaluation set is available it is used instead). Let $t_{\mathrm{in}}$ be the first epoch with $\mathrm{nMSE} < 0.01$, and let $\bar E_{\mathrm{init}}$ be the average of $\bar E_t$ over the epochs $t \le t_{\mathrm{in}}$. On the synthetic datasets training terminates at the first epoch $t$ at which all three conditions hold; on the real datasets the oscillation condition may not occur, and those runs end at a fixed iteration limit, with all retained checkpoints written out at that point:
\begin{itemize}
  \item \textbf{Entry.} $t_{\mathrm{in}}$ has occurred.
  \item \textbf{Oscillation.} At some epoch after $t_{\mathrm{in}}$, the mean of nMSE over the preceding $100$ epochs exceeded $0.01$. Once observed, this condition stays satisfied.
  \item \textbf{Energy.} $\bar E_t \le 0.25\, \bar E_{\mathrm{init}}$.
\end{itemize}
The oscillation condition needs a full window, so a run continues for at least $100$ epochs past $t_{\mathrm{in}}$.

\begin{itemize}
\item \textbf{Observation.} On the synthetic datasets, after the solver first reaches nMSE $< 0.01$, the reconstruction can repeatedly leave and re-enter the tolerance while the active count stays high. We call this behavior oscillation. It does not always appear on the real datasets. The detector is an auxiliary stopping signal; when the minimum component count is the primary goal, running longer is preferable.
\item \textbf{Energy ratio.} On the synthetic datasets, oscillation typically begins while the ratio $\bar E_t / \bar E_{\mathrm{init}}$ is still clearly above $0.25$. In typical runs the ratio later falls to about $0.05$.
\end{itemize}

\subsection{Decoding and Reuse}
\label{app:b.5}

Decoding a sample, during the per-epoch evaluation and in reuse (Corollary~\ref{cor:reuse}; \S\ref{sec:exp-real}), runs the frozen model in evaluation mode: the exploration noise is off, the gate runs at $\tau_{\mathrm{eval}}$, and $\delta_j = \mathbf{1}[\,[f_e(x)]_j > \tau_{\mathrm{thres}}\,]$. The reconstruction is $\hat x = \sum_j \delta_j\, |f_c(x)|_j\, \mathbf{s}_j$. The reuse experiment of \S\ref{sec:exp-real} proceeds as follows.
\begin{itemize}
  \item \textbf{Checkpoint.} Every retained checkpoint of the Carbs run is decoded on the $21$ training samples; the frozen checkpoint is the one with the fewest active components across the retained bands, ties broken by the higher training $R^2$ (the 0.96--0.965 band, three components); the known component count is not used.
  \item \textbf{Test set.} Concentration triples are placed on the simplex $c_1 + c_2 + c_3 = 1$ with step $0.025$ ($861$ points). Each clean spectrum is $c^\top S$ with $S$ the reference profiles, plus noise drawn uniformly from $[0,\ 0.03\, I_{\max}]$ per channel, where $I_{\max}$ is the maximum intensity of the noiseless training data; $20$ realizations per point give $17{,}220$ test samples.
  \item \textbf{Scoring.} Each test sample is decoded as above. The per-sample nMSE is $\|x - \hat x\|^2 / \|x\|^2$, and the active count is the number of $j$ with $\delta_j = 1$.
  \item \textbf{Boundary mixtures.} Of the $861$ grid points, $120$ have one concentration equal to zero. The frozen model keeps three components on these as well, and the absent component decodes to a nonzero concentration: its median is $0.22$ of the smallest present concentration in the same sample (90th percentile $0.52$), while the nMSE stays below $0.01$. The per-sample support is read from the gate; \S\ref{sec:limits} records the same behaviour on the synthetic benchmark.
  \item \textbf{Run-to-run variation.} A second run under the same protocol, selected by the same rule, gives the same picture as Figure~\ref{fig:real}(c). An earlier run at the same training $R^2$ reached nMSE below $0.01$ on $75\%$ of the simplex. Generalization away from the training mixtures varies between runs that the certificate, which is evaluated on the training mixtures, does not distinguish.
\end{itemize}

\subsection{Algorithm}
\label{app:b.4}

The training procedure is summarized in Algorithm~\ref{alg:1}. The order of the per-epoch steps follows the implementation: evaluation, termination test, $\lambda$ update, checkpoint update, temperature update.

\begin{algorithm}[t]
\caption{EB-gMCR: training and checkpointing}
\label{alg:1}
\begin{algorithmic}[1]
\REQUIRE Observed samples $D$; pool size $N$; $R^2$ bands; constants of Table~\ref{tab:params}
\STATE Initialize the pool $f_s$, the gate $f_e$, and the concentration network $f_c$;\; $\lambda \gets 0$;\; no checkpoint in any band
\REPEAT
  \STATE \textbf{Train one epoch:} for each minibatch, select components with the noisy gate (Eqs.~\ref{eq:gumbel} and \ref{eq:5}), reconstruct (Eq.~\ref{eq:model}), and take an optimizer step on the loss of Eq.~\ref{eq:9}
  \STATE \textbf{Evaluate:} decode $D$ with the hard gate; record $R^2$, the number of components in use $|A|$, and the mean selection energy $\bar E$
  \STATE \textbf{Checkpoint:} if $R^2$ falls in a band, keep the current model for that band when it uses fewer components than the stored one (ties broken by $R^2$)
  \STATE \textbf{Schedule:} once the fit is established, move $\lambda$ toward its target; lower $\tau$ on its schedule (Appendix~\ref{app:b.1})
\UNTIL{the termination rule of Appendix~\ref{app:b.3} fires, or the iteration limit is reached}
\STATE \textbf{return} the stored checkpoint of every band entered
\end{algorithmic}
\end{algorithm}

\newpage
\section{Details of Constructing the Synthetic MCR Dataset}
\label{app:c}

Component profiles are vectors in $\mathbb{R}^{512}$, the resolution of common chemical instruments.
A pool of $N$ non-negative profiles is produced by sampling an orthonormal matrix, taking the element-wise absolute value, and normalizing each row to unit $\ell_2$ norm; the profiles are distinct by construction.
For every sample the active component count $k$ is drawn uniformly from $\{1, \ldots, 4\}$, as laboratory samples usually contain few dominant constituents, and each active concentration is drawn uniformly from $[1, 10)$, so that the component signals stay on a comparable scale.
The noiseless sample is the sum of the active profiles weighted by their concentrations.
Zero-mean Gaussian noise is then added with a per-sample variance set from the target SNR: the sample's mean power over its nonzero entries divided by $10^{\mathrm{SNR}/10}$.
The noisy sample is finally clamped from below at $10^{-6}$, so that every entry stays positive.
The synthetic benchmarks and the runtime table ran on Ubuntu 24.04 LTS with an Intel i9-14900K CPU and NVIDIA RTX A6000 GPUs; the Carbs rerun and the ARD-NMF seed extension ran on an Intel i9-12900K with NVIDIA RTX 3090 GPUs, so wall times across the two sets are not comparable.

\begin{algorithm}[H]
\caption{Mixed Data Sampling Process}
\label{alg:2}
\begin{algorithmic}[1]
\REQUIRE Component matrix $S \in \mathbb{R}^{N \times d}$ with non-negative rows of unit $\ell_2$ norm, dataset size $M$, active-count range $[K_{\min}, K_{\max}]$, concentration range $[c_{\text{low}}, c_{\text{high}}]$, target SNR (dB)
\FOR{$i = 1$ \textbf{to} $M$}
    \STATE Sample $k \sim \mathcal{U}\{K_{\min}, K_{\max}\}$
    \STATE Select $k$ component indices $J_i \subset \{1, \ldots, N\}$
    \FOR{each $j \in J_i$}
        \STATE Sample $c_j \sim \mathcal{U}(c_{\text{low}}, c_{\text{high}})$
    \ENDFOR
    \STATE $X_i = \sum_{j \in J_i} c_j \cdot S_{j,:}$
    \IF{SNR specified}
        \STATE $\sigma_i^2 \gets \bigl(\text{mean of } X_{i,l}^2 \text{ over the entries } l \text{ with } X_{i,l} \neq 0\bigr) \big/ 10^{\mathrm{SNR}/10}$
        \STATE $X_i \gets X_i + \varepsilon_i$ on those entries, $\varepsilon_i \sim \mathcal{N}(0, \sigma_i^2 I)$
    \ENDIF
    \STATE $X_i \gets \max(X_i,\, 10^{-6})$ element-wise
\ENDFOR
\STATE \textbf{return} $X \in \mathbb{R}^{M \times d}$
\end{algorithmic}
\end{algorithm}

\section{Benchmark Search Wrapper for Baseline MCR Solvers}
\label{app:d}

Every baseline except ARD-NMF needs the component count as input. On the synthetic data the true count $K^*$ is known, so each baseline is run inside a search around it (Algorithm~\ref{alg:3}).
The wrapper tries $K = \lfloor s K^* \rfloor$ for $s$ from $1.20$ down to $0.80$ in steps of $0.05$ and returns the smallest $K$ whose $R^2$ falls in the success band of the setting: $[0.98, 0.985)$ at 20\,dB and $[0.99, 0.995)$ at 30\,dB for the $4N$ data of Table~\ref{tab:e2}, and $[0.975, 0.98)$ for the $8N$ data at 20\,dB.
When no $K$ reaches the band, the wrapper returns the $K$ with the highest $R^2$.

\begin{algorithm}[H]
\caption{Component-Count Search for Baseline Methods}
\label{alg:3}
\begin{algorithmic}[1]
\REQUIRE Data $D$, solver $f(\cdot, K)$, true count $K^*$, ratios $R_{\text{search}} = \{1.20, 1.15, \ldots, 0.80\}$ in descending order, success band $[r_{\text{lo}}, r_{\text{hi}})$
\STATE Initialize: $R^2_{\text{sel}} \gets -\infty$, $\text{hit} \gets \text{false}$
\FOR{$s$ \textbf{in} $R_{\text{search}}$}
    \STATE $K_{\text{try}} \gets \lfloor s \cdot K^* \rfloor$
    \STATE $R^2 \gets f(D, K_{\text{try}})$
    \IF{$r_{\text{lo}} \le R^2 < r_{\text{hi}}$}
        \STATE $\text{hit} \gets \text{true}$; $R^2_{\text{sel}} \gets R^2$; $K_{\text{sel}} \gets K_{\text{try}}$ \COMMENT{$s$ descends, so the last hit is the smallest $K$ in band}
    \ELSIF{\textbf{not} hit \textbf{and} $R^2 > R^2_{\text{sel}}$}
        \STATE $R^2_{\text{sel}} \gets R^2$; $K_{\text{sel}} \gets K_{\text{try}}$
    \ENDIF
\ENDFOR
\STATE \textbf{return} $R^2_{\text{sel}}, K_{\text{sel}}$
\end{algorithmic}
\end{algorithm}

\section{Synthetic-Data Evaluation}
\label{app:e}

Every result on the synthetic mixtures of Appendix~\ref{app:c}
is collected here: the component counts against the baselines
(\S\ref{app:e-count}), the counts per checkpoint band
(\S\ref{app:e-bands}), ARD-NMF on the same grid
(\S\ref{app:e-ard}), wall-clock times (\S\ref{app:e-time}), and
the per-sample decoding behind the support finding of
\S\ref{sec:discussion} (\S\ref{app:e-support}). EB-gMCR is run
five times per setting and the baselines one hundred times;
every entry is a mean $\pm$ one standard deviation over runs.
The ablation at $N = 256$ is Table~\ref{tab:ablation} in
\S\ref{sec:exp-ablation}.

\subsection{Component Count Against the Baselines}
\label{app:e-count}

Tables~\ref{tab:e2} and \ref{tab:e2b} report, for each true
component count $N$ and noise level, the success rate, the
reconstruction $R^2$, and the estimated component count (EC) of
every solver on $4N$ samples. The success rate is the fraction
of runs whose $R^2$ reaches $0.98$ at 20\,dB or $0.99$ at
30\,dB; it is a fit criterion only, which is why ARD-NMF at the
oracle $\phi$ passes it at 30\,dB while returning about twice
the true count. The EC of EB-gMCR is read from the checkpoint
band that matches the wrapper's success band
(Appendix~\ref{app:d}), $[0.98, 0.985)$ at 20\,dB and
$[0.99, 0.995)$ at 30\,dB; every other band is in
Table~\ref{tab:e1}. The baselines receive the count through the
search of Appendix~\ref{app:d}, so their EC is the count the
search returned. These two tables are the numbers behind
Figure~\ref{fig:count} and the count statements of
\S\ref{sec:exp-synth}.

\begin{table*}[t]
\centering
\caption{Comparison of EB-gMCR with six baseline solvers on $4N$ synthetic samples, $N = 16$ to $96$. ARD-NMF (oracle $\phi$) sets $\phi$ to the true noise variance of the draw; ARD-NMF (tuned) uses $\phi = 0.03$, chosen by hand at 30\,dB. Success at 30\,dB counts the $R^2$ criterion only, which the oracle setting passes while returning about twice the true count.}
\label{tab:e2}
\scriptsize
\begin{tabular}{clccc|ccc}
\toprule
\multirow{2}{*}{\textbf{Components}} & \multirow{2}{*}{\textbf{Method}}
  & \multicolumn{3}{c|}{\textbf{20dB (success $R^2 \geq 0.98$)}}
  & \multicolumn{3}{c}{\textbf{30dB (success $R^2 \geq 0.99$)}} \\
\cmidrule(r){3-5} \cmidrule(l){6-8}
& & \textbf{Succ.} & \textbf{$R^2$} & \textbf{EC}
  & \textbf{Succ.} & \textbf{$R^2$} & \textbf{EC} \\
\midrule
\multirow{8}{*}{16}
& NMF & 0.33 & 0.978±0.002 & 18.0±1.1 & 1.00 & 0.995±0.002 & 17.0±1.4 \\
& Sparse-NMF & 0.00 & 0.761±0.034 & 13.7±2.0 & 0.00 & 0.776±0.031 & 13.4±1.9 \\
& Bayes-NMF & 0.18 & 0.978±0.002 & 18.3±0.9 & 1.00 & \textbf{0.996±0.002} & \textbf{17.6±1.4} \\
& ICA & 0.00 & 0.106±0.172 & 14.7±1.9 & 0.00 & 0.079±0.199 & 13.8±1.4 \\
& MCR-ALS & 0.80 & 0.981±0.003 & 16.4±0.9 & 0.92 & 0.994±0.008 & 16.1±1.4 \\
& EB-gMCR & \textbf{-} & \textbf{0.983±0.001} & \textbf{16.2±1.1} & - & 0.993±0.002 & 15.4±0.5 \\
& ARD-NMF (oracle $\phi$) & 0.94 & 0.983±0.002 & 18.4±1.2 & 1.00 & 0.999±0.000 & 45.9±4.2 \\
& ARD-NMF (tuned $\phi{=}0.03$) & -- & -- & -- & 1.00 & 0.998±0.000 & 16.1±0.3 \\
\midrule
\multirow{8}{*}{32}
& NMF         & 0.01 & 0.976±0.002 & 36.4±1.6 & 1.00 & 0.992±0.001 & 32.7±1.2 \\
& Sparse-NMF & 0.00 & 0.580±0.034 & 25.5±1.8 & 0.00 & 0.596±0.034 & 25.4±1.5 \\
& Bayes-NMF  & 0.08 & 0.978±0.002 & 37.3±1.3 & 1.00 & \textbf{0.993±0.002} & \textbf{33.1±2.1} \\
& ICA         & 0.00 & 0.230±0.123 & 30.4±3.5 & 0.00 & 0.131±0.188 & 29.1±2.9 \\
& MCR-ALS     & 0.12 & 0.956±0.048 & 29.5±2.8 & 0.52 & 0.986±0.015 & 30.3±2.1 \\
& EB-gMCR     & -    & \textbf{0.982±0.001} & \textbf{31.4±0.9} & -    & \textbf{0.992±0.001} & \textbf{31.4±2.2} \\
& ARD-NMF (oracle $\phi$) & 1.00 & 0.983±0.001 & 35.1±1.5 & 1.00 & 0.999±0.000 & 85.6±6.0 \\
& ARD-NMF (tuned $\phi{=}0.03$) & -- & -- & -- & 1.00 & 0.998±0.000 & 32.1±0.4 \\
\midrule
\multirow{8}{*}{48}
& NMF         & 0.00 & 0.972±0.003 & 55.8±1.6 & 0.83 & 0.991±0.001 & 53.9±2.7 \\
& Sparse-NMF & 0.00 & 0.524±0.019 & 38.8±2.2 & 0.00 & 0.536±0.019 & 38.9±3.0 \\
& Bayes-NMF  & 0.03 & 0.977±0.002 & 56.2±1.4 & 1.00 & \textbf{0.992±0.001} & \textbf{49.5±1.8} \\
& ICA         & 0.00 & 0.200±0.141 & 47.8±5.9 & 0.00 & 0.126±0.180 & 43.8±3.6 \\
& MCR-ALS     & 0.00 & 0.870±0.139 & 41.5±3.7 & 0.11 & 0.923±0.108 & 42.1±3.2 \\
& EB-gMCR     & -    & \textbf{0.981±0.001} & \textbf{48.0±1.2} & -    & 0.992±0.001 & 50.0±4.2 \\
& ARD-NMF (oracle $\phi$) & 1.00 & 0.983±0.001 & 50.6±1.3 & 1.00 & 0.999±0.000 & 119.5±7.2 \\
& ARD-NMF (tuned $\phi{=}0.03$) & -- & -- & -- & 1.00 & 0.998±0.000 & 48.0±0.3 \\
\midrule
\multirow{8}{*}{64}
& NMF         & 0.00 & 0.967±0.003 & 74.7±2.0 & 0.09 & 0.987±0.002 & 74.4±2.3 \\
& Sparse-NMF & 0.00 & 0.509±0.019 & 51.0±0.0 & 0.00 & 0.518±0.018 & 51.1±0.6 \\
& Bayes-NMF  & 0.00 & 0.976±0.002 & 74.9±1.8 & 1.00 & \textbf{0.992±0.001} & \textbf{67.9±2.3} \\
& ICA         & 0.00 & 0.215±0.114 & 63.3±8.3 & 0.00 & 0.172±0.143 & 58.5±4.7 \\
& MCR-ALS     & 0.00 & 0.770±0.226 & 54.2±4.1 & 0.00 & 0.876±0.183 & 56.5±4.9 \\
& EB-gMCR     & -    & \textbf{0.981±0.001} & \textbf{64.8±2.6} & -    & 0.991±0.001 & 74.4±5.0 \\
& ARD-NMF (oracle $\phi$) & 1.00 & 0.983±0.001 & 66.3±1.4 & 1.00 & 0.999±0.000 & 150.0±8.7 \\
& ARD-NMF (tuned $\phi{=}0.03$) & -- & -- & -- & 1.00 & 0.998±0.000 & 64.0±0.4 \\
\midrule
\multirow{8}{*}{96}
& NMF         & 0.00 & 0.955±0.004 & 113.3±2.9 & 0.00 & 0.977±0.003 & 113.7±2.4 \\
& Sparse-NMF & 0.00 & 0.504±0.015 & 76.3±1.6  & 0.00 & 0.515±0.015 & 76.1±1.0 \\
& Bayes-NMF  & 0.00 & 0.974±0.002 & 114.2±1.9 & 1.00 & 0.991±0.001 & 108.0±3.8 \\
& ICA         & 0.00 & 0.245±0.077 & 96.0±12.0 & 0.00 & 0.194±0.100 & 89.5±7.8 \\
& MCR-ALS     & 0.00 & 0.782±0.237 & 86.4±6.9  & 0.00 & 0.882±0.041 & 85.9±6.2 \\
& EB-gMCR     & -    & \textbf{0.981±0.001} & \textbf{103.8±12.0} & -    & \textbf{0.991±0.000} & \textbf{102.4±4.8} \\
& ARD-NMF (oracle $\phi$) & 1.00 & 0.983±0.001 & 97.3±1.3 & 1.00 & 0.999±0.000 & 201.2±8.6 \\
& ARD-NMF (tuned $\phi{=}0.03$) & -- & -- & -- & 1.00 & 0.998±0.000 & 95.9±0.3 \\
\bottomrule
\end{tabular}
\end{table*}

\begin{table*}[t]
\centering
\caption{Comparison of EB-gMCR with six baseline solvers on $4N$ synthetic samples, $N = 128$ to $256$; columns as in Table~\ref{tab:e2}.}
\label{tab:e2b}
\scriptsize
\begin{tabular}{clccc|ccc}
\toprule
\multirow{2}{*}{\textbf{Components}} & \multirow{2}{*}{\textbf{Method}}
  & \multicolumn{3}{c|}{\textbf{20dB (success $R^2 \geq 0.98$)}}
  & \multicolumn{3}{c}{\textbf{30dB (success $R^2 \geq 0.99$)}} \\
\cmidrule(r){3-5} \cmidrule(l){6-8}
& & \textbf{Succ.} & \textbf{$R^2$} & \textbf{EC}
  & \textbf{Succ.} & \textbf{$R^2$} & \textbf{EC} \\
\midrule
\multirow{8}{*}{128}
& NMF         & 0.00 & 0.938±0.005 & 152.0±2.2 & 0.00 & 0.962±0.006 & 151.4±3.0 \\
& Sparse-NMF & 0.00 & 0.501±0.011 & 102.0±0.0 & 0.00 & 0.513±0.014 & 102.0±0.0 \\
& Bayes-NMF  & 0.00 & 0.970±0.002 & 152.0±2.2 & 0.66 & 0.990±0.001 & 150.7±3.5 \\
& ICA         & 0.00 & 0.275±0.065 & 129.3±13.7 & 0.00 & 0.235±0.085 & 120.0±9.9 \\
& MCR-ALS     & 0.00 & 0.703±0.020 & 116.6±8.8  & 0.00 & 0.763±0.035 & 118.3±7.4 \\
& EB-gMCR     & -    & \textbf{0.981±0.000} & \textbf{147.8±14.9} & -    & \textbf{0.991±0.000} & \textbf{137.6±3.8} \\
& ARD-NMF (oracle $\phi$) & 1.00 & 0.984±0.001 & 128.7±0.9 & 1.00 & 0.999±0.000 & 248.6±9.9 \\
& ARD-NMF (tuned $\phi{=}0.03$) & -- & -- & -- & 1.00 & 0.998±0.000 & 127.8±0.4 \\
\midrule
\multirow{8}{*}{160}
& NMF         & 0.00 & 0.923±0.005 & 191.2±2.7 & 0.00 & 0.945±0.005 & 190.8±2.9 \\
& Sparse-NMF & 0.00 & 0.500±0.012 & 128.0±0.0 & 0.00 & 0.513±0.011 & 128.0±0.0 \\
& Bayes-NMF  & 0.00 & 0.967±0.002 & 191.4±2.2 & 0.18 & 0.988±0.002 & 191.3±2.3 \\
& ICA         & 0.00 & 0.309±0.050 & 151.4±17.0 & 0.00 & 0.270±0.068 & 146.3±11.6 \\
& MCR-ALS     & 0.00 & 0.445±0.055 & 140.0±10.9 & 0.00 & 0.480±0.080 & 142.6±10.9 \\
& EB-gMCR     & -    & \textbf{0.979±0.002} & \textbf{187.2±15.1} & -    & \textbf{0.991±0.001} & \textbf{169.6±5.3} \\
& ARD-NMF (oracle $\phi$) & 1.00 & 0.984±0.000 & 160.6±0.9 & 1.00 & 0.999±0.000 & 293.8±9.6 \\
& ARD-NMF (tuned $\phi{=}0.03$) & -- & -- & -- & 1.00 & 0.997±0.000 & 159.7±0.5 \\
\midrule
\multirow{8}{*}{192}
& NMF         & 0.00 & 0.906±0.005 & 229.2±2.7 & 0.00 & 0.928±0.005 & 228.2±4.1 \\
& Sparse-NMF & 0.00 & 0.497±0.010 & 153.0±0.0 & 0.00 & 0.510±0.009 & 153.0±0.0 \\
& Bayes-NMF  & 0.00 & 0.962±0.002 & 228.9±3.4 & 0.00 & 0.985±0.002 & 229.4±2.4 \\
& ICA         & 0.00 & 0.316±0.048 & 171.4±17.2 & 0.00 & 0.285±0.054 & 170.7±12.7 \\
& MCR-ALS     & 0.00 & -0.029±0.117 & 161.3±9.8  & 0.00 & -0.002±0.126 & 164.1±12.2 \\
& EB-gMCR     & -    & \textbf{0.979±0.000} & \textbf{213.5±11.6} & -    & \textbf{0.989±0.002} & \textbf{191.8±4.6} \\
& ARD-NMF (oracle $\phi$) & 1.00 & 0.984±0.000 & 192.0±1.0 & 1.00 & 0.999±0.000 & 340.9±11.6 \\
& ARD-NMF (tuned $\phi{=}0.03$) & -- & -- & -- & 1.00 & 0.997±0.000 & 191.5±0.7 \\
\midrule
\multirow{8}{*}{224}
& NMF         & 0.00 & 0.893±0.004 & 266.9±3.3 & 0.00 & 0.914±0.004 & 267.2±2.8 \\
& Sparse-NMF & 0.00 & 0.497±0.009 & 268.0±0.0 & 0.00 & 0.510±0.009 & 268.0±0.0 \\
& Bayes-NMF  & 0.00 & 0.957±0.002 & 267.7±1.9 & 0.00 & 0.981±0.002 & 267.2±2.8 \\
& ICA         & 0.00 & 0.321±0.042 & 196.0±14.0 & 0.00 & 0.292±0.049 & 191.6±11.6 \\
& MCR-ALS     & 0.00 & -0.676±0.164 & 185.9±9.2  & 0.00 & -0.617±0.150 & 186.4±10.2 \\
& EB-gMCR     & -    & \textbf{0.979±0.002} & \textbf{251.5±13.0} & -    & \textbf{0.989±0.002} & \textbf{224.8±6.3} \\
& ARD-NMF (oracle $\phi$) & 1.00 & 0.984±0.000 & 223.7±0.9 & 1.00 & 0.998±0.000 & 386.1±11.6 \\
& ARD-NMF (tuned $\phi{=}0.03$) & -- & -- & -- & 1.00 & 0.997±0.000 & 223.1±0.8 \\
\midrule
\multirow{8}{*}{256}
& NMF         & 0.00 & 0.884±0.004 & 304.9±4.8 & 0.00 & 0.902±0.004 & 305.1±4.6 \\
& Sparse-NMF & 0.00 & 0.497±0.009 & 307.0±0.0 & 0.00 & 0.509±0.009 & 307.0±0.0 \\
& Bayes-NMF  & 0.00 & 0.951±0.003 & 306.0±3.5 & 0.00 & 0.977±0.002 & 306.6±2.2 \\
& ICA         & 0.00 & 0.341±0.031 & 212.8±9.0 & 0.00 & 0.307±0.036 & 214.3±9.2 \\
& MCR-ALS     & 0.00 & -1.473±0.242 & 212.2±11.4  & 0.00 & -1.420±0.250 & 213.6±12.7 \\
& EB-gMCR     & -    & \textbf{0.978±0.003} & \textbf{286.7±6.4} & -    & \textbf{0.990±0.000} & \textbf{248.8±1.1} \\
& ARD-NMF (oracle $\phi$) & 1.00 & 0.984±0.000 & 255.4±1.0 & 1.00 & 0.998±0.000 & 435.2±12.1 \\
& ARD-NMF (tuned $\phi{=}0.03$) & -- & -- & -- & 1.00 & 0.997±0.000 & 254.6±1.1 \\
\bottomrule
\end{tabular}
\end{table*}B

\subsection{Checkpoint Bands}
\label{app:e-bands}

\begin{figure}[t]
\centering
\includegraphics[width=\textwidth]{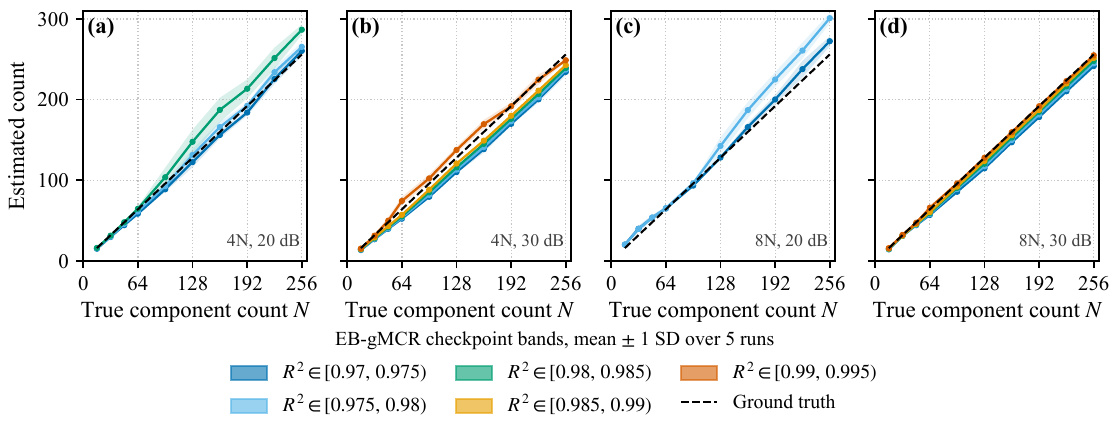}
\caption{EB-gMCR checkpoints on the synthetic benchmark:
estimated against true component count per $R^2$ band, for
$4N$ and $8N$ samples at 20 and 30\,dB (mean $\pm$ one SD over
five runs).}
\label{fig:bands}
\end{figure}

Table~\ref{tab:e1} lists the EC of EB-gMCR in every retained
$R^2$ band, for $4N$ and $8N$ samples at both noise levels, and
Figure~\ref{fig:bands} plots the same numbers. Two facts read off
the table. Within one run the EC rises with the band, since a
stricter fit admits components and never removes them; and the
band that holds the true count moves with the noise level, which
is the sense in which the band is the solver's statement about
the noise (\S\ref{sec:exp-synth}). At 30\,dB the top band is
within a few percent of the truth at every $N$; at 20\,dB the
top band exceeds the truth, by more with $8N$ samples than with
$4N$ at small $N$.

\begin{landscape}
\begin{table}[ht]
\centering
\caption{Exact numerical values underlying Fig.~\ref{fig:bands}: mean $\pm$ 1 SD estimated component counts returned by EB-gMCR for each dataset size (4$N$, 8$N$) and noise level (20~dB, 30~dB).}
\label{tab:e1}
\scriptsize
\begin{tabular}{lcccccccccc}
\toprule
\textbf{R² band} & \textbf{16} & \textbf{32} & \textbf{48} & \textbf{64} & \textbf{96} & \textbf{128} & \textbf{160} & \textbf{192} & \textbf{224} & \textbf{256} \\
\midrule
\multicolumn{11}{l}{\textbf{Data size = 4N (SNR = 20dB)}} \\
0.97–0.975 & 15.2$\pm$1.6 & 29.8$\pm$0.4 & 44.4$\pm$0.5 & 58.6$\pm$1.1 & 88.8$\pm$3.0 & 122.6$\pm$6.5 & 156.2$\pm$4.3 & 184.0$\pm$3.7 & 225.6$\pm$6.3 & 260.4$\pm$6.6 \\
0.975–0.98 & 15.8$\pm$1.3 & 30.4$\pm$0.5 & 46.0$\pm$0.7 & 61.4$\pm$1.3 & 92.8$\pm$4.9 & 131.4$\pm$6.9 & 166.0$\pm$6.8 & 192.4$\pm$3.0 & 234.0$\pm$6.0 & 265.4$\pm$5.6 \\
0.98–0.985 & 16.2$\pm$1.1 & 31.4$\pm$0.9 & 48.0$\pm$1.2 & 64.8$\pm$2.6 & 103.8$\pm$12.0 & 147.8$\pm$14.9 & 187.2$\pm$15.1 & 213.5$\pm$11.6 & 251.5$\pm$13.0 & 286.7$\pm$6.4 \\
\midrule
\multicolumn{11}{l}{\textbf{Data size = 4N (SNR = 30dB)}} \\
0.97--0.975 & 13.6$\pm$0.5 & 27.0$\pm$0.7 & 39.8$\pm$0.4 & 52.4$\pm$1.1 & 79.6$\pm$2.1 & 110.2$\pm$1.9 & 138.6$\pm$4.3 & 170.0$\pm$2.0 & 200.4$\pm$3.8 & 234.4$\pm$4.4 \\
0.975--0.98 & 14.2$\pm$0.4 & 27.6$\pm$0.5 & 40.4$\pm$0.5 & 53.8$\pm$0.8 & 82.0$\pm$2.4 & 113.2$\pm$0.8 & 141.2$\pm$3.0 & 173.2$\pm$1.8 & 202.8$\pm$2.5 & 237.2$\pm$2.6 \\
0.98--0.985 & 14.4$\pm$0.5 & 28.2$\pm$0.8 & 41.8$\pm$0.4 & 55.2$\pm$1.1 & 85.2$\pm$2.8 & 116.0$\pm$1.9 & 145.2$\pm$2.2 & 176.2$\pm$0.8 & 207.0$\pm$2.0 & 239.2$\pm$1.9 \\
0.985--0.99 & 14.8$\pm$0.4 & 29.2$\pm$0.8 & 43.2$\pm$0.4 & 57.2$\pm$1.1 & 88.4$\pm$1.9 & 120.8$\pm$1.1 & 149.2$\pm$1.6 & 180.0$\pm$1.0 & 211.2$\pm$1.5 & 242.8$\pm$1.3 \\
0.99--0.995 & 15.4$\pm$0.5 & 31.4$\pm$2.2 & 50.0$\pm$4.2 & 74.4$\pm$5.0 & 102.4$\pm$4.8 & 137.6$\pm$3.8 & 169.6$\pm$5.3 & 191.8$\pm$4.6 & 224.8$\pm$6.3 & 248.8$\pm$1.1 \\
\midrule
\multicolumn{11}{l}{\textbf{Data size = 8N (SNR = 20dB)}} \\
0.97--0.975   & 20.4$\pm$1.9 & 40.0$\pm$3.7 & 53.8$\pm$2.9 & 65.0$\pm$2.2 & 93.4$\pm$0.9 & 128.2$\pm$2.5 & 166.0$\pm$4.8 & 200.2$\pm$5.5 & 237.8$\pm$10.4 & 272.4$\pm$6.3 \\
0.975--0.98   & 20.2$\pm$1.6 & 39.8$\pm$4.4 & 53.8$\pm$2.9 & 65.6$\pm$1.9 & 96.4$\pm$0.9 & 142.6$\pm$11.0 & 187.0$\pm$10.0 & 225.0$\pm$9.0 & 260.8$\pm$9.9 & 301.0$\pm$6.4 \\
\midrule
\multicolumn{11}{l}{\textbf{Data size = 8N (SNR = 30dB)}} \\
0.97--0.975   & 14.4$\pm$0.5 & 31.2$\pm$0.8 & 44.4$\pm$1.7 & 57.0$\pm$0.7 & 85.8$\pm$0.8 & 114.8$\pm$1.8 & 147.2$\pm$1.9 & 178.4$\pm$1.7 & 210.4$\pm$0.7 & 241.6$\pm$2.3 \\
0.975--0.98   & 14.8$\pm$0.4 & 31.2$\pm$0.8 & 44.8$\pm$1.3 & 58.2$\pm$0.6 & 89.2$\pm$2.5 & 117.4$\pm$1.5 & 150.0$\pm$1.9 & 181.6$\pm$1.1 & 213.2$\pm$3.3 & 245.0$\pm$1.1 \\
0.98--0.985   & 15.0$\pm$0.4 & 31.2$\pm$0.8 & 45.4$\pm$0.9 & 59.2$\pm$0.7 & 90.6$\pm$2.2 & 120.6$\pm$1.1 & 153.0$\pm$1.9 & 185.0$\pm$1.4 & 216.4$\pm$2.3 & 248.0$\pm$1.1 \\
0.985--0.99   & 15.4$\pm$0.5 & 31.2$\pm$0.5 & 46.0$\pm$0.7 & 61.0$\pm$0.5 & 92.2$\pm$1.6 & 124.0$\pm$1.6 & 156.0$\pm$1.2 & 187.2$\pm$1.1 & 220.0$\pm$1.3 & 251.6$\pm$0.5 \\
0.99--0.995   & 16.0$\pm$0.5 & 32.4$\pm$0.5 & 47.4$\pm$0.5 & 66.0$\pm$0.5 & 96.2$\pm$0.8 & 128.0$\pm$1.6 & 159.6$\pm$1.1 & 191.8$\pm$0.4 & 223.0$\pm$0.8 & 255.0$\pm$1.4 \\
\bottomrule
\end{tabular}
\end{table}
\end{landscape}

\subsection{ARD-NMF on the Synthetic Grid}
\label{app:e-ard}

\begin{figure}[t]
\centering
\includegraphics[width=0.55\textwidth]{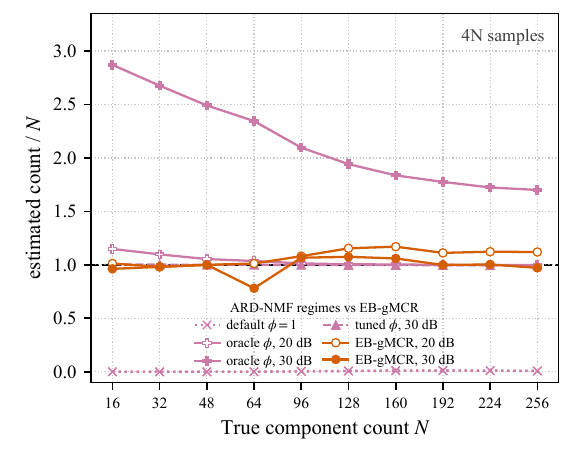}
\caption{ARD-NMF on the synthetic grid: estimated over true
component count against $N$ on $4N$ samples, at the default
$\phi$, at $\phi$ equal to the true noise variance, and at the
hand-chosen $\phi$, with EB-gMCR for reference.}
\label{fig:ardnmf-synth}
\end{figure}

ARD-NMF is the one baseline that chooses its count, through the
noise-scale hyperparameter $\phi$. Figure~\ref{fig:ardnmf-synth}
plots its EC over the true count against $N$ on $4N$ samples in
three settings: the default $\phi = 1$, which returns one or two
components at every $N$; $\phi$ set to the true noise variance
of the draw, which returns the true count at 20\,dB and $1.7$ to
$2.9$ times the truth at 30\,dB while reporting
$R^2 \approx 0.998$; and $\phi = 0.03$, chosen with the true
count in view at 30\,dB, which returns the exact count. The rows
of Tables~\ref{tab:e2} and \ref{tab:e2b} give the numbers.
Which $\phi$ is right cannot be read from a single run, since
$\phi$ is a variance scale with no check of its own; the
corresponding input of EB-gMCR, the $R^2$ band, is the fit
condition of the certificate and is checked on every checkpoint.
The same sweep on the real datasets is in Appendix~\ref{app:f}.

\subsection{Runtime}
\label{app:e-time}

Table~\ref{tab:runtime} gives wall-clock times. EB-gMCR takes hours per run where the NMF family completes a search over counts in minutes and ARD-NMF fits once in about a minute; it is slower by one to two orders of magnitude. The solver is a realization of the formulation of \S\ref{sec:solver}, run with one set of hyperparameters (Table~\ref{tab:params}) across every setting; runtime was not tuned, and the search it removes is the choice of an interval of counts, not the cost of a fit.

\begin{table}[t]
\centering
\caption{Cumulative wall-clock time (hours) to search up to $N=256$ at different intervals (SNR=20\,dB). EB-gMCR automatically determines the component count without interval specification. ARD-NMF is a single fit rather than a search (median $74$\,s per fit at $N=256$, 4N, 20\,dB, on different hardware) and is not entered.}
\label{tab:runtime}
\small
\begin{tabular}{c|ccccc|c}
\toprule
\textbf{Interval} & \textbf{NMF} & \textbf{Sp-NMF} & \textbf{Ba-NMF} & \textbf{ICA} & \textbf{MCR-ALS} & \textbf{EB-gMCR} \\
\midrule
\multicolumn{7}{c}{\textit{4N samples (1024)}} \\
\midrule
32 & 0.03 & 0.01 & 0.07 & 1.06 & 0.39 & \multirow{5}{*}{3.3±0.8} \\
16 & 0.06 & 0.02 & 0.14 & 1.75 & 0.75 & \\
8 & 0.10 & 0.03 & 0.27 & 3.16 & 1.41 & \\
4 & 0.20 & 0.06 & 0.52 & 6.03 & 2.71 & \\
2 & 0.27 & 0.10 & 0.68 & 10.1 & 4.28 & \\
\midrule
\multicolumn{7}{c}{\textit{8N samples (2048)}} \\
\midrule
32 & 0.01 & 0.01 & 0.02 & 1.49 & 0.48 & \multirow{5}{*}{2.1±0.2} \\
16 & 0.01 & 0.02 & 0.03 & 2.71 & 0.89 & \\
8 & 0.02 & 0.03 & 0.06 & 5.15 & 1.81 & \\
4 & 0.03 & 0.05 & 0.08 & 7.46 & 1.64 & \\
2 & 0.07 & 0.10 & 0.17 & 15.8 & 3.77 & \\
\bottomrule
\end{tabular}
\end{table}

\subsection{Per-Sample Support Recovery}
\label{app:e-support}

The tables above report the component count, the size of the
pool a run keeps. Theorem~\ref{thm:cert} is a statement about
each sample's support, and its certificate has two conditions:
the fit of every sample, which is checked without the truth, and
the total usage, which is compared with the true total and is
known here because the data are synthetic. Whether the supports
are the true ones is a separate question. The theory answers it
under one condition on the data: at minimal usage the solution
set is a relabeling of the pool when every component has an
isolating pair, two samples that share exactly that component,
and otherwise contains rotations inside each group of components
the samples never separate (\S\ref{sec:theory},
Appendix~\ref{app:toy}).

Table~\ref{tab:support} reports three kinds of quantity, and
they should be read apart. The usage ratio is the certificate's
usage condition and needs no matching. Min cos and misses report
the profile matching: whether each true component has a learned
slot of its own. Exact and no-miss report the gate's per-sample
selection against the matched slots. Read this way the table
says one thing in most settings: the profiles are recovered and pinned (min cos above $0.97$ and no misses at $N \le 32$ at 20\,dB and at $N = 16$ at 30\,dB; a few blended slots with min cos $0.80$ to $0.90$ elsewhere), and the failure is confined to the gate, which selects
seven to eleven times the true usage with the true components
almost always inside the selection (no-miss near one, exact
zero). The two 30\,dB rows with usage near one are the reverse
case: the gate is near usage-minimal, and the matching is the
weakest in the table. The isolating-pairs column has two
readings. The count on the gate's supports is zero wherever the
gate over-selects, since an over-full support cannot share
exactly one component with another; that part is the solver's.
The ceiling in parentheses is evaluated on the true supports and
is a property of the sampled data: at $4N$ samples between a
quarter and a third of the components admit a pair, so the
condition of Conjecture~R is not met by these data for any
solver, and the exact column is not a test of that conjecture.

Table~\ref{tab:support-bands} decodes the same runs at every
retained checkpoint band and shows the two states in one
trajectory. Early checkpoints, taken within the first fifth of a
run at the highest $R^2$ bands, have usage ratios of $1.1$ to
$2.4$: the gate is near usage-minimal, while the pool is still
incomplete (no-miss $0.69$ to $0.97$). Later checkpoints have a
complete and pinned pool (no-miss up to $1.00$, min cos up to
$0.98$) and a saturated gate (usage $4$ to $19$) whose extra
slots carry decoded concentrations of $0.01$ to $0.10$ against
true concentrations of at least one, which is why a threshold at $0.5$ recovers the exact support on $56$ to $89\%$ of samples at those checkpoints.
The two conditions of the certificate are each reached, at
different times, and are not reached together at any retained
checkpoint. That is the precise form of the limitation
\S\ref{sec:discussion} states and of the attempt the solver
represents.

\paragraph{Columns of Tables~\ref{tab:support} and \ref{tab:support-bands}.}
Both tables decode the runs that produce the component counts of
Table~\ref{tab:e2}, sample by sample. Table~\ref{tab:support}
reports the checkpoint band of that table; Table~\ref{tab:support-bands}
reports every retained band, with the fraction of the run at
which each checkpoint was taken (epoch). A learned slot is
matched to a true component by the Hungarian assignment on
profile cosine, over the slots active in at least one sample.
Usage ratio is the total number of components the gate selects
over all samples divided by the true total; one is
usage-minimal. Exact is the fraction of samples whose gate
support equals the true support after the matching; no-miss is
the fraction whose gate support contains the true support. Min
cos is the smallest cosine between a learned profile and its
matched true profile. Misses is the number of true components
left without a slot because fewer slots were active than
components; no cosine cutoff is applied. Isolating pairs counts
the components that admit a pair in the sense of Conjecture~R,
evaluated on the gate's supports, over the $N$ true components,
all of which are active in these data; the ceiling in
parentheses is the same count on the true supports. The two
columns headed $\hat c > 0.5$ recompute usage and exact after
each gate support is intersected with the set of slots whose
decoded concentration exceeds $0.5$, a level fixed in advance
below the smallest true concentration of one; this cut lowers
no-miss by at most $0.012$ in any setting. Spurious $\hat c$ in
Table~\ref{tab:support-bands} is the median decoded
concentration on selected slots outside the true support.

\paragraph{Reading the result under the theory.}
Three parts of the theory say what these numbers should look
like, and the numbers follow them. First, the direction of the
gate's failure is the one Theorem~\ref{thm:ident} gives for a
usage penalty below its $\lambda$-window: extra components that
fit noise are kept, and no true component is dropped. The table
shows this direction exactly, no-miss near one with exact zero,
and never the opposite. The solver's per-slot penalty is
$\lambda / N$ with $N = 1024$ (\S\ref{sec:solver-objective}),
and its schedule sets $\lambda$ from the energy scale, not from
the window; \S\ref{sec:solver-objective} states that entering
the window is not claimed. Second, the certificate itself does
not read a support off the gate alone. Condition (C8) of the
formal certificate (Appendix~\ref{app:thm2}) requires every used
coefficient to clear a floor $\vartheta$, so the support the
certificate evaluates is the set of selected slots whose
coefficient clears the floor. With $\vartheta = 0.5$, below the
true floor of one, the returned solutions have usage ratios of
$0.96$ to $1.10$ and exact supports on $68$ to $89\%$ of samples at the reported checkpoints (Table~\ref{tab:support}); across all bands the range is $14$ to $96\%$, with the low values at early checkpoints where the pool is still incomplete (Table~\ref{tab:support-bands}).
Under the certificate's own definition the returned solutions
are close to usage-minimal; what over-selects is the gate as a
stand-alone reader of the support. Third, the theory does not
cover the formation of the pool before the anchor of
Proposition~\ref{prop:dyn}, and the trajectory of
Table~\ref{tab:support-bands} lives there: the checkpoints at
which the gate is near usage-minimal are the early ones, taken
while the pool is still incomplete, and the checkpoints with a
complete pool are the late ones, taken after the gate has
saturated. The open problem is therefore stated precisely: a
schedule that holds the per-sample penalty inside the
$\lambda$-window while the pool completes. Theorem~\ref{thm:ident}
gives the window's location in terms of the noise level, so the
direction of that change is known, and the certificate is the
test of whether any such schedule succeeds. What the theory
leaves to the data is the isolating-pair ceiling: at $4N$
samples with random supports of one to four components, between
a quarter and a third of the components admit a pair even under
the true supports, so the condition of Conjecture~R is not met
here for any solver, and the exact column is a description of
these runs, not a test of that conjecture.

\begin{table*}[t]
\centering
\caption{Per-sample support recovery of EB-gMCR on the synthetic benchmark at the checkpoint band of Table~\ref{tab:e2} (pool $1024$, $4N$ samples, five seeds). Columns are defined in the text.}
\label{tab:support}
\scriptsize
\resizebox{\textwidth}{!}{%
\begin{tabular}{cc|cccccc|cc}
\toprule
$N$ & SNR & usage ratio & exact & no-miss & min cos & misses & isolating pairs (ceiling) & usage, $\hat c>0.5$ & exact, $\hat c>0.5$ \\
\midrule
16 & 20 & 7.025$\pm$0.828 & 0.000$\pm$0.000 & 1.000$\pm$0.000 & 0.978$\pm$0.010 & 0.0$\pm$0.0 & 0.0$\pm$0.0 / 16 (5.4) & 1.104$\pm$0.062 & 0.756$\pm$0.140 \\
16 & 30 & 1.837$\pm$0.607 & 0.144$\pm$0.196 & 0.931$\pm$0.041 & 0.991$\pm$0.008 & 0.2$\pm$0.4 & 0.6$\pm$1.3 / 16 (5.4) & 1.038$\pm$0.089 & 0.831$\pm$0.187 \\
32 & 20 & 10.581$\pm$0.948 & 0.000$\pm$0.000 & 0.991$\pm$0.007 & 0.980$\pm$0.007 & 0.0$\pm$0.0 & 0.0$\pm$0.0 / 32 (8.6) & 1.042$\pm$0.043 & 0.889$\pm$0.097 \\
32 & 30 & 9.849$\pm$3.795 & 0.000$\pm$0.000 & 0.886$\pm$0.045 & 0.904$\pm$0.058 & 1.8$\pm$1.1 & 0.0$\pm$0.0 / 32 (8.6) & 1.102$\pm$0.046 & 0.775$\pm$0.132 \\
64 & 20 & 11.491$\pm$7.872 & 0.000$\pm$0.000 & 0.885$\pm$0.097 & 0.877$\pm$0.118 & 0.4$\pm$0.5 & 0.0$\pm$0.0 / 64 (19.4) & 1.062$\pm$0.069 & 0.780$\pm$0.191 \\
64 & 30 & 1.110$\pm$0.210 & 0.393$\pm$0.339 & 0.742$\pm$0.028 & 0.800$\pm$0.171 & 0.2$\pm$0.4 & 10.6$\pm$9.8 / 64 (19.4) & 0.962$\pm$0.061 & 0.675$\pm$0.078 \\
\bottomrule
\end{tabular}}

\end{table*}

\begin{table*}[t]
\centering
\caption{The runs of Table~\ref{tab:support} decoded at every retained checkpoint band ($n$ runs retained the band). Columns are defined in the text.}
\label{tab:support-bands}
\scriptsize
\resizebox{\textwidth}{!}{%
\begin{tabular}{ccc c c c c c c c c}
\toprule
$N$ & SNR & band & $n$ & epoch & usage ratio & no-miss & min cos & exact & exact, $\hat c>0.5$ & spurious $\hat c$ \\
\midrule
16 & 20 & 0.97--0.975 & 5 & 0.53$\pm$0.27 & 6.9$\pm$0.6 & 0.97$\pm$0.04 & 0.94$\pm$0.07 & 0.00$\pm$0.00 & 0.73$\pm$0.17 & 0.06$\pm$0.03 \\
16 & 20 & 0.975--0.98 & 5 & 0.52$\pm$0.27 & 6.8$\pm$0.8 & 0.98$\pm$0.04 & 0.96$\pm$0.03 & 0.00$\pm$0.00 & 0.79$\pm$0.10 & 0.04$\pm$0.02 \\
16 & 20 & 0.98--0.985 & 5 & 0.47$\pm$0.19 & 7.0$\pm$0.8 & 1.00$\pm$0.00 & 0.98$\pm$0.01 & 0.00$\pm$0.00 & 0.76$\pm$0.14 & 0.03$\pm$0.01 \\
16 & 20 & 0.985--0.99 & 5 & 0.91$\pm$0.10 & 8.3$\pm$0.6 & 1.00$\pm$0.00 & 0.97$\pm$0.02 & 0.00$\pm$0.00 & 0.74$\pm$0.11 & 0.01$\pm$0.00 \\
16 & 20 & 0.99--0.995 & 4 & 0.08$\pm$0.02 & 2.0$\pm$0.3 & 0.77$\pm$0.07 & 0.93$\pm$0.03 & 0.05$\pm$0.05 & 0.14$\pm$0.03 & 1.33$\pm$0.95 \\
\midrule
16 & 30 & 0.97--0.975 & 5 & 0.77$\pm$0.24 & 5.6$\pm$0.3 & 0.79$\pm$0.04 & 0.96$\pm$0.02 & 0.00$\pm$0.00 & 0.61$\pm$0.12 & 0.05$\pm$0.04 \\
16 & 30 & 0.975--0.98 & 5 & 0.46$\pm$0.18 & 4.3$\pm$1.9 & 0.82$\pm$0.07 & 0.97$\pm$0.01 & 0.00$\pm$0.00 & 0.70$\pm$0.12 & 0.06$\pm$0.04 \\
16 & 30 & 0.98--0.985 & 5 & 0.64$\pm$0.31 & 4.6$\pm$1.7 & 0.87$\pm$0.04 & 0.95$\pm$0.07 & 0.00$\pm$0.01 & 0.66$\pm$0.15 & 0.07$\pm$0.07 \\
16 & 30 & 0.985--0.99 & 5 & 0.41$\pm$0.31 & 3.7$\pm$2.4 & 0.91$\pm$0.05 & 0.98$\pm$0.01 & 0.17$\pm$0.33 & 0.78$\pm$0.12 & 0.09$\pm$0.07 \\
16 & 30 & 0.99--0.995 & 5 & 0.19$\pm$0.08 & 1.8$\pm$0.6 & 0.93$\pm$0.04 & 0.99$\pm$0.01 & 0.14$\pm$0.20 & 0.83$\pm$0.19 & 0.10$\pm$0.05 \\
16 & 30 & 0.995--1.0 & 5 & 0.20$\pm$0.04 & 2.1$\pm$0.5 & 0.97$\pm$0.03 & 1.00$\pm$0.00 & 0.03$\pm$0.07 & 0.96$\pm$0.04 & 0.10$\pm$0.03 \\
\midrule
32 & 20 & 0.97--0.975 & 5 & 0.45$\pm$0.15 & 9.6$\pm$1.7 & 0.91$\pm$0.04 & 0.91$\pm$0.09 & 0.00$\pm$0.00 & 0.78$\pm$0.08 & 0.03$\pm$0.01 \\
32 & 20 & 0.975--0.98 & 5 & 0.58$\pm$0.19 & 10.6$\pm$0.7 & 0.96$\pm$0.01 & 0.93$\pm$0.09 & 0.00$\pm$0.00 & 0.84$\pm$0.12 & 0.02$\pm$0.01 \\
32 & 20 & 0.98--0.985 & 5 & 0.55$\pm$0.17 & 10.6$\pm$0.9 & 0.99$\pm$0.01 & 0.98$\pm$0.01 & 0.00$\pm$0.00 & 0.89$\pm$0.10 & 0.02$\pm$0.01 \\
32 & 20 & 0.985--0.99 & 5 & 0.48$\pm$0.40 & 9.6$\pm$5.9 & 0.89$\pm$0.15 & 0.97$\pm$0.01 & 0.00$\pm$0.00 & 0.56$\pm$0.20 & 0.05$\pm$0.06 \\
\midrule
32 & 30 & 0.97--0.975 & 5 & 0.55$\pm$0.22 & 9.0$\pm$1.3 & 0.74$\pm$0.03 & 0.94$\pm$0.02 & 0.00$\pm$0.00 & 0.60$\pm$0.08 & 0.04$\pm$0.02 \\
32 & 30 & 0.975--0.98 & 5 & 0.53$\pm$0.18 & 9.8$\pm$1.2 & 0.78$\pm$0.02 & 0.93$\pm$0.03 & 0.00$\pm$0.00 & 0.63$\pm$0.10 & 0.03$\pm$0.01 \\
32 & 30 & 0.98--0.985 & 5 & 0.65$\pm$0.27 & 10.6$\pm$0.9 & 0.82$\pm$0.04 & 0.93$\pm$0.02 & 0.00$\pm$0.00 & 0.69$\pm$0.08 & 0.02$\pm$0.01 \\
32 & 30 & 0.985--0.99 & 5 & 0.74$\pm$0.23 & 11.3$\pm$0.6 & 0.88$\pm$0.03 & 0.88$\pm$0.12 & 0.00$\pm$0.00 & 0.69$\pm$0.14 & 0.02$\pm$0.00 \\
32 & 30 & 0.99--0.995 & 5 & 0.69$\pm$0.29 & 9.8$\pm$3.8 & 0.89$\pm$0.04 & 0.90$\pm$0.06 & 0.00$\pm$0.00 & 0.78$\pm$0.13 & 0.04$\pm$0.04 \\
32 & 30 & 0.995--1.0 & 4 & 0.14$\pm$0.04 & 2.4$\pm$0.5 & 0.78$\pm$0.07 & 0.99$\pm$0.01 & 0.00$\pm$0.00 & 0.62$\pm$0.18 & 0.16$\pm$0.06 \\
\midrule
64 & 20 & 0.97--0.975 & 5 & 0.78$\pm$0.21 & 17.4$\pm$4.4 & 0.86$\pm$0.01 & 0.88$\pm$0.08 & 0.00$\pm$0.00 & 0.79$\pm$0.01 & 0.01$\pm$0.00 \\
64 & 20 & 0.975--0.98 & 5 & 0.65$\pm$0.22 & 15.7$\pm$6.2 & 0.90$\pm$0.02 & 0.86$\pm$0.11 & 0.00$\pm$0.00 & 0.84$\pm$0.04 & 0.01$\pm$0.01 \\
64 & 20 & 0.98--0.985 & 5 & 0.41$\pm$0.32 & 11.5$\pm$7.9 & 0.89$\pm$0.10 & 0.88$\pm$0.12 & 0.00$\pm$0.00 & 0.78$\pm$0.19 & 0.03$\pm$0.02 \\
64 & 20 & 0.985--0.99 & 3 & 0.36$\pm$0.48 & 11.4$\pm$13.3 & 0.83$\pm$0.14 & 0.96$\pm$0.02 & 0.00$\pm$0.00 & 0.57$\pm$0.20 & 0.06$\pm$0.05 \\
\midrule
64 & 30 & 0.97--0.975 & 5 & 0.54$\pm$0.23 & 12.3$\pm$4.6 & 0.71$\pm$0.03 & 0.91$\pm$0.01 & 0.00$\pm$0.00 & 0.56$\pm$0.03 & 0.02$\pm$0.01 \\
64 & 30 & 0.975--0.98 & 5 & 0.71$\pm$0.26 & 15.2$\pm$5.6 & 0.76$\pm$0.02 & 0.89$\pm$0.03 & 0.00$\pm$0.00 & 0.63$\pm$0.03 & 0.02$\pm$0.01 \\
64 & 30 & 0.98--0.985 & 5 & 0.65$\pm$0.13 & 15.9$\pm$6.5 & 0.79$\pm$0.01 & 0.92$\pm$0.03 & 0.00$\pm$0.00 & 0.73$\pm$0.02 & 0.02$\pm$0.01 \\
64 & 30 & 0.985--0.99 & 5 & 0.79$\pm$0.28 & 19.0$\pm$8.7 & 0.83$\pm$0.02 & 0.92$\pm$0.04 & 0.00$\pm$0.00 & 0.74$\pm$0.06 & 0.02$\pm$0.03 \\
64 & 30 & 0.99--0.995 & 5 & 0.12$\pm$0.07 & 1.1$\pm$0.2 & 0.74$\pm$0.03 & 0.80$\pm$0.17 & 0.39$\pm$0.34 & 0.68$\pm$0.08 & 0.65$\pm$0.64 \\
64 & 30 & 0.995--1.0 & 1 & 0.09$\pm$0.00 & 1.1$\pm$0.0 & 0.69$\pm$0.00 & 0.97$\pm$0.00 & 0.45$\pm$0.00 & 0.50$\pm$0.00 & 1.75$\pm$0.00 \\
\bottomrule
\end{tabular}}
\end{table*}

\section{Real-Data Evaluation Details}
\label{app:f}

\paragraph{Datasets.}
\textbf{Carbs}: $M=21$ samples in $\mathbb{R}^{1401}$, linear mixtures of $N=3$ non-negative components plus noise.
\textbf{NIR}: $M=166$ samples in $\mathbb{R}^{235}$; the signals take negative values; the concentrations of two analytes are known, the component profiles are not, and the two-analyte count is taken as the reference count.
On NIR the absolute-value steps of $f_c$ and $f_s$ (Appendix~\ref{app:repro}) are removed to admit signed signals; without positivity, Theorem~\ref{thm:cert} resolves each component only up to sign, and recovery on NIR is correct up to sign.
On Carbs the candidate pool has more slots than there are samples ($128 > M = 21$). In this regime the usage condition of the certificate alone does not exclude a decomposition with one slot per sample, and the learned-pool guarantee rests on assumption (C7) of Appendix~\ref{app:thm2}, which is not checked; the three-component result is consistent with the certificate under that assumption.

\paragraph{Baselines and preprocessing.}
The baselines choose their count by a variance sweep, increasing $N$ from 1 until $R^2$ stabilizes (cap 10); this is common practice, although other selection criteria have been proposed \citep{smith2019mcrals}.
For methods constrained to non-negativity, a sign-splitting transform is applied.
EB-gMCR starts from 128 candidate components and stops at its termination rule or, on the real datasets, at a fixed iteration limit (Appendix~\ref{app:repro}).
Checkpoint bands begin at $R^2 = 0.95$. On NIR two components reach at most about $R^2 = 0.91$ for every method, so every retained band holds three or more components, and the two-component solution reported for EB-gMCR is the state at the end of the run.

\paragraph{Figure~\ref{fig:real}.}
EB-gMCR is plotted at the counts its checkpoints and its end state returned, ARD-NMF at its best $\phi$ for each count. Series outside the axis range are omitted: all methods at $K \le 2$ on Carbs and at $K = 1$ on NIR, ICA on Carbs, and Sparse-NMF, ICA, and MCR-ALS on NIR.

\paragraph{Metrics.}
Components and concentrations are aligned to the references by linear assignment.
Reported: $R^2(D)\uparrow$, Cos.$(S)\uparrow$, RMSE$(S)\downarrow$, RMSE$(C)\downarrow$.

\begin{figure}[t]
\centering
\includegraphics[width=0.7\textwidth]{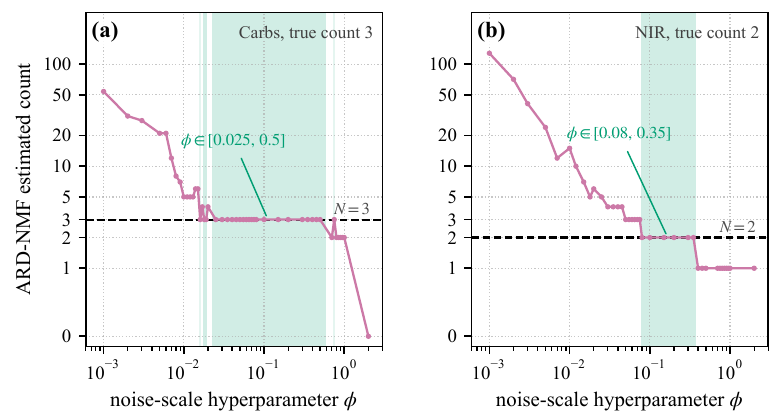}
\caption{ARD-NMF and its noise-scale hyperparameter $\phi$ on the
real datasets: returned count against $\phi$ on (a) Carbs, true
count 3, and (b) NIR, true count 2, shading the $\phi$ values
that return the true count.}
\label{fig:ardnmf}
\end{figure}

\paragraph{ARD-NMF and the certificate.} The certificate of
Theorem~\ref{thm:cert} does not depend on the solver: any returned
decomposition can be checked for fit, and its count read against
the truth where the truth is known. Figure~\ref{fig:ardnmf} shows
ARD-NMF's $\phi$ sweep under that check. The $\phi$ values that
return the true count, about $0.025$ to $0.5$ on Carbs (with
isolated values outside) and $0.08$ to $0.35$ on NIR, are the
region in which its output has
the right count; outside that region the output still
reconstructs well, and nothing in a single run tells the two
cases apart, since $\phi$ is a variance scale with no check of
its own. The same sweep on the synthetic grid is in
Appendix~\ref{app:e-ard}. The corresponding input of EB-gMCR is the $R^2$ band of
\S\ref{sec:solver-checkpoint}, a reconstruction quality that is
the fit condition of the certificate itself, checkable on every
checkpoint, with several bands returned per run. At a $\phi$
inside its region, ARD-NMF's decomposition is as good as any in
Table~\ref{tab:f1}.

\begin{figure}[t]
\centering
\includegraphics[width=0.5\textwidth]{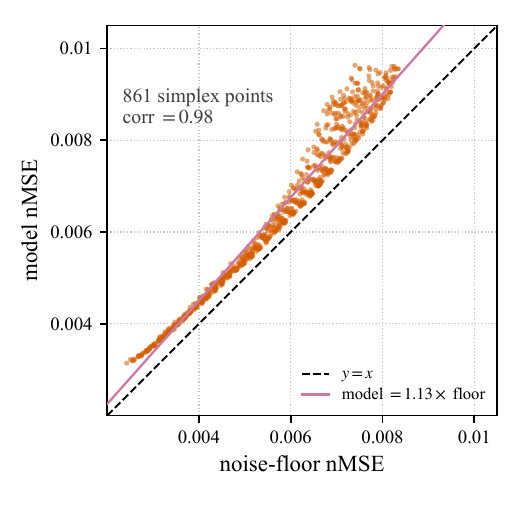}
\caption{Frozen-model reuse on the Carbs simplex: model nMSE
against the noise-floor nMSE at each of the $861$ unseen
mixtures, with the line $y = x$ and the fitted constant ratio.}
\label{fig:floor}
\end{figure}

\begin{table}[t]
\centering
\caption{Real-data decomposability at the ground-truth component count. ARD-NMF is scored at the best-$R^2$ value of its noise-scale hyperparameter $\phi$ among those that return the true count.}
\label{tab:f1}
\small
\begin{tabular}{lcccc|cc}
\toprule
& \multicolumn{4}{c|}{Carbs ($N=3$)} & \multicolumn{2}{c}{NIR ($N=2$)} \\
\cmidrule(lr){2-5} \cmidrule(lr){6-7}
Solver & $R^2(D)$ & RMSE$(C)$ & Cos.$(S)$ & RMSE$(S)$ & $R^2(D)$ & RMSE$(C)$ \\
\midrule
NMF & 0.963 & 0.070 & 0.992 & 0.929 & 0.876 & 32.46 \\
Sparse-NMF & 0.963 & 0.068 & 0.993 & 0.924 & -131.4 & 42.33 \\
Bayes-NMF & 0.962 & 0.072 & 0.992 & 0.931 & 0.888 & 34.06 \\
ICA & -6.206 & 0.338 & 0.462 & 6.564 & -66.99 & 23.47 \\
MCR-ALS & \textbf{0.964} & \textbf{0.026} & 0.996 & 0.610 & -25.17 & 27.63 \\
ARD-NMF (tuned $\phi$) & 0.951 & \textbf{0.026} & \textbf{0.997} & \textbf{0.490} & 0.902 & 31.72 \\
EB-gMCR & \textbf{0.964} & 0.032 & 0.996 & 0.628 & \textbf{0.910} & 24.49 \\
\bottomrule
\end{tabular}
\end{table}

\end{document}

%% file: math_commands.tex
\usepackage{amsmath,amsfonts,bm}

\def\eqref#1{equation~\ref{#1}}
\def\1{\bm{1}}

\DeclareMathAlphabet{\mathsfit}{\encodingdefault}{\sfdefault}{m}{sl}
\SetMathAlphabet{\mathsfit}{bold}{\encodingdefault}{\sfdefault}{bx}{n}

\newcommand{\E}{\mathbb{E}}

\newcommand{\R}{\mathbb{R}}